\documentclass[pdflatex,sn-apa]{sn-jnl}

\usepackage{amsmath,amssymb,amsfonts}
\usepackage{amsthm}
\usepackage{mathrsfs}
\usepackage{mathtools}
\usepackage{mleftright}
\usepackage{dblfloatfix}
\usepackage{placeins}
\theoremstyle{thmstyleone}

\theoremstyle{thmstyletwo}

\theoremstyle{thmstylethree}

\usepackage{algorithm}
\usepackage{algpseudocode}

\algrenewcommand\algorithmicrequire{\textbf{Input:}}
\algrenewcommand\algorithmicensure{\textbf{Output:}}

\usepackage{graphicx}
\usepackage{subcaption}          % Use subcaption OR subfigure — NOT both
\usepackage{booktabs}
\usepackage{multirow}
\usepackage{tabularx}
\usepackage{array}
\usepackage{adjustbox}
\usepackage{makecell}
\usepackage{siunitx}
\usepackage{float}
\usepackage[table]{xcolor}       % Load ONCE with the table option

\usepackage{tikz}
\usetikzlibrary{positioning, calc, arrows.meta, fit, backgrounds}

\usepackage[most,breakable,skins,listings]{tcolorbox}  % Load ONCE with all options

\usepackage{microtype}
\usepackage{lmodern}
\usepackage{textcomp}
\usepackage{alltt}
\usepackage{fvextra}
\usepackage{enumitem}
\usepackage{multicol}
\usepackage{needspace}
\usepackage{capt-of}
\usepackage{url}
\usepackage[title]{appendix}
\usepackage{chngcntr}

\newcounter{mainfigurecounter}
\newcounter{maintablecounter}

\usepackage{etoc}
\usepackage{algorithm}
\usepackage{algorithmicx}
\usepackage{algpseudocode}
\usepackage{listings}
\usepackage{manyfoot}
\usepackage{pifont}
\usepackage{svg}
\usepackage[textsize=tiny]{todonotes}

\usepackage{hyperref}            % Load ONCE, always last

\algrenewcommand\algorithmiccomment[1]{\hfill$\triangleright$ #1}

\begin{document}

\title[Article Title]{STaR-DRO: Stateful Tsallis Reweighting for Group-Robust Structured Prediction}


% \author*[1,2]{\fnm{First} \sur{Author}}\email{iauthor@gmail.com}

% % \equalcont{These authors contributed equally to this work.}


% \affil*[1]{\orgdiv{Department}, \orgname{Organization}, \orgaddress{\street{Street}, \city{City}, \postcode{100190}, \state{State}, \country{Country}}}

% \affil[2]{\orgdiv{Department}, \orgname{Organization}, \orgaddress{\street{Street}, \city{City}, \postcode{10587}, \state{State}, \country{Country}}}

% \affil[3]{\orgdiv{Department}, \orgname{Organization}, \orgaddress{\street{Street}, \city{City}, \postcode{610101}, \state{State}, \country{Country}}}

\author*[1]{\fnm{Samah} \sur{Fodeh}}\email{samah.fodeh@yale.edu}
\author[1]{\fnm{Ganesh} \sur{Puthiaraju}}
\author[1]{\fnm{Elyas} \sur{Irankhah}}
\author[1]{\fnm{Afshan} \sur{Khan}}
\author[1]{\fnm{Sreeraj} \sur{Ramachandran}}
\author[1]{\fnm{Linhai} \sur{Ma}}
\author[1]{\fnm{Srivani} \sur{Talakokkul}}
\author[2]{\fnm{Sarah} \sur{Schellhorn}}

\affil[1]{\orgname{Yale University}, \orgaddress{\city{New Haven}, \state{CT}, \country{USA}}}

\affil[2]{\orgname{Medical Oncology, Yale School of Medicine}, \orgaddress{\city{New Haven}, \state{CT}, \country{USA}}}

\abstract{
Structured prediction with large language models requires more than accurate labels: outputs must satisfy ontology constraints, preserve valid structure, and ensure evidence-grounded decisions under label imbalance and heterogeneous group difficulty. To resolve these challenges, we present a unified framework for robust ontology-constrained generation. First, we introduce a modular prompt-engineering architecture combining XML-style structure, expert disambiguation rules, chain-of-thought reasoning, metadata-aware decision logic, schema contracts, and a self-validation gate. This targets recurrent in-context structured generation failures, including format drift, label ambiguity, evidence hallucination, and metadata-conditioned confusion. Second, we propose STaR-DRO, which combines Tsallis mirror ascent with sparse entmax-style primal mapback, EMA-smoothed group-loss tracking, rescaled ascent signals, and bounded excess-only multipliers. This upweights only persistently hard groups without suppressing easier ones, addressing conventional DRO’s dense Shannon-entropy exponentiated-gradient updates, high-variance stochastic reweighting, positive adversarial mass on groups that are not persistently hard, and performance costs from simplex competition. We evaluate the framework on EPPC Miner, a carefully chosen, clinically grounded representation of high-stakes structured prediction requiring hierarchical label prediction and evidence-span extraction from patient-provider secure messages. Across Llama models spanning 1B to 70B parameters, our prompt engineering improves zero-shot extraction, with an average label F1 gain of +14.46 and a +17.40 Span F1 gain. Building on supervised fine-tuning, STaR-DRO further improves accuracy and robustness, increasing average label F1 by +1.08/+2.20 and reducing average group-wise validation cross-entropy by 21.3\%/14.8\% relative to SFT and standard DRO, respectively, significantly advancing the reliability of automated communication mining for patient-centered clinical care analysis.}

\keywords{Structured prediction, Prompt engineering, Group-robust learning, Tsallis mirror descent, Clinical information extraction, Large language models}

\maketitle

\section{Introduction}
\label{sec:introduction}

Structured prediction lies at the core of many high-value language technologies. In these settings, a model must generate outputs that satisfy explicit structural constraints, respect ontology or schema dependencies, and often ground each decision in supporting evidence from the input. This requirement arises across information extraction, where errors cause \emph{missed evidence}; legal document analysis, where they introduce \emph{compliance risk}; and biomedical NLP, where they propagate \emph{scientific noise}. The challenge is especially consequential in medicine, where digital communication has become an increasingly important substrate of care. Patient--clinician secure messaging now carries symptom escalation, medication management, diagnostic follow-up, care coordination, and increasingly the kind of between-visit medical decision-making that can supplement or substitute for in-person care, while also supporting the deliberative exchange central to patient-centered and shared decision-making practice \citep{wec2025measurement,north2020retrospective}.

Large language models have made such tasks newly approachable by recasting them as instruction-following generation problems \citep{ouyang2022training}. Recent instruction-tuned models show strong zero-shot and few-shot capabilities across a wide range of biomedical and general NLP settings, and benchmark efforts such as BLURB and CBLUE have helped clarify both the promise and the limits of current systems on specialized biomedical tasks \citep{gu2022domain,zhang2022cblue}. More broadly, the prompting literature has shown that carefully structured instructions, intermediate reasoning, and self-feedback can substantially improve model behavior \citep{wei2022chain}. Yet structured prediction remains a difficult regime for prompting alone. The failure modes are systematic and identifiable: format drift over long structured outputs, label ambiguity at ontological boundaries, reasoning shortcuts that bypass evidence verification, hallucinated evidence spans, and metadata-conditioned confusion in tasks whose label semantics depend on source, role, direction, or local discourse state. What is often missing is not another task-specific prompt tweak, but a reusable prompt architecture that directly targets these recurrent structural failure modes.

A second obstacle arises during learning. In many structured prediction problems, the output space admits a natural grouping into semantically coherent categories, for example, communication codes, entity types, clause families, or event classes, and the distribution over these groups is strongly heterogeneous in both frequency and intrinsic difficulty. This heterogeneity creates a fundamental tension for model training. Standard empirical risk minimization optimizes average loss and can therefore mask systematic failures on minority groups: a model with low overall loss may still incur high loss on semantically ambiguous, structurally complex, or low-frequency categories \citep{sagawa2020distributionally,menon2021longtail}. The problem is amplified in hierarchical structured prediction, where errors compound across levels: a wrong entity type can invalidate downstream relation extraction, a wrong clinical code can distort quality-of-care surveillance, and a misclassified legal clause can alter obligation detection. Group distributionally robust optimization offers a principled response by prioritizing worst-group behavior over average-case fit \citep{sagawa2020distributionally, namkoong2016stochastic}. However, standard group DRO is not an ideal fit for modern structured prediction pipelines. Its stochastic update is typically implemented through exponentiated-gradient mirror ascent under Shannon-entropy geometry, which yields dense adversarial weights over all groups. In practice, this leads to five intertwined limitations: the update is \emph{stateless} with respect to longer-horizon group difficulty, \emph{variance-sensitive} when minibatches sparsely cover groups, often translated into \emph{asymmetric} or highly variable optimizer-facing reweighting, intrinsically \emph{dense} in its redistribution of mass, and prone to worsening the robustness--performance trade-off by turning worst-case simplex competition into unnecessary reweighting pressure on groups that are not actually hard.

This paper addresses both problems through a unified framework. First, we introduce a \textbf{modular prompt engineering strategy for structured prediction}. The framework decomposes the prompt into explicit components for XML-style instruction organization, expert-curated disambiguation rules, verification-style chain-of-thought reasoning, metadata-aware decision logic, strict schema contracts, and a self-validation quality gate. Although instantiated here on a clinical annotation task, its logic is general: the modules are designed to make structured decision problems legible to language models whenever outputs must be valid, grounded, and reproducible. Each module targets a distinct, identified failure mode, and each is defined through replaceable placeholders, so the template adapts to new domains by substituting label inventories, disambiguation criteria, metadata variables, and output schemas while retaining the architectural benefits.

Second, we introduce STaR-DRO, a \textbf{stateful robust optimization method for supervised learning under meaningful group heterogeneity}. STaR-DRO replaces dense Shannon-entropy mirror updates with Tsallis mirror ascent \citep{tsallis1988possible}, uses sparse entmax-style primal mapback to focus adversarial mass on persistently hard groups \citep{peters2019sparse}, tracks group difficulty through momentum-smoothed and centered loss signals rather than raw single-batch fluctuations, and converts adversarial weights into bounded \emph{excess-only} multipliers so that only groups above the neutral baseline are upweighted. At the optimization level, this yields a more selective and more stable robust objective. At the modeling level, it avoids turning worst-case simplex competition into explicit downweighting of easier groups, instead translating robust emphasis into selective upweighting of genuinely harder groups. At the representation level, it can be applied at annotation-token granularity, directing extra gradient signal to the semantically decisive regions of a structured output rather than diffusing robust pressure across an entire example. At the temporal level, exponential moving averages provide stateful estimates of group difficulty, allowing the adversarial update to respond to persistent hardness rather than transient minibatch noise.

We evaluate this framework on EPPC Miner \citep{carini2021impact, fodeh2026pvminer, fodeh2026eppcminerben, fodeh2026tab, fodeh2026pvminerllm}, but EPPC Miner is best understood here as a demanding clinical instantiation of a broader methodological setting. The task requires the model to predict structured communication annotations from electronic patient--provider secure messages, including hierarchical Code/Sub-code assignments and exact supporting evidence spans. This makes it a strong stress test for general structured prediction because success requires simultaneous semantic discrimination, structural validity, and grounded extraction. At the same time, its clinical significance is immediate. The difficult and infrequent groups are not arbitrary tail labels; they correspond to communication behaviors such as shared decision-making, socioemotional support, partnership, and social-determinant-related exchange: signals that matter for care escalation, treatment follow-through, relational continuity, and intervention design. In this setting, robustness is not merely a statistical preference. It is directly tied to whether automated communication mining remains clinically informative on the cases that are easiest to miss and most important to understand.

\paragraph{Contributions.}
\begin{itemize}[leftmargin=*]

\item \textbf{Modular prompt engineering for structured prediction.} We introduce a six-module, failure-mode-targeted,  reusable and domain-general prompt architecture for ontology-constrained generation with grounded evidence. On EPPC Miner, it improves zero-shot performance across four Llama models by an average of $+15.90$ Code~F1, $+13.02$ Sub-code~F1, and $+17.40$ Span~F1. It also rescues \texttt{Llama-3.1-8B-Instruct} from complete zero-shot semantic collapse ($0.00/0.00$ Code/Sub-code~F1) to task-usable extraction ($52.84/38.26$), and delivers the largest cross-family Span gain on \texttt{gemma-2-2b-it} ($+27.31$ F1 points), showing that the framework transfers consistently across Llama, Qwen, and Gemma without architecture-specific redesign.

\item \textbf{STaR-DRO: a task-agnostic robust fine-tuning method.} We introduce a stateful sparse DRO method that combines Tsallis mirror ascent, entmax-style sparse primal mapback, EMA-smoothed group tracking, rescaled ascent signals, and bounded excess-only multipliers. Building on SFT, STaR-DRO improves Code and Sub-code~F1 on all four Llama scales, reaching $81.47$ Code~F1 and $69.30$ Sub-code~F1 on \texttt{Llama-3.3-70B-Instruct}, while keeping Span~F1 tightly clustered above $91$. In contrast, standard group DRO degrades the three smaller Llama models, with average changes of $-1.91$ Code~F1 and $-1.64$ Sub-code~F1, showing that the sparse, stateful, and bounded excess-only multiplier design is essential rather than incremental.

\item \textbf{Cross-architecture robustness under group heterogeneity.} STaR-DRO reduces group-wise validation cross-entropy in $70$ of $72$ model--group combinations, with average reductions reaching up to $36.0\%$. The same clinically difficult groups (\emph{PartnershipPatient}, \emph{PartnershipProvider}, and \emph{SharedDecisionPatient}) often receive among the largest reductions across Llama, Qwen, and Gemma, indicating that the gains are driven by task-level group structure rather than architecture-specific dynamics. Ablation studies further identify that stronger regularization improves all nine groups while compressing the inter-group validation-loss spread from approximately $[0.22,0.34]$ to $[0.21,0.29]$, a roughly $2\times$ reduction.

\end{itemize}

\begin{figure*}[!htbp]
    \centering
    \includegraphics[width=0.93\textwidth]{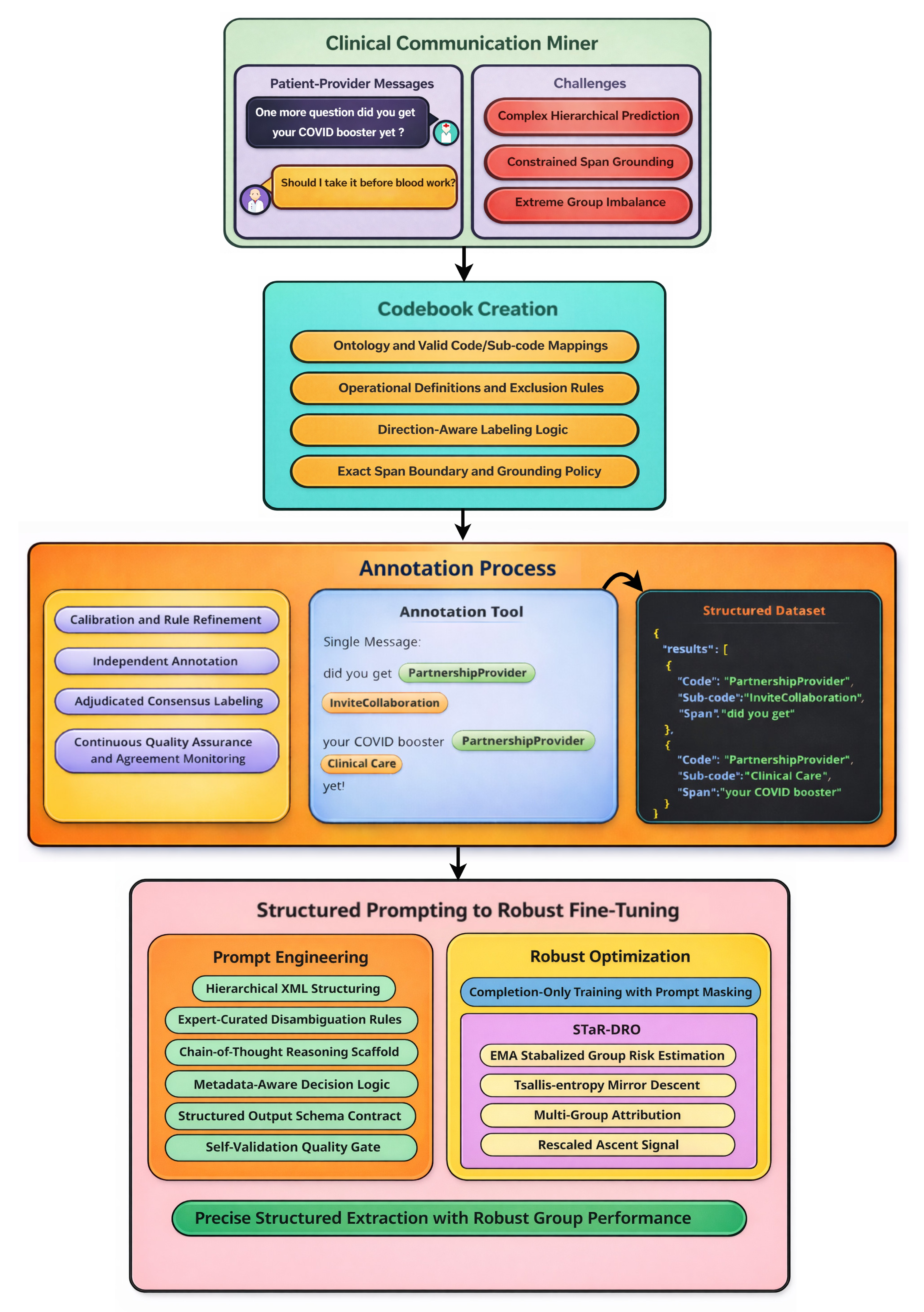}
    \caption{Pipeline of EPPC Miner. The pipeline progresses from clinical communication mining and codebook creation to annotation-driven dataset construction, followed by structured prompting, supervised fine-tuning, and STaR-DRO-based robust optimization.}
    \label{fig:workflow}
\end{figure*}

As shown in Figure~\ref{fig:workflow}, the pipeline spans clinical communication mining, codebook construction, annotation and dataset creation, and structured prompting with robust fine-tuning, with evaluation at the Code, Sub-code, and Span levels together with group-wise analysis.

% \midrule

\section{EPPC Miner}
While the experiments in this work are conducted on EPPC Miner, the methodological scope of the paper is broader and is fundamentally task-agnostic. We consider a class of hierarchical structured generation and prediction problems in which the output consists of semantically constrained labels and grounded textual evidence, and in which the underlying label space admits a natural grouping into semantically coherent categories. When the group-wise difficulty distribution across these categories is heterogeneous, as is common in real-world datasets spanning clinical, biomedical, legal, and general information extraction domains, this structure can be exploited to trade off average-case performance against worst-group robustness, an objective of particular importance in safety-critical and compliance-sensitive settings where
underperformance on rare but consequential categories carries disproportionate downstream risk. This setting encompasses annotation, named entity recognition, evidence extraction, and related information extraction tasks with ontology constraints or multi-level labeling structure.

EPPC Miner serves here as a rigorously constrained, representative instantiation of this problem class. It provides a clinically meaningful benchmark that demands hierarchical multi-labeling, strict Span grounding, direction-sensitive semantics, and pronounced heterogeneity in group difficulty. To systematically establish this benchmark, the remainder of this section formalizes the grounded structured classification task (Subsection~\ref{sec:task_formulation}), outlines the data acquisition and curation process (Subsection~\ref{sec:data_overview}), details the rigorous multi-stage annotation protocol (Subsection~\ref{sec:annotation_protocol}), and compares EPPC Miner against existing clinical NLP benchmarks (Subsection~\ref{sec:benchmark_comparison}).

\subsection{Task Formulation}
\label{sec:task_formulation}
 Formally, we cast EPPC Miner as a hierarchical communication pattern extraction problem over clinical secure messages. The model takes as input a target sentence $s$ alongside its local discourse context 
\[
C = (s_{\text{previous}}, s_{\text{next}}),
\]
where $s_{\text{previous}}$ and $s_{\text{next}}$ denote the immediately preceding and following sentences, respectively, and a direction indicator 
\[
d \in \{\texttt{Y}, \texttt{N}\},
\]
which specifies whether the message is directed to the patient or originates from the patient. 

Let 
\[
L_{\text{Code}} = \{\kappa_1, \dots, \kappa_K\}
\]
denote the inventory of communication Codes, and let
\[
L_{\text{Sub-code}} = \{\varsigma_1, \dots, \varsigma_M\}
\]
denote the inventory of Sub-codes. The hierarchical dependency between Codes and Sub-codes is defined by a validity mapping
\[
\mathcal{H}: L_{\text{Code}} \rightarrow 2^{L_{\text{Sub-code}}},
\]
where $\mathcal{H}(\kappa)$ gives the set of valid Sub-codes associated with Code $\kappa$ (Figure~\ref{fig:code_and_subcode}). The goal is to predict a structured output consisting of one or more grounded triples of the form $(\kappa,\varsigma,\pi)$, where $\kappa \in L_{\text{Code}}$ is a Code, $\varsigma \in \mathcal{H}(\kappa)$ is a valid Sub-code under $\kappa$, and $\pi$ is a supporting evidence Span drawn from the target sentence $s$. The model is therefore defined as
\begin{equation*}
f_{\theta} : (s, C, d) \longmapsto \widehat{\mathcal{Y}},
\end{equation*}
where
\begin{equation*}
\widehat{\mathcal{Y}} \subseteq \left\{ (\kappa,\varsigma,\pi)\ \middle|\ \kappa \in L_{\text{Code}},\ \varsigma \in \mathcal{H}(\kappa),\ \pi \subseteq s \right\}.
\end{equation*}

\subsection{Data acquisition and curation}\label{sec:data_overview}

We evaluate on a secure-message corpus curated from Yale New Haven Health (YNHH), partitioned into 448 training messages (272 patient-authored, 176 provider-authored) and 752 test messages (449 patient-authored, 303 provider-authored). After sentence segmentation, the corpus comprises 1{,}268 training sentences and 1{,}933 test sentences, totaling 15{,}746 training words and 27{,}849 test words. Message lengths vary substantially. The mean is $37.03 \pm 33.35$ words (test) and $35.15 \pm 30.85$ words (train), with maxima of 248 and 215 words, respectively. This heterogeneity creates a challenging extraction setting. Models must remain stable across terse single-sentence messages and extended multi-paragraph narratives. Secure messages are conversational in register. They often contain informal phrasing, implicit referents, and overlapping communicative intents within a single sentence. These conditions make boundaries between closely related codes inherently ambiguous. They also motivate the structured prompt engineering framework introduced in Section~\ref{sec:prompt_engineering}.

The resulting label distribution is heavily skewed at both levels of
the ontology, as shown in Figure~\ref{fig:eppc_subcodes}. At the Code level, the two most frequent
categories, \emph{InfoGive} (930 annotations) and \emph{PartnershipPatient} (839), together account for more than 65\% of all code-level
labels, while \emph{InfoSeekSDOH} (26) and \emph{SharedDecisionProvider} (57) collectively represent less than 4\%. At the Sub-code
level, the imbalance is more extreme: the top two Sub-codes, \emph{salutation}
(333) and \emph{signoff} (296), outnumber the bottom two, \emph{EncourageQuestions} (2) and \emph{PositiveRemarks} (6), by nearly 80-fold. The frequency ratio between the most and least common Sub-code
exceeds 160:1. This imbalance is not merely statistical: the rare
categories often capture clinically consequential communicative behaviors,
including shared decision-making, empathetic support, and
social-determinant disclosure, that are precisely the patterns most
valuable for downstream quality improvement and intervention design.
Standard empirical risk minimization, which optimizes average loss across
all groups, can mask systematic underperformance on these minority
categories. This observation provides the direct empirical motivation
for the group-robust objective developed in Section~\ref{sec:stardro}.

Annotation follows a structured EPPC codebook adapted from the Roter Interaction Analysis System (RIAS). The codebook defines nine communication codes and 38 Sub-codes in a strict hierarchy. Figure \ref{fig:eppc_four_categories} presents the full EPPC Miner communication ontology as a Code–Sub-code hierarchy. The ontology spans informational exchange, socioemotional and empathic communication, partnership behaviors, shared decision-making, and SDOH-related exchange, with each Sub-code valid only under its parent Code. This hierarchy defines the permissible Code–Sub-code pairings used throughout annotation and structured prediction in EPPC Miner.

\subsection{Annotation protocol and quality assurance}%
\label{sec:annotation_protocol}

Annotation proceeded in four stages designed to maximize consistency,
validity, and reproducibility.

\paragraph{Stage~1: Calibration.}
The full research team jointly annotated an initial set of 100 messages,
reviewed in batches of 20 across multiple calibration meetings. This
phase established shared interpretation of EPPC Code definitions,
surfaced systematic ambiguities, and refined operational coding rules
before large-scale labeling.

\paragraph{Stage~2: Independent dual annotation.}
Following calibration, two trained annotators independently labeled each
secure-message thread using the EPPC codebook. A thread was defined as a
contiguous sequence of patient-provider exchanges centered on a single
clinical concern or episode of care. In total, 13 batches of 50 messages
each were annotated under this protocol, yielding 650 doubly coded
messages.

\paragraph{Stage~3: Adjudication and consensus resolution.}
Disagreements between annotators were resolved through structured
adjudication meetings led by senior investigators with expertise in
communication coding and qualitative methods. Each disagreement was
reviewed against the codebook definitions and the disambiguation rules
established during calibration, and the consensus label was recorded as
the final gold standard.

\paragraph{Stage~4: Ongoing quality assurance.}
Inter-annotator agreement was monitored continuously. The mean Cohen's
$\kappa$ across all 13 batches was 0.74, with a steady upward trend
(batch~1: $\kappa = 0.67$; median: $\kappa = 0.74$; batch~13:
$\kappa = 0.76$), indicating progressive calibration convergence. In
parallel, a senior reviewer performed random audits and targeted spot
checks to verify codebook adherence and to identify emerging ambiguity
patterns. Feedback from these audits was incorporated into the annotation
manual iteratively, ensuring that evolving interpretive standards were
documented and applied consistently across all subsequent batches.

\subsection{Comparison to existing clinical NLP benchmarks}%
\label{sec:benchmark_comparison}

Existing biomedical NLP benchmarks capture only part of the EPPC problem
space. Resources such as BLURB~\citep{gu2021domain} and
CBLUE~\citep{zhang2022cblue} emphasize conventional extraction and
classification over biomedical text. Dialogue-focused datasets such as
MedDG~\citep{liu2022meddg} and MediTOD~\citep{saley2024meditod} evaluate
task-oriented interactions but do not represent the relational and
socio-emotional characteristics of real patient-provider secure messaging.
ReMeDi~\citep{yan2022remedi} is closer to clinical dialogue reasoning, yet
it is not designed for secure portal communication and does not support
multi-label EPPC annotation with strict Span grounding. EPPC Miner
addresses these gaps in a single benchmark. It targets U.S.\ patient-provider
secure messages, models bidirectional roles, and requires hierarchical
Code/Sub-code prediction with evidence-Span grounding. This combination
reflects the practical demands of communication mining in real clinical
workflows.

\begin{table*}[!htbp]
\centering
\small
\renewcommand{\arraystretch}{1.15}
\caption{
Comparison between EPPC Miner and existing biomedical NLP benchmarks.
$\checkmark$~indicates strong support, $\times$~indicates no support,
and $\triangle$~indicates partial or limited support.
Column abbreviations are as follows:
Rel./Socio-Emo. = relational and socio-emotional communication;
Bidir. Inter. = bidirectional interaction;
Multi-Label = multi-label annotation;
Secure Msg. = secure messaging.
Additional task abbreviations are:
NER = named entity recognition;
RE = relation extraction;
QA = question answering.
}
\label{tab:ben_compare_wide}
\resizebox{\textwidth}{!}{%
\begin{tabular}{@{}p{1.6cm}p{1.6cm}p{2.2cm}p{2.2cm}cccc@{}}
\toprule
\textbf{Bench-mark} &
\textbf{Domain} &
\textbf{Focus} &
\textbf{Data Source} &
\makecell[c]{\textbf{Rel./}\\\textbf{Socio-}\\\textbf{Emo.}} &
\makecell[c]{\textbf{Bidir.}\\\textbf{Inter.}} &
\makecell[c]{\textbf{Multi-}\\\textbf{Label}} &
\makecell[c]{\textbf{Secure}\\\textbf{Msg.}} \\
\midrule
CBLUE \cite{zhang2022cblue} &
Chinese biomedical &
NER, RE, classification &
Clinical text, dialogues &
$\times$ & $\times$ & $\times$ & $\times$ \\[3pt]

MedDG \cite{liu2022meddg} &
Chinese medical dialogue &
Diagnosis, symptom inquiry &
Simulated dialogues &
$\times$ & $\times$ & $\times$ & $\times$ \\[3pt]

ReMeDi \cite{yan2022remedi} &
English medical dialogue &
Medication, treatment reasoning &
Clinical conversations &
$\triangle$ & $\triangle$ & $\times$ & $\times$ \\[3pt]

MediTOD \cite{saley2024meditod} &
English history-taking &
Symptom elicitation, reasoning &
Simulated dialogues &
$\times$ & $\times$ & $\times$ & $\times$ \\[3pt]

BLURB \cite{gu2021domain} &
Biomedical NLP &
NER, RE, QA, summarization &
Biomedical literature &
$\times$ & $\times$ & $\times$ & $\times$ \\
\midrule

\textbf{EPPC Miner (Ours)} &
U.S.\ secure messaging &
Communication functions, relational behaviors &
De-identified secure msg. from portals &
$\checkmark$ & $\checkmark$ & $\checkmark$ & $\checkmark$ \\
\bottomrule
\end{tabular}%
}
\end{table*}

\section{Methods: Task-Aware Adaptation and Robust Optimization}
This section outlines the end-to-end pipeline for robust structured classification. We first introduce a modular prompt engineering framework to mitigate common in-context generation failures (Subsection~\ref{sec:prompt_engineering}) and establish a comprehensive prompting benchmark to characterize baseline task difficulty. Building upon this foundation, we detail our completion-only supervised fine-tuning approach (Subsection~\ref{sec:sft}), and finally present STaR-DRO, our stateful Tsallis robust optimization strategy designed to systematically target persistently hard semantic groups in hierarchical extraction (Subsection~\ref{sec:stardro}).

\subsection{Modular Prompt Engineering for Structured Prediction} 
\label{sec:prompt_engineering} 

The prompt engineering framework introduced in this work is designed for a broad class of structured prediction problems in which a model must emit ontology-constrained labels together with grounded textual evidence under explicit control variables. Although we instantiate and evaluate it on the EPPC Miner clinical annotation task, the underlying design is general purpose: each component addresses a generic failure mode of in-context structured generation: format drift, label ambiguity, reasoning shortcuts, evidence hallucination, and role-conditioned confusion that arises across structured prediction settings with hierarchical or multi-level labeling structure and strict Span fidelity requirements.

The prompt decomposes structured prediction into six explicitly
controlled modules.
\textbf{M1} (XML structuring) organizes the instruction into
semantically tagged blocks that separate global behavioral instructions from task-specific local
decision rules, improving constraint clarity in long contexts
\citep{white2023prompt}.
\textbf{M2} (disambiguation rules) encodes expert-derived decision
boundaries for confusable label pairs, operationalizing the annotation
manual inside the prompt
\citep{pang2023guideline,sainz2024gollie}.
\textbf{M3} (reasoning scaffold) imposes a four-step chain-of-thought style verification routine consisting of context analysis, phrase decomposition, Span extraction, and cross-validation before output emission
\citep{wei2022chain}.
\textbf{M4} (metadata-aware control) exposes task metadata as explicit
control variables, converting latent contextual attributes into
observed signals that guide context-aware structured extraction.
\citep{kong2024better,ouyang2022training}.
\textbf{M5} (schema contract) enforces machine-parseable output with
strict specifications on Span grounding, label-pair validity, and
multi-label coverage
\citep{li2024simple,sainz2024gollie}.
\textbf{M6} (quality gate) adds a single-turn self-validation
checklist that audits parse-ability, label validity, Span exactness,
and rule coverage within one inference pass, complementing but
distinct from iterative self-refinement approaches
\citep{madaan2023selfrefine,huang2024cannot}. Because each module is defined through replaceable placeholders,
the template readily adapts to other structured prediction problems
by substituting domain-specific label inventories, disambiguation
criteria, metadata variables, and output schemas while retaining the
architectural benefits. Figure \ref{fig:prompt_template} presents the
general-purpose template; the full EPPC Miner instantiation and
detailed component-level discussion are provided in
Appendix~\ref{app:prompt-engineering}.

\clearpage

% ─── Color definitions for the template box ─────────────────────────────────
\definecolor{tplFrame}{HTML}{4A148C}
\definecolor{tplBack}{HTML}{F3E5F5}
\definecolor{tplTitle}{HTML}{FFFFFF}
\definecolor{tplTitleBg}{HTML}{6A1B9A}
\definecolor{modXML}{HTML}{1565C0}
\definecolor{modDIS}{HTML}{7B1FA2}
\definecolor{modCOT}{HTML}{E65100}
\definecolor{modVAL}{HTML}{C62828}
\definecolor{modLOG}{HTML}{00838F}
\definecolor{modSCH}{HTML}{2E7D32}
\definecolor{slotGray}{HTML}{616161}

\begin{figure*}[!htbp]
\label{fig:prompt_template}
\begin{tcolorbox}[
  enhanced,
  breakable,
  colframe=tplFrame,
  colback=tplBack,
  colbacktitle=tplTitleBg,
  coltitle=tplTitle,
  title={\small\bfseries Prompt-Engineered Instruction Template for General Structured Prediction},
  fonttitle=\bfseries\large,
  boxrule=1.2pt,
  arc=3.5mm,
  top=4pt, bottom=4pt, left=8pt, right=8pt,
  drop fuzzy shadow=black!35,
]

\small

\vspace{-4pt}

\begin{tcolorbox}[
  enhanced,
  colback=white,
  colframe=modXML!60,
  boxrule=0.8pt,
  arc=2mm,
  left=6pt, right=6pt, top=4pt, bottom=3pt,
  title={\textcolor{modXML}{\textbf{M1\;\;$\vert$\;\;Hierarchical XML Structuring}}},
  fonttitle=\small\bfseries,
  colbacktitle=modXML!10,
  coltitle=modXML, 
]
\footnotesize\ttfamily
\textcolor{modXML}{<role>} \textcolor{slotGray}{\textit{[Domain expert persona and behavioral prior]}} \textcolor{modXML}{</role>}\\[1pt]
\textcolor{modXML}{<performance\_target>} \textcolor{slotGray}{\textit{[Precision-oriented accuracy expectations]}} \textcolor{modXML}{</performance\_target>}\\[1pt]
\textcolor{modXML}{<task>} \textcolor{slotGray}{\textit{[Task definition, input/output specification, multi-label constraint]}} \textcolor{modXML}{</task>}\\[1pt]
\textcolor{modXML}{<codes\_definitions>} \textcolor{slotGray}{\textit{[Authoritative label inventory with operational definitions]}} \textcolor{modXML}{</codes\_definitions>}\\[2pt]
\rmfamily\textcolor{slotGray}{\textit{Separates global instructions from local decision rules; improves constraint salience in long contexts.}}
\end{tcolorbox}

\vspace{-8pt}

\begin{tcolorbox}[
  enhanced,
  colback=white,
  colframe=modDIS!60,
  boxrule=0.8pt,
  arc=2mm,
  left=6pt, right=6pt, top=4pt, bottom=3pt,
  title={\textcolor{modDIS}{\textbf{M2\;\;$\vert$\;\;Expert-Curated Disambiguation Rules}}},
  fonttitle=\small\bfseries,
  colbacktitle=modDIS!10,
  coltitle=modDIS,
]
\footnotesize\ttfamily
\textcolor{modDIS}{<disambiguation\_rules>}\\
\quad \textcolor{slotGray}{\textit{[Boundary condition 1: decision criteria for confusable label pair A vs.\ B]}}\\
\quad \textcolor{slotGray}{\textit{[Boundary condition 2: positional, specificity, or semantic cues]}}\\
\quad \textcolor{slotGray}{\quad\dots}\\
\quad \textcolor{slotGray}{\textit{[Boundary condition K: role-dependent or context-dependent distinctions]}}\\
\textcolor{modDIS}{</disambiguation\_rules>}\\[2pt]
\rmfamily\textcolor{slotGray}{\textit{Operationalizes the annotation manual inside the prompt; resolves ontology boundary errors.}}
\end{tcolorbox}

\vspace{-8pt}

\begin{tcolorbox}[
  enhanced,
  colback=white,
  colframe=modCOT!60,
  boxrule=0.8pt,
  arc=2mm,
  left=6pt, right=6pt, top=4pt, bottom=3pt,
  title={\textcolor{modCOT}{\textbf{M3\;\;$\vert$\;\;Chain-of-Thought Reasoning Scaffold}}},
  fonttitle=\small\bfseries,
  colbacktitle=modCOT!10,
  coltitle=modCOT,
]
\footnotesize\ttfamily
\textcolor{modCOT}{<reasoning\_process>}\\
\quad \textcolor{slotGray}{\textit{Step 1: Context and metadata analysis}}\\
\quad \textcolor{slotGray}{\textit{Step 2: Phrase decomposition and label matching}}\\
\quad \textcolor{slotGray}{\textit{Step 3: Evidence extraction with boundary verification}}\\
\quad \textcolor{slotGray}{\textit{Step 4: Cross-validation with loop-back conditions}}\\
\textcolor{modCOT}{</reasoning\_process>}\\[2pt]
\rmfamily\textcolor{slotGray}{\textit{Imposes a disciplined internal verification sequence before structured output emission.}}
\end{tcolorbox}

\vspace{-8pt}

\begin{tcolorbox}[
  enhanced,
  colback=white,
  colframe=modLOG!60,
  boxrule=0.8pt,
  arc=2mm,
  left=6pt, right=6pt, top=4pt, bottom=3pt,
  title={\textcolor{modLOG}{\textbf{M4\;\;$\vert$\;\;Metadata-Aware Decision Logic}}},
  fonttitle=\small\bfseries,
  colbacktitle=modLOG!10,
  coltitle=modLOG,
]
\footnotesize\ttfamily
\textcolor{modLOG}{<metadata\_control>}\\
\quad \textcolor{slotGray}{\textit{[Expose task metadata as explicit control variable (e.g., speaker role, document type)]}}\\
\textcolor{modLOG}{</metadata\_control>}\\[2pt]
\rmfamily\textcolor{slotGray}{\textit{Converts metadata values, which are latent contextual attributes, into observed control signals; narrows the label search space.}}
\end{tcolorbox}

\vspace{-8pt}

\begin{tcolorbox}[
  enhanced,
  colback=white,
  colframe=modSCH!60,
  boxrule=0.8pt,
  arc=2mm,
  left=6pt, right=6pt, top=4pt, bottom=3pt,
  title={\textcolor{modSCH}{\textbf{M5\;\;$\vert$\;\;Structured Output Schema Contract}}},
  fonttitle=\small\bfseries,
  colbacktitle=modSCH!10,
  coltitle=modSCH,
]
\footnotesize\ttfamily
\textcolor{modSCH}{<output\_format>}\\
\quad \textcolor{slotGray}{\textit{[Machine-parseable hard constraints: schema (e.g., JSON array of verbatim Spans, valid label pairings, and multi-label coverage)]}}\\
\textcolor{modSCH}{</output\_format>}\\[2pt]
\rmfamily\textcolor{slotGray}{\textit{Shifts the task from open-ended generation to structured generation with explicit validity conditions.}}
\end{tcolorbox}

\vspace{-8pt}

\begin{tcolorbox}[
  enhanced,
  colback=white,
  colframe=modVAL!60,
  boxrule=0.8pt,
  arc=2mm,
  left=6pt, right=6pt, top=4pt, bottom=3pt,
  title={\textcolor{modVAL}{\textbf{M6\;\;$\vert$\;\;Self-Validation Quality Gate}}},
  fonttitle=\small\bfseries,
  colbacktitle=modVAL!10,
  coltitle=modVAL,
]
\footnotesize\ttfamily
\textcolor{modVAL}{<quality\_gate>}\\
\quad \textcolor{slotGray}{\textit{[1. Output is parseable under the declared schema]}}\\
\quad \textcolor{slotGray}{\textit{[2. All sub-labels are valid for their parent labels]}}\\
\quad \textcolor{slotGray}{\quad\dots}\\
\quad \textcolor{slotGray}{\textit{[K. High-confidence output; all disambiguation rules applied]}}\\
\textcolor{modVAL}{</quality\_gate>}\\[2pt]
\rmfamily\textcolor{slotGray}{\textit{Single-turn intra-generation audit; constrains the model to verify concrete failure conditions within one pass.}}
\end{tcolorbox}
\vspace{-8pt}
\end{tcolorbox}
% \refstepcounter{figure}
\caption{General modular prompt template for ontology-constrained structured prediction and evidence extraction.}

\label{fig:prompt_template}

\end{figure*}
 
\clearpage

On the EPPC Miner benchmark, this modular architecture yields
substantial improvements in zero-shot structured annotation
(Table~\ref{tab:llama_allshot_baseline_vs_prompt}). 

\subsection{In-Context Prompting Benchmark for EPPC Miner}

We establish an in context prompting benchmark for EPPC Miner to characterize task difficulty before supervised adaptation. Given a patient to provider secure message $s$ and direction indicator $d$, models are prompted to generate one or more structured $(\text{Code}, \text{Sub-code}, \text{Span})$ tuples under strict ontology constraints. We evaluate this benchmark in zero-shot, one-shot, and two-shot settings (Table~\ref{tab:llama_allshot_baseline_vs_prompt}), providing a controlled reference point for understanding which aspects of the task are learnable from prompting alone and whether these gains persist after supervised adaptation (Table~\ref{tab:sft_instruction_comparison}).

To isolate label surface effects, we additionally study label encoding sensitivity. Because autoregressive LLMs generate labels token by token, the surface form of a label directly affects the model's conditional generation probabilities. We therefore evaluate four semantically equivalent encodings: T1 (\emph{Original}), T2 (\emph{Token-Formatted}), T3 (\emph{Numeric IDs}), and T4 (\emph{Simplified Aliases}). These encodings preserve the same ontology while varying semantic transparency and tokenization granularity \citep{wang2023label,gao2023benefits}. A key result is that, after supervised adaptation, the large zero-shot differences across encodings largely collapse: under plain SFT, performance becomes tightly clustered across T1, T2, T3, and T4, and under SFT with token weighting the same qualitative pattern holds, with only modest model dependent shifts and no encoding dominating uniformly across models or metrics. This indicates that label verbalization is not the primary driver of the gains in the main benchmark; instead, the dominant factors are stronger task specification through prompt engineered instruction and supervised adaptation. Full encoding definitions, results, and discussion are provided in Appendix~\ref{sec:label_encoding}.

\subsection{Supervised Fine-Tuning (SFT)}
\label{sec:sft}

We use supervised fine-tuning (SFT) as the primary task adaptation step after in-context prompting. In EPPC Miner, each training example consists of a target sentence $s$, its local discourse context $C = (s_{\text{previous}}, s_{\text{next}})$, a direction indicator $d \in {\texttt{Y},\texttt{N}}$, and a gold structured output set $\mathcal{Y}$ containing one or more grounded $(\text{Code}, \text{Sub-code}, \text{Span})$ triples. For training, we linearize $\mathcal{Y}$ into a JSON completion string $y$. The model input is formed by concatenating task instructions with instance content:
\begin{equation*}
x \;=\; \texttt{Task\ Instructions} \,\Vert\, \texttt{\textbackslash n} \,\Vert\, s \,\Vert\, \texttt{\textbackslash n} \,\Vert\, d ,
\end{equation*}
where $\Vert$ denotes concatenation and $\texttt{\textbackslash n}$ denotes a newline token. 
The causal-LM sequence is then
\begin{equation*}
\mathbf{z} \;=\; \mathrm{Tok}\!\left(x \,\Vert\, \texttt{\textbackslash n} \,\Vert\, y\right) \;=\; [z_1,\ldots,z_T],
\end{equation*}

where \(\mathrm{Tok}(\cdot)\) denotes the model tokenizer. Following completion-only instruction tuning, we compute loss only on completion tokens and mask prompt tokens from gradient updates. Let \(\mathbf{r}\in\{0,1\}^{T}\) be a token indicator with \(r_j=1\) if token \(j\) belongs to the completion. We optimize the masked negative log-likelihood:
\begin{equation*}
\mathcal{L}_{\mathrm{SFT}}(\theta)
=
-\mathbb{E}_{(x,\mathcal{Y})\sim \mathcal{D}}
\left[
\frac{\sum_{j=1}^{T} r_j \log \pi_{\theta}(z_j \mid z_{<j})}{\sum_{j=1}^{T} r_j}
\right].
\end{equation*}
Here, $\pi_{\theta}$ denotes the language model parameterized by $\theta$. 
This objective prevents the model from learning to copy instruction text and concentrates optimization on schema-conformant structured generation. We fine-tune with parameter-efficient adapters using QLoRA. Low-rank updates are applied to attention projection modules while base model parameters remain frozen.

\subsection{STaR-DRO: Stateful Tsallis Reweighting for Robust Structured Completion Tuning}
\label{sec:stardro}
STaR-DRO extends group-robust fine-tuning to structured completion tasks. We begin by reviewing conventional group DRO and the limitations that motivate our method.
% ---- 3.4.1: DRO Motivation and Group DRO Foundation ----

\subsubsection{From Conventional Group DRO to Stateful Sparse Tsallis Reweighting}
\label{sec:dro_background}

Standard supervised fine-tuning minimizes the average loss across all training examples. For structured prediction tasks with heterogeneous group difficulty, this empirical risk minimization (ERM) objective can mask systematic failures on persistently hard groups: a model that achieves low average loss may still incur high loss on semantically ambiguous, structurally complex, or otherwise hard-to-learn regions of the label space~\citep{sagawa2020distributionally,duchi2021learning}. Distributionally robust optimization (DRO) addresses this failure mode by instead minimizing the worst-case expected loss over a family of distributions~\citep{namkoong2016stochastic,duchi2021learning}.

To motivate our method, consider the standard group DRO  formulation of~\citet{sagawa2020distributionally}. Suppose the training data can be partitioned into $G$ predefined groups $\mathcal{G}=\{1,\dots,G\}$, each with population risk $\mathcal{R}_g(\theta)=\mathbb{E}_{\xi\sim\mathcal{P}_g}[\ell(\theta;\xi)]$. Group DRO seeks parameters that minimize the maximum group risk:
\begin{equation}
\min_{\theta\in\Theta}\;\max_{g\in\mathcal{G}}\;\mathcal{R}_g(\theta).
\label{eq:gdro_obj}
\end{equation}
Equivalently, this objective can be written as a minimax problem over the probability simplex $\Delta_G=\{q\in\mathbb{R}_{\ge 0}^G:\sum_{g}q_g=1\}$:
\begin{equation}
\min_{\theta\in\Theta}\;\max_{q\in\Delta_G}\;\sum_{g=1}^{G}q_g\,\mathcal{R}_g(\theta).
\label{eq:gdro_minimax}
\end{equation}
Here, $q$ acts as an adversarial distribution that concentrates mass on the worst-performing groups, while the model parameters $\theta$ are optimized against this worst case. The inner maximization over $q$ recovers the maximum group risk, so Eqs.~\eqref{eq:gdro_obj} and~\eqref{eq:gdro_minimax} are equivalent.

In the stochastic setting, the population risks $\mathcal{R}_g(\theta)$ are replaced by minibatch estimates $\widehat{L}_{t,g}$, and the adversarial distribution is updated online. \citet{sagawa2020distributionally} solve the inner problem via exponentiated gradient ascent on $q$, which corresponds to mirror ascent with the negative Shannon entropy as the mirror map. At each step~$t$, their update takes the form:
\begin{equation}
q_{t+1,g}\propto q_{t,g}\exp\!\bigl(\eta\,\widehat{L}_{t,g}\bigr),\quad g=1,\dots,G,
\label{eq:exp_grad}
\end{equation}
followed by renormalization to the simplex.

While group DRO provides a principled framework, applying Eq.~\eqref{eq:exp_grad} to structured prediction with LLMs introduces four practical challenges that motivate the design of STaR-DRO:

\begin{enumerate}[leftmargin=*,label=(\roman*)]
\item \textbf{Dense reweighting.} The exponentiated gradient update in Eq.~\eqref{eq:exp_grad} produces a dense distribution: every group receives strictly positive mass at every step, regardless of whether it is persistently difficult or already well-learned. When only a subset of groups is hard, dense updates dilute optimization pressure across all groups, reducing the effective gradient signal on the groups that need it most.

\item \textbf{Minibatch variance.} In structured prediction with fine-grained groups, many groups may be absent from a given minibatch or represented by very few examples. Driving the adversarial update directly from raw minibatch losses produces high-variance group-loss estimates, leading to erratic oscillations in $q$ that destabilize training.

\item \textbf{Unbounded optimizer-facing weights.} In practical robust-training pipelines, adversarial group weights may be translated into sample- or token-level reweighting factors. Under severe group skew, raw importance-ratio-style weights, as in ~\citet{sagawa2020distributionally}, can become highly variable, causing unstable gradients and brittle optimization.

\item \textbf{Worst-case simplex competition.} Because the adversarial weights lie on the simplex, increasing mass on hard groups necessarily decreases mass on other groups. As a result, standard worst-case reweighting does not merely emphasize difficult groups; it also relatively downweights easier groups. For structured prediction, this can induce an unnecessarily sharp robustness--performance trade-off.
\end{enumerate}

STaR-DRO addresses each of these challenges within a unified mirror-ascent framework, as we describe next.

% ---- 3.4.2: The STaR-DRO Algorithm (Task-Agnostic) ----

\subsubsection{STaR-DRO Algorithm}
\label{sec:stardro_algorithm}

We present STaR-DRO as a general-purpose robust reweighting method for any supervised learning task whose training data admits a meaningful group structure. The method is parameterized by four hyperparameters: a Tsallis index $\alpha>1$, a mirror ascent step size $\eta_t>0$, an EMA smoothing coefficient $\rho_t\in(0,1]$, and multiplier shaping parameters $(U,\gamma)$. Algorithm~\ref{alg:stardro} summarizes the complete procedure; we derive each component below.

\paragraph{Setup.}
Let $\mathcal{G}=\{1,\dots,G\}$ be a finite group inventory induced by the task's label structure, and let $\Delta_G$ denote the $G$-simplex. Each training example $i$ in a minibatch $\mathcal{B}_t$ has an associated group membership set $S_i\subseteq\mathcal{G}$ with cardinality $\nu_i=|S_i|$, and an update signal $u_i(\theta)$ derived from the task-specific loss (Section~\ref{sec:stardro_eppc} instantiates these for EPPC Miner). The adversarial distribution $q_t\in\Delta_G$ is initialized uniformly: $q_0=\mathbf{1}/G$.

\paragraph{Step 1: Minibatch group-loss estimation with overlap correction.}
When a single example belongs to multiple groups ($\nu_i>1$), naively accumulating its loss into every group inflates the estimated difficulty of multi-membership examples. STaR-DRO corrects for this overlap by weighting each example's contribution to group $g$ by $1/\nu_i$:
\begin{equation}
\widehat{L}_{t,g}
=
\frac{
  \displaystyle\sum_{i\in\mathcal{B}_t}
  \mathbf{1}[g\in S_i]\;\frac{1}{\nu_i}\;u_i(\theta)
}{
  \displaystyle\sum_{i\in\mathcal{B}_t}
  \mathbf{1}[g\in S_i]\;\frac{1}{\nu_i}
},
\qquad g\in\mathcal{G}.
\label{eq:group_loss_est}
\end{equation}

\paragraph{Step 2: Momentum-smoothed group losses.}
Rather than driving the adversarial update with noisy minibatch estimates, STaR-DRO maintains an exponential moving average (EMA) of group losses. Let $\mathcal{P}_t\subseteq\mathcal{G}$ denote the set of groups observed in minibatch $\mathcal{B}_t$. The smoothed group loss is:
\begin{equation}
L_{t,g}
=
\begin{cases}
\widehat{L}_{t,g}, & \text{if $g$ is observed for the first time at step $t$},\\[4pt]
(1-\rho_t)\,L_{t-1,g}+\rho_t\,\widehat{L}_{t,g}, & \text{if } g\in\mathcal{P}_t \text{ and previously observed},\\[4pt]
L_{t-1,g}, & \text{if } g\notin\mathcal{P}_t.
\end{cases}
\label{eq:ema_loss}
\end{equation}
The coefficient $\rho_t\in(0,1]$ controls the memory horizon: small $\rho_t$ yields long-horizon estimates; large $\rho_t$ increases responsiveness to recent observations. This smoothing addresses challenge~(ii): the mirror-ascent step receives a lower-variance, stateful estimate of group difficulty rather than a single-batch snapshot. 

\paragraph{Step 3: Rescaled ascent signal.}
Because the EMA-smoothed losses $L_{t,g}$ retain information from earlier training stages, when losses are typically larger, their absolute magnitudes may cause earlier high-loss phases to exert disproportionate influence on the current adversarial update relative to the present loss landscape. STaR-DRO normalizes the smoothed losses of present groups by their weighted mean:
\begin{equation*}
s_t
=
\sum_{g\in\mathcal{P}_t}\pi_t(g)\,L_{t,g},
\label{eq:scale_factor}
\end{equation*}
where $\pi_t(g)=n_{t,g}/\!\sum_{h\in\mathcal{P}_t}n_{t,h}$ is the normalized count weight for group $g$, and $n_{t,g}$ is the number of examples contributing to group $g$ in $\mathcal{B}_t$. The rescaled ascent signal is then:
\begin{equation}
a_{t,g}
=
\begin{cases}
L_{t,g}\,/\,s_t, & g\in\mathcal{P}_t,\\[4pt]
0, & g\notin\mathcal{P}_t.
\end{cases}
\label{eq:ascent_signal}
\end{equation}
By construction, $\sum_{g\in\mathcal{P}_t}\pi_t(g)\,a_{t,g}=1$, so the mirror step operates on \emph{relative} group difficulty: a group with $a_{t,g}>1$ is harder than the batch-weighted average, while $a_{t,g}<1$ is relatively easier. As a result, the adversarial update is governed by the current hardness landscape among present groups rather than by absolute loss magnitudes inherited from earlier training stages.

\paragraph{Step 4: Tsallis mirror ascent.}
Standard group DRO uses the negative Shannon entropy (equivalently, the entropic regularizer) as the mirror map, yielding the dense exponentiated-gradient update in Eq.~\eqref{eq:exp_grad}. To address challenge~(i) and obtain \emph{sparse} adversarial distributions, STaR-DRO replaces this with a Tsallis-type mirror map of order $\alpha>1$:
\begin{equation}
\Psi_\alpha(q)
=
\frac{1}{\alpha(\alpha-1)}
\sum_{g=1}^{G}q_g^\alpha
+I_{\Delta_G}(q),
\label{eq:tsallis_mirror}
\end{equation}
where $I_{\Delta_G}$ is the indicator function of the simplex~\citep{zimmert2021tsallis,martins2022sparse}. As $\alpha\to 1^+$, $\Psi_\alpha$ approaches the entropic regularizer associated with the standard dense update. For $\alpha>1$, the corresponding mirror step yields a thresholded power-law map back to the simplex, which permits exact zeros and therefore sparse adversarial distributions.

The mirror-ascent step maximizes a linearized reward penalized by Bregman divergence:
\begin{equation}
q_{t+1}
=
\operatorname*{arg\,max}_{q\in\Delta_G}
\bigl\{
\eta_t\langle a_t,\,q\rangle - D_{\Psi_\alpha}(q\,\|\,q_t)
\bigr\}.
\label{eq:mirror_step}
\end{equation}
To  obtain a closed-form update, we express the mirror-ascent step in the dual coordinate system. The gradient of $\Psi_\alpha$ maps the primal weights to dual coordinates:
\begin{equation}
\xi_t = \nabla\Psi_\alpha(q_t) = \frac{1}{\alpha-1}\bigl(q_{t,1}^{\alpha-1},\dots,q_{t,G}^{\alpha-1}\bigr).
\label{eq:primal_to_dual}
\end{equation}
Under this dual parameterization, the mirror-ascent step takes the form of an additive update:
\begin{equation}
\xi_{t+\frac{1}{2}} = \xi_t + \eta_t\,a_t.
\label{eq:dual_additive}
\end{equation}
The resulting map back to the primal simplex differs qualitatively from the Shannon case: under the Shannon geometry it is the softmax, which is everywhere positive and therefore dense, whereas under the Tsallis geometry with $\alpha>1$ it becomes a thresholded power-law transformation that can assign exact zeros, as we show next.

Defining the scaled dual variable $\mathbf{u}_t:=(\alpha-1)\xi_t=q_t^{\alpha-1}$ (component wise), the update becomes:
\begin{equation}
\mathbf{u}_{t+\frac{1}{2},g}
=
q_{t,g}^{\alpha-1} + (\alpha-1)\,\eta_t\,a_{t,g}.
\label{eq:scaled_dual}
\end{equation}

\paragraph{Step 5: Entmax-style dual-to-primal projection.}
The map from the updated dual coordinate $\mathbf{u}_{t+1/2}$ back to the primal simplex is obtained from the convex conjugate of $\Psi_\alpha$. The KKT conditions yield a thresholded power-law normalization:
\begin{equation}
q_{t+1,g}
=
\bigl[\mathbf{u}_{t+\frac{1}{2},g}-\lambda_t\bigr]_+^{1/(\alpha-1)},
\qquad
\sum_{g=1}^{G}q_{t+1,g}=1,
\label{eq:entmax_proj}
\end{equation}
where $[x]_+=\max(x,0)$ and $\lambda_t$ is the unique threshold satisfying the simplex constraint, found by one-dimensional bisection.

Eq.~\eqref{eq:entmax_proj} is the defining operation of the \emph{entmax} family~\citep{peters2019sparse,blondel2020learning}: coordinates whose dual value $\mathbf{u}_{t+1/2,g}$ falls below the threshold $\lambda_t$ are mapped to \emph{exactly zero}. This is the mechanism through which STaR-DRO achieves sparse adversarial distributions. Groups that are not persistently difficult receive zero adversarial mass and are effectively deactivated from the robust reweighting, while hard groups receive concentrated emphasis. The degree of sparsity is controlled by $\alpha$: larger $\alpha$ lowers the geometric barrier against zero entries and increases the number of groups that are thresholded out. 

\paragraph{Step 6: From adversarial weights to bounded training multipliers.}
Directly using the raw simplex weights $q_t$ as optimizer-facing loss multipliers would reintroduce the instability described in challenge~(iii): as in standard group DRO, the resulting importance-ratio-style training weights can become highly variable and lead to brittle optimization. STaR-DRO instead transforms $q_t$ into bounded, interpretable training multipliers through a three-stage mapping.

First, compute the \emph{density ratio} relative to the uniform baseline:
\begin{equation}
r_{t,g} = \frac{q_{t,g}}{1/G} = G\,q_{t,g}.
\label{eq:density_ratio}
\end{equation}
Under uniform weighting, $r_{t,g}=1$; under full concentration on a single group, $r_{t,g}=G$.
Second, define the \emph{normalized hardness excess}:
\begin{equation}
e_{t,g}
=
\Bigl[\frac{r_{t,g}-1}{G-1}\Bigr]_{[0,1]},
\label{eq:hardness_excess}
\end{equation}
where $[x]_{[0,1]}=\min(1,\max(0,x))$. This extracts only the \emph{excess} adversarial mass above the uniform baseline: when $q_{t,g}=1/G$, we have $r_{t,g}=1$ and hence $e_{t,g}=0$, while $e_{t,g}>0$ only for groups whose simplex weight exceeds uniform. In this way, the map discards relative decreases below uniform and retains only evidence that a group is currently harder than the neutral baseline.

Third, apply a power-law shaping to obtain the final per-group multiplier:
\begin{equation}
m_{t,g}
=
\begin{cases}
1 + (U-1)\,e_{t,g}^{\,\gamma}, & g\in\mathcal{P}_t,\\[4pt]
1, & g\notin\mathcal{P}_t,
\end{cases}
\label{eq:multiplier}
\end{equation}

\begin{algorithm}[t]
\caption{STaR-DRO: Stateful Tsallis Reweighting for Distributionally Robust Optimization}
\label{alg:stardro}
\small
\begin{algorithmic}[1]
\Require Groups $\mathcal{G}=\{1,\dots,G\}$; Tsallis order $\alpha>1$; step size $\eta_t$; EMA coefficient $\rho_t$; multiplier ceiling $U>1$; curvature $\gamma>0$
\State $q_{1,g}\gets 1/G$ and $L_{0,g}\gets \textsc{None}$ for all $g\in\mathcal{G}$

\For{$t=1,2,\dots$}
  \State Draw minibatch $\mathcal{B}_t$ and set $\mathcal{P}_t\gets\{g\in\mathcal{G}:\exists\,i\in\mathcal{B}_t,\ g\in S_i\}$ \Comment{present groups}

  \State $\displaystyle
    \widehat{L}_{t,g}\gets
    \frac{
      \sum_{i\in\mathcal{B}_t}\mathbf{1}[g\in S_i]\,\nu_i^{-1}\,u_i(\theta_t)
    }{
      \sum_{i\in\mathcal{B}_t}\mathbf{1}[g\in S_i]\,\nu_i^{-1}
    }
    \quad \forall\, g\in\mathcal{P}_t$
    \Comment{overlap-corrected group losses}

  \State $\displaystyle
    L_{t,g}\gets
    \begin{cases}
      \widehat{L}_{t,g}, & g\in\mathcal{P}_t,\ L_{t-1,g}=\textsc{None},\\
      (1-\rho_t)L_{t-1,g}+\rho_t\,\widehat{L}_{t,g}, & g\in\mathcal{P}_t,\ L_{t-1,g}\neq\textsc{None},\\
      L_{t-1,g}, & g\notin\mathcal{P}_t
    \end{cases}
    \quad \forall\, g\in\mathcal{G}$
    \Comment{EMA-smoothed group difficulty}

  \State $\displaystyle
    s_t\gets \sum_{g\in\mathcal{P}_t}
    \frac{n_{t,g}}{\sum_{h\in\mathcal{P}_t}n_{t,h}}\,L_{t,g}$
    \Comment{normalized scale}

  \State $a_{t,g}\gets L_{t,g}/s_t$ for $g\in\mathcal{P}_t$, and $a_{t,g}\gets 0$ otherwise \Comment{relative hardness signal}

  \State $u_{t+\frac{1}{2},g}\gets q_{t,g}^{\,\alpha-1}+(\alpha-1)\eta_t a_{t,g}\quad \forall\, g\in\mathcal{G}$ \Comment{scaled dual ascent}

  \State Find $\lambda_t$ such that $\displaystyle\sum_{g=1}^{G}[u_{t+\frac{1}{2},g}-\lambda_t]_+^{1/(\alpha-1)}=1$ \Comment{bisection}

  \State $q_{t+1,g}\gets [u_{t+\frac{1}{2},g}-\lambda_t]_+^{1/(\alpha-1)}\quad \forall\, g\in\mathcal{G}$ \Comment{entmax projection}

  \State $m_{t,g}\gets 1+(U-1)\Bigl[\dfrac{Gq_{t+1,g}-1}{G-1}\Bigr]_{[0,1]}^{\gamma}$ for $g\in\mathcal{P}_t$, and $m_{t,g}\gets 1$ otherwise \Comment{bounded group multipliers}

  \State $m_{t,i}\gets \nu_i^{-1}\sum_{g\in S_i}m_{t,g}\quad \forall\, i\in\mathcal{B}_t$ \Comment{mean aggregation}

  \State $\displaystyle
    \mathcal{L}^{\mathrm{STaR}}_t(\theta)\gets
    \frac{
      \sum_{i\in\mathcal{B}_t} \,m_{t,i}\,\bar{\ell}_i(\theta)
    }{
      \sum_{i\in\mathcal{B}_t} \,m_{t,i}
    }$
    \Comment{robust objective}

  \State Update $\theta$ with one optimizer step on $\mathcal{L}^{\mathrm{STaR}}_t(\theta)$
\EndFor
\end{algorithmic}
\end{algorithm}

where $U>1$ is the multiplier ceiling and $\gamma>0$ controls curvature. The multiplier is bounded in $[1,U]$ by construction: $m_{t,g}=1$ when $q_{t,g}\le 1/G$ and $m_{t,g}=U$ when $q_{t,g}=1$. This directly addresses challenge~(iv): rather than translating worst-case simplex competition into both upweighting of hard groups and downweighting of easier ones, STaR-DRO converts only excess mass above uniform into additional emphasis, while keeping all non-hard groups at the neutral weight $1$. As a result, the method prioritizes hard groups without unnecessarily suppressing easier groups that still provide useful supervision for overall structured prediction performance. The curvature parameter $\gamma$ governs how aggressively intermediate hardness levels are upweighted: $\gamma<1$ amplifies moderate-hardness groups, whereas $\gamma>1$ concentrates emphasis more sharply on the hardest groups.

\paragraph{Step 7: Multi-group aggregation.}
When an example belongs to multiple groups ($\nu_i>1$), its per-group multipliers must be aggregated into a single training weight. STaR-DRO computes the example-level multiplier as the arithmetic mean over its group memberships:
\begin{equation*}
m_{t,i}
=
\frac{1}{\nu_i}\sum_{g\in S_i}m_{t,g}.
\label{eq:agg_mean}
\end{equation*}
The mean aggregation ensures that an example's robust weight reflects its average difficulty across all groups it participates in, rather than being dominated by a single hard group. This is particularly important for structured prediction, where a single example may contain annotations Spanning both easy and hard groups.

The robust training objective then takes the form:
\begin{equation*}
\mathcal{L}_{\mathrm{STaR}}(\theta)
=
\frac{
  \sum_{i\in\mathcal{B}_t} \,m_{t,i}\,\bar{\ell}_i(\theta)
}{
  \sum_{i\in\mathcal{B}_t} \,m_{t,i}
},
\label{eq:stardro_obj}
\end{equation*}
where $\bar{\ell}_i(\theta)$ is the per-example loss.

\paragraph{Summary of design principles.}
STaR-DRO integrates four design choices that directly address the limitations of standard group DRO for structured prediction:
\begin{enumerate}[leftmargin=*,label=(\roman*)]

\item \textbf{Sparse adversarial redistribution.} The Tsallis mirror map with entmax-style projection (Eqs.~\eqref{eq:tsallis_mirror}--\eqref{eq:entmax_proj}) concentrates adversarial mass on persistently hard groups while allowing exactly zero weight on noncompetitive groups. This avoids the dense reweighting induced by exponentiated-gradient updates and focuses optimization pressure where it is most needed.

\item \textbf{Stateful group-difficulty tracking.} EMA-smoothed group losses (Eq.~\eqref{eq:ema_loss}) replace raw minibatch losses with lower-variance, stateful estimates of group difficulty. The normalized ascent signal $a_t$ (Eq.~\eqref{eq:ascent_signal}) further ensures that the adversarial update is driven by relative current hardness rather than absolute loss magnitude, yielding a more stable signal in fine-grained grouped structured prediction.

\item \textbf{Bounded optimizer-facing reweighting.} The density-ratio $\to$ excess $\to$ power-law multiplier map (Eqs.~\eqref{eq:density_ratio}--\eqref{eq:multiplier}) converts adversarial simplex weights into training multipliers bounded in $[1,U]$. This prevents the unstable gradients and brittle optimization that can arise when raw group weights are translated directly into sample- or token-level loss weights under severe skew.

\item \textbf{Excess-only emphasis on hard groups.} By converting only adversarial mass \emph{above} the uniform baseline into additional weight, while leaving all groups at or below uniform at the neutral multiplier $1$, STaR-DRO avoids translating worst-case simplex competition into explicit downweighting of easier groups. This yields a more favorable robustness--performance trade-off for structured prediction, where non-hard groups still provide useful supervision for overall semantic and structural accuracy.

\end{enumerate}

% ---- 3.4.3: Application to EPPC Miner ----

\subsubsection{Application to EPPC Miner}
\label{sec:stardro_eppc}

We now instantiate the STaR-DRO framework for the EPPC Miner structured extraction task. This requires specifying: (a) how the task's label structure induces groups, (b) what update signal $u_i(\theta)$ is used, and (c) at what granularity robust multipliers are applied.

\paragraph{Structured outputs and group construction.}
For an input instance $x_i=(s_i,C_i,d_i)$, the gold structured output is a set of grounded annotation triples:
\begin{equation*}
\mathcal{Y}_i = \{(\kappa_{ia},\varsigma_{ia},\pi_{ia})\}_{a=1}^{A_i},
\end{equation*}
where $\kappa_{ia}\in L_{\mathrm{Code}}$ is a communication Code, $\varsigma_{ia}\in\mathcal{H}(\kappa_{ia})$ is a valid Sub-code, $\pi_{ia}$ is the grounded evidence Span, and $A_i$ is the number of annotations. STaR-DRO defines the group inventory $\mathcal{G}$ from these structured labels. We consider five progressively finer grouping strategies:

\begin{enumerate}[leftmargin=*,label=(\alph*)]
\item \textbf{Code.} $\mathcal{G}=L_{\mathrm{Code}}$. Groups correspond to the nine top-level communication codes. This is the coarsest semantic grouping.
\item \textbf{Sub-code.} $\mathcal{G}=L_{\mathrm{Sub-code}}$. Groups correspond to the 38 individual Sub-codes.
\item \textbf{Code$\times$Sub-code.} $\mathcal{G}=\{(\kappa,\varsigma):\kappa\in\mathcal{L}_{\mathrm{Code}},\;\varsigma\in\mathcal{H}(\kappa)\}$. Groups are joint Code--Sub-code pairs.
\item \textbf{Number of Annotations.} $\mathcal{G}=\{\mathrm{NA}_k:k\in\{1,2,\dots\}\}$, where $\mathrm{NA}_k$ denotes the bucket of examples with exactly $k$ annotations. This captures output-complexity difficulty.
\item \textbf{Code$\times$Sub-code$\times$Number of Annotations.} $\mathcal{G}=\{(\kappa,\varsigma,\mathrm{NA}_k)\}$. The finest grouping, crossing semantic identity with structural complexity.
\end{enumerate}

Under groupings (a)--(c), a single example with $A_i>1$ annotations may belong to multiple groups (one per annotation); hence $S_i\subseteq\mathcal{G}$ with $|S_i|\ge 1$, and the overlap correction in Eq.~\eqref{eq:group_loss_est} applies. Under grouping~(d), all annotations in an example share the same group $\mathrm{NA}_{A_i}$, so $|S_i|=1$. Grouping~(e) combines both properties.

\paragraph{Update signal: sample-level vs.\ annotation-level.}
The update signal $u_i(\theta)$ that drives the group-loss estimates can be computed either at the sample level or at the annotation level. Using the completion-only masking introduced in Subsection~\ref{sec:sft}, we first define the masked token loss for example \(i\) at position \(j\) as
\[
\ell_{i,j}(\theta)
=
-r_{i,j}\log \pi_\theta(z_{i,j}\mid z_{i,<j}),
\]
where \(r_{i,j}=1\) for completion tokens and \(r_{i,j}=0\) for prompt tokens.

In \emph{sample-level} mode, the signal is the average completion loss over all completion tokens of example \(i\):
\[
u_i(\theta)
=
\frac{\sum_j \ell_{i,j}(\theta)}
{\sum_j r_{i,j}}.
\]

In \emph{annotation-level} mode, each annotation \(a\) of example \(i\) is associated with the token set \(\mathcal{T}_{i,a}\) covering its JSON object in the linearized completion. The per-annotation signal is the length-normalized loss over that annotation:
\[
u_{i,a}(\theta)
=
\frac{1}{|\mathcal{T}_{i,a}|}
\sum_{j\in \mathcal{T}_{i,a}} \ell_{i,j}(\theta).
\]

Each annotation is assigned to exactly one group \(g_{i,a}\) under the chosen grouping scheme, and the group-loss estimate is computed over annotation-level pairs \((i,a)\) rather than whole examples. Annotation-level mode enables the robust mechanism to target the exact structured regions, namely label-bearing tokens and local evidence spans, that realize hard Code/Sub-code/Span combinations, rather than reweighting the full completion uniformly.

\paragraph{Illustrative example: group-loss attribution.}
Appendix~\ref{app:eppc_group_attribution} provides a concrete example of how the same
multi-annotation EPPC sample is assigned to groups under the grouping strategies used
by STaR-DRO. Under Code-, Sub-code-, and Code$\times$Sub-code-based grouping, a
sample with multiple annotations can contribute to multiple groups at the sample level,
with overlap correction applied as in Eq.~\ref{eq:group_loss_est}. In annotation-level attribution, by
contrast, each annotation is assigned to its corresponding group, allowing the robust
signal to be localized to the specific structured-output tokens associated with that
annotation. Under Number-of-Annotations $\mathrm{NA}_k$ grouping, the entire sample and
all of its annotations share the same annotation-count group. The appendix therefore
clarifies the distinction between sample-level multi-group attribution and
annotation-level group assignment, which motivates the annotation-level robust
objective introduced below.

\paragraph{Annotation-level robust objective.}
When annotation-level robustness is used, STaR-DRO maps the per-group multipliers back onto individual tokens through the annotation structure. If token \(j\) in example \(i\) belongs to annotation \(a\), and that annotation is assigned to group \(g_{i,a}\), its robust weight is
\[
d_{i,j} = m_{g_{i,a}}, \quad j \in \mathcal{T}_{i,a},
\]
and \(d_{i,j}=1\) for completion tokens not covered by any annotation. The annotation-level robust objective is then
\[
\mathcal{L}^{\mathrm{ann}}_{\mathrm{STaR}}(\theta)
=
\frac{
\sum_i \sum_j d_{i,j}\ell_{i,j}(\theta)
}{
\sum_i \sum_j d_{i,j}r_{i,j}
}.
\]
Because each annotated token inherits the multiplier of exactly one annotation group, this formulation directs robust emphasis to the semantically decisive parts of the output, including label tokens and evidence-span content associated with hard groups, while leaving structurally easy completion tokens such as JSON delimiters at the neutral weight.

\subsection{Experimental Settings}
\label{sec:exp_settings}
For reproducibility, the complete experimental setup is summarized in Table~\ref{tab:exp_config}. The table consolidates the model families, parameter-efficient tuning configuration, optimization settings, STaR-DRO hyperparameters, training schedule, inference stack, and hardware used across the main experiments. Unless explicitly noted otherwise in the corresponding sections, all results reported in this paper follow the protocol specified in Table~\ref{tab:exp_config}.

\section{Results}
\label{sec:results}

We evaluate the proposed framework along two dimensions: structured prediction accuracy and group-wise robustness.

\subsection{Evaluation Protocol}
\label{sec:eval-protocol}

We quantify structured prediction accuracy using three complementary metrics that reflect the hierarchical structure of EPPC Miner.

\textbf{Code F1} measures multi-label classification performance at the coarsest semantic level. Precision, recall, and F1 are computed from their set overlap aggregated across the evaluation corpus.

\textbf{Sub-code F1} applies the same multi-label protocol at the finer-grained Sub-code level, where the ontology expands from 9 Codes to 38 Sub-codes and each Sub-code must remain valid under its parent Code. This is the most semantically demanding metric, since it penalizes both incorrect parent--child assignments and confusions among closely related Sub-codes within the same Code family.

\textbf{Span F1} evaluates evidence extraction quality using a relaxed token-level matching criterion. A predicted Span $\hat{\pi}$ is counted as correct if it either fully contains a gold Span $\pi$, is fully contained by it, or attains token-level Jaccard similarity of at least $0.6$. This relaxation tolerates minor boundary variation while still requiring high overlap fidelity. Formal definitions are provided in Appendix~\ref{append:metric}.

 Improvements on Code and Sub-code primarily reflect stronger \emph{semantic} decision-making, whereas improvements on Span reflect better \emph{extraction} behavior.

\subsection{Prompt Engineering Improves Structured Annotation Across Shot Regimes}
\label{sec:results-prompt}
 
Table~\ref{tab:llama_allshot_baseline_vs_prompt} compares the baseline instruction with the prompt-engineered instruction across zero-shot, one-shot, and two-shot evaluation for four Llama instruction-tuned models spanning 1B to 70B parameters. Figures~\ref{fig:qwen25_7b_eppc_results}, \ref{fig:qwen25_14b_eppc_results}, \ref{fig:gemma_2b_eppc_results}, and \ref{fig:gemma_9b_eppc_results} further extend this comparison to the Qwen and Gemma families.

\FloatBarrier
\begin{table*}[!t]
\centering
\caption{Experimental configuration and evaluation protocol.}
\label{tab:exp_config}
\small
\setlength{\tabcolsep}{5pt}
\renewcommand{\arraystretch}{1.12}
\begin{tabularx}{\textwidth}{>{\raggedright\arraybackslash}p{3.6cm} >{\raggedright\arraybackslash}X}
\toprule
\textbf{Component} & \textbf{Configuration} \\
\midrule
Models & Instruction-tuned Llama family (Llama-3.2-1B, Llama-3.2-3B, Llama-3.1-8B, Llama-3.3-70B), Qwen2.5 family (Qwen2.5-7B, Qwen2.5-14B), and Gemma-2 family (Gemma-2-2B, Gemma-2-9B). \\
\midrule
Parameter-efficient tuning & LoRA with rank $r=16$, scaling factor $32$, dropout $0.05$. \\
\midrule
Optimization (default) & bfloat16, gradient checkpointing, AdamW, cosine learning-rate decay, warmup ratio $0.03$, learning rate $2\times10^{-4}$, per-device batch size $3$, gradient accumulation $1$, total epochs $3$. \\
\midrule
Regularization & Weight decay tuned over $\{0.01,\,0.05,\,0.1\}$. \\
\midrule
Quantization strategy & QLoRA-style 4-bit adaptation for 70B only; all other models fine-tuned without 4-bit quantization. \\
\midrule
Compute setup & 70B trained on two GPUs; all other models trained on a single GPU. \\
\midrule
STaR-DRO hyperparameters & Tsallis order $\alpha \in (1.1,\,1.3]$; EMA coefficient $\rho_t$ cosine-annealed from $0.3$ to $0.03$; mirror-ascent step size $\eta \in [6\times10^{-4},\,3\times10^{-3}]$; multiplier ceiling $U \in [10,\,30]$; curvature $\gamma=0.75$. The values of $\eta$ and $U$ were chosen according to the number of groups and the corresponding group cardinalities. \\
\midrule
DRO schedule & Robust reweighting active in epochs 2--3 only; Standard DRO follows the same schedule. \\
\midrule
Token weighting & Used only in label-encoding sensitivity experiments; not part of the default main robustness configuration. \\
\midrule
Training procedure & Single continuous optimization trajectory; robust objectives introduced from epoch 2 as scheduled loss reweighting, without merge-and-restart transitions. \\
\midrule
Inference/evaluation stack & Zero-shot inference with \texttt{lm-eval} and \texttt{vLLM} under task-specific chat templates. \\
\midrule
Primary metric & $F1_{\text{Code}}, F1_{\text{Sub-code}}, F1_{\text{Span}}$. \\
\midrule
Hardware & NVIDIA B200, H200, and A100 Tensor Core GPUs, and NVIDIA RTX PRO 6000 Blackwell Server Edition GPUs. \\
\bottomrule
\end{tabularx}
\end{table*}
\FloatBarrier

\subsubsection{Zero-Shot Results}

\paragraph{Llama Family Results.}

Prompt engineering yields large gains in zero-shot structured annotation. Averaged across the four Llama variants, the prompt-engineered instruction improves Code F1 by $+15.90$, Sub-code F1 by $+13.02$, and Span F1 by $+17.40$ over the baseline instruction. The largest model, \texttt{Llama-3.3-70B-Instruct}, improves from 64.53 to 68.92 Code F1, from 41.80 to 50.34 Sub-code F1, and from 71.02 to 84.70 Span F1. These gains are especially notable for Span extraction, where explicit grounding constraints appear to materially improve evidence localization.

\begin{table*}[!t]
\centering
\caption{Zero-shot, one-shot, and two-shot performance of Llama instruction variants on the EPPC Miner benchmark.}
\label{tab:llama_allshot_baseline_vs_prompt}
\resizebox{\textwidth}{!}{%
\begin{tabular}{lllccccccccc}
\toprule
\multirow{2}{*}{\textbf{Shot}} 
& \multirow{2}{*}{\textbf{Category}} 
& \multirow{2}{*}{\textbf{Model}} 
& \multicolumn{3}{c}{\textbf{Code}} 
& \multicolumn{3}{c}{\textbf{Sub-code}} 
& \multicolumn{3}{c}{\textbf{Span}} \\
\cmidrule(lr){4-6} \cmidrule(lr){7-9} \cmidrule(lr){10-12}
& & & \textbf{Precision} & \textbf{Recall} & \textbf{F1}
& \textbf{Precision} & \textbf{Recall} & \textbf{F1}
& \textbf{Precision} & \textbf{Recall} & \textbf{F1} \\
\midrule

\multirow{8}{*}{Zero-shot}
& \multirow{4}{*}{Baseline Instruction}
 & Llama-3.3-70B-Instruct & 57.87 & 72.92 & 64.53 & 34.70 & 52.57 & 41.80 & 58.63 & 90.03 & 71.02 \\
& & Llama-3.1-8B-Instruct  & 0.00  & 0.00  & 0.00  & 0.00  & 0.00  & 0.00  & 28.19 & 87.45 & 42.65 \\
& & Llama-3.2-3B-Instruct  & 19.92 & 63.78 & 30.36 & 7.76  & 33.22 & 12.57 & 8.82  & 71.52 & 15.70 \\
& & Llama-3.2-1B-Instruct  & 22.50 & 35.69 & 27.60 & 3.51  & 10.07 & 5.21  & 5.60  & 65.97 & 10.31 \\
\cmidrule(lr){2-12}
& \multirow{4}{*}{Prompt-Engineered Instruction}
 & Llama-3.3-70B-Instruct & 69.16 & 68.67 & 68.92 & 50.72 & 49.96 & 50.34 & 84.52 & 84.89 & 84.70 \\
& & Llama-3.1-8B-Instruct  & 53.44 & 52.26 & 52.84 & 37.64 & 38.91 & 38.26 & 58.98 & 74.12 & 65.69 \\
& & Llama-3.2-3B-Instruct  & 38.53 & 40.85 & 39.66 & 15.58 & 19.30 & 17.24 & 40.90 & 65.60 & 50.39 \\
& & Llama-3.2-1B-Instruct  & 20.84 & 30.16 & 24.65 & 4.84  & 7.23  & 5.80  & 4.77  & 39.33 & 8.51 \\
\midrule

\multirow{8}{*}{One-shot}
& \multirow{4}{*}{Baseline Instruction}
 & Llama-3.3-70B-Instruct & 60.66 & 69.08 & 64.60 & 37.60 & 48.06 & 42.19 & 54.98 & 85.35 & 66.88 \\
& & Llama-3.1-8B-Instruct  & 49.11 & 53.22 & 51.08 & 19.59 & 33.25 & 24.66 & 24.04 & 75.10 & 36.42 \\
& & Llama-3.2-3B-Instruct  & 37.25 & 42.65 & 39.77 & 12.19 & 21.85 & 15.65 & 22.30 & 59.70 & 32.47 \\
& & Llama-3.2-1B-Instruct  & 18.58 & 38.01 & 24.96 & 5.48  & 16.79 & 8.26  & 3.70  & 44.40 & 6.84  \\
\cmidrule(lr){2-12}
& \multirow{4}{*}{Prompt-Engineered Instruction}
 & Llama-3.3-70B-Instruct & 67.91 & 71.43 & 69.63 & 47.98 & 51.24 & 49.56 & 79.07 & 85.83 & 82.31 \\
& & Llama-3.1-8B-Instruct  & 52.14 & 58.10 & 54.96 & 35.70 & 42.02 & 38.60 & 48.14 & 76.04 & 58.96 \\
& & Llama-3.2-3B-Instruct  & 44.87 & 41.98 & 43.38 & 16.99 & 15.55 & 16.24 & 51.26 & 54.48 & 52.82 \\
& & Llama-3.2-1B-Instruct  & 28.39 & 34.13 & 30.99 & 7.34  & 7.53  & 7.44  & 10.05 & 45.70 & 16.48 \\
\midrule

\multirow{8}{*}{Two-shot}
& \multirow{4}{*}{Baseline Instruction}
 & Llama-3.3-70B-Instruct & 57.87 & 72.92 & 64.53 & 34.70 & 52.57 & 41.80 & 58.63 & 90.03 & 71.02 \\
& & Llama-3.1-8B-Instruct  & 0.00  & 0.00  & 0.00  & 0.00  & 0.00  & 0.00  & 28.19 & 87.45 & 42.65 \\
& & Llama-3.2-3B-Instruct  & 19.92 & 63.78 & 30.36 & 7.76  & 33.22 & 12.57 & 8.82  & 71.52 & 15.70 \\
& & Llama-3.2-1B-Instruct  & 22.50 & 35.69 & 27.60 & 3.51  & 10.07 & 5.21  & 5.60  & 65.97 & 10.31 \\
\cmidrule(lr){2-12}
& \multirow{4}{*}{Prompt-Engineered Instruction}
 & Llama-3.3-70B-Instruct & 70.13 & 71.80 & 70.96 & 50.15 & 51.20 & 50.67 & 82.45 & 84.96 & 83.68 \\
& & Llama-3.1-8B-Instruct  & 56.52 & 63.91 & 59.99 & 36.57 & 43.07 & 39.55 & 48.22 & 76.69 & 59.21 \\
& & Llama-3.2-3B-Instruct  & 50.19 & 50.04 & 50.12 & 19.95 & 22.11 & 20.97 & 46.49 & 61.47 & 52.94 \\
& & Llama-3.2-1B-Instruct  & 35.90 & 34.84 & 35.36 & 9.23  & 8.10  & 8.62  & 22.91 & 43.57 & 30.03 \\
\bottomrule
\end{tabular}%
}
\end{table*}

The clearest qualitative shift appears for \texttt{Llama-3.1-8B-Instruct}. Under the baseline instruction, zero-shot Code and Sub-code F1 collapse to $0.00/0.00$, despite a non-trivial Span F1 of 42.65. Under prompt engineering, the same model reaches 52.84 Code F1, 38.26 Sub-code F1, and 65.69 Span F1. Under the evaluation protocol, this represents a transition from severe semantic failure to task-usable structured extraction. The prompt-engineered instruction supplies the structural scaffolding missing from the baseline prompt, particularly through XML organization, a stricter schema contract, and the final self-validation gate.

\texttt{Llama-3.2-3B-Instruct} shows a related pattern. Zero-shot performance rises from 30.36/12.57/15.70 to 39.66/17.24/50.39 for Code/Sub-code/Span F1. The disproportionate gain on Span suggests that explicit extraction rules, including verbatim copying constraints and boundary verification, are especially helpful for mid-scale models that can follow instructions but still struggle with unconstrained structured generation.

%%% CHANGE 1: Cross-family prompt-engineering results %%%

\paragraph{Cross-Family Transfer: Results on Qwen and Gemma.}

The prompt-engineering gains are not specific to the Llama family. The same qualitative pattern holds across Qwen and Gemma architectures. The most striking cross-family result is \texttt{gemma-2-2b-it}, which mirrors the Llama-3.1-8B pattern at a different scale: Code F1 rises from 0.3044 to 0.4323, and Span F1 nearly doubles from 0.3483 to 0.6214. This $+$0.2731 Span gain is the largest absolute Span improvement observed across all model families and confirms that explicit extraction constraints are especially impactful for smaller models regardless of architecture. Across the three model families, the pattern is consistent: prompt engineering improves structured annotation most where the baseline instruction is least stable, and the gains concentrate on the same metric dimensions, Span and Sub-code, that the designed modules most directly target.

\subsubsection{Few-Shot Results}

The gains from prompt engineering persist in one-shot and two-shot evaluation. In one-shot evaluation average gains of $+4.64$ Code F1, $+5.27$ Sub-code F1, and $+16.99$ Span F1 are observed. The smaller gain on Code and Sub-code relative to zero-shot is expected: a single demonstration already provides partial task specification. Span gains, however, remain large, indicating that examples alone are not sufficient to reliably enforce exact evidence extraction behavior.

In the two-shot setting prompt engineering again delivers strong gains, especially for smaller models. For \texttt{Llama-3.2-3B-Instruct}, Code F1 improves from 30.36 to 50.12 and Span F1 from 15.70 to 52.94. For \texttt{Llama-3.2-1B-Instruct}, which was adversely affected by the denser prompt in zero-shot, two-shot prompting now improves all three metrics, including Code F1 from 27.60 to 35.36 and Span F1 from 10.31 to 30.03. This reversal suggests that demonstrations help smaller models interpret and execute the more structured instruction template.

\begin{figure*}[!t]
    \centering
    \includegraphics[width=\textwidth]{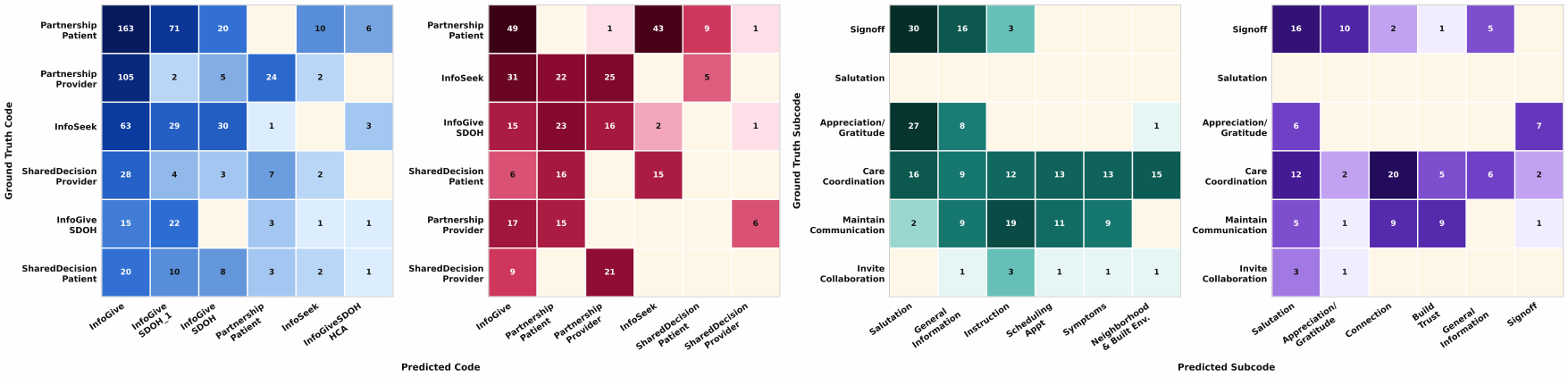}
    \caption{Confusion matrices of selected representative misclassifications in zero-shot Llama-3.2-3B-Instruct, comparing Baseline Instruction (left) and Prompt-Engineered Instruction (right). The left two panels show code-level confusions and the right two panels show subcode-level confusions, highlighting careful error reduction under prompt engineering}
    \label{fig:confusion_matrices}
\end{figure*}

\subsubsection{Confusion Matrix Analysis}

Figure~\ref{fig:confusion_matrices} provides mechanistic evidence for these gains by comparing baseline and prompt-engineered confusion matrices for zero-shot \texttt{Llama-3.2-3B-Instruct}.

\paragraph{Direction-aware error reduction.}
At the Code level, the baseline instruction exhibits prominent confusion between \emph{PartnershipProvider} and \textit{PartnershipPatient}. These are precisely the label pairs whose semantics depend on message direction rather than surface content alone. Under prompt engineering, these confusions contract substantially. This pattern is consistent with the role of the metadata-aware decision logic, which exposes the direction indicator as an explicit control variable and narrows the search space for direction-sensitive Code pairs.

\paragraph{Disambiguation-driven Sub-code improvements.}
At the Sub-code level, the baseline instruction shows recurring confusion between \textit{salutation} and \textit{signoff}, two frequent labels distinguished primarily by position rather than topic. Under prompt engineering, these confusions are markedly reduced. This pattern is consistent with the contribution of the expert-curated disambiguation rules, which encode positional criteria explicitly and operationalize the annotation manual inside the prompt itself.

\paragraph{Compound effects across modules.}
More broadly, the confusion matrices show a general reduction in off-diagonal mass. This suggests that the gains are not attributable to any single module alone. Rather, XML structuring, disambiguation rules, the reasoning scaffold, the schema contract, and the self-validation gate appear to act together to reduce different failure modes: structural drift, semantic ambiguity, and malformed output generation.

\subsection{SFT Closes Much of the Gap, but Prompt Quality Still Matters}
\label{sec:results-sft}

Supervised fine-tuning (SFT) substantially improves performance over all in-context settings, even when trained with the baseline instruction, as expected for completion-only adaptation directly optimized for schema-conformant structured generation (Table~\ref{tab:sft_instruction_comparison}). Span performance is both high and tightly clustered across scales, indicating that SFT reliably teaches structured output format and evidence extraction behavior even at relatively small model sizes, despite the baseline instruction lacking the fully structured design of the prompt-engineered template.

\begin{table*}[!t]
\caption{Performance of supervised fine-tuning with baseline and prompt-engineered instructions on the EPPC Miner benchmark.}
\centering
\resizebox{\textwidth}{!}{
\begin{tabular}{llccccccccc}
\toprule
\multirow{2}{*}{\textbf{Category}} & \multirow{2}{*}{\textbf{Model}} 
& \multicolumn{3}{c}{\textbf{Code}} 
& \multicolumn{3}{c}{\textbf{Sub-code}} 
& \multicolumn{3}{c}{\textbf{Span}} \\
\cmidrule(lr){3-5} \cmidrule(lr){6-8} \cmidrule(lr){9-11}
& & \textbf{Precision} & \textbf{Recall} & \textbf{F1}
& \textbf{Precision} & \textbf{Recall} & \textbf{F1}
& \textbf{Precision} & \textbf{Recall} & \textbf{F1} \\
\midrule

\multirow{4}{*}{SFT with Baseline Instruction}
 & Llama-3.3-70B-Instruct & 75.98 & 82.21 & 78.98 & 64.24 & 70.84 & 67.38 & 85.08 & 96.34 & 90.36 \\
 & Llama-3.1-8B-Instruct  & 75.59 & 75.94 & 75.77 & 62.77 & 68.70 & 65.60 & 85.69 & 96.99 & 90.99 \\
 & Llama-3.2-3B-Instruct  & 73.88 & 76.32 & 75.08 & 58.61 & 62.52 & 60.50 & 86.67 & 95.90 & 91.05 \\
 & Llama-3.2-1B-Instruct  & 67.70 & 65.41 & 66.54 & 53.72 & 52.55 & 53.13 & 87.58 & 93.04 & 90.23 \\
\midrule

\multirow{4}{*}{SFT with Prompt-Engineered Instruction}
 & Llama-3.3-70B-Instruct & 77.41 & 81.16 & 79.24 & 65.92 & 69.75 & 67.78 & 89.97 & 94.93 & 92.38 \\
 & Llama-3.1-8B-Instruct  & 75.96 & 81.45 & 78.61 & 63.58 & 69.75 & 66.52 & 86.86 & 96.52 & 91.43 \\
 & Llama-3.2-3B-Instruct  & 76.29 & 81.58 & 78.85 & 62.05 & 68.03 & 64.90 & 86.84 & 96.59 & 91.46 \\
 & Llama-3.2-1B-Instruct  & 75.65 & 80.20 & 77.86 & 61.41 & 65.97 & 63.21 & 86.91 & 96.01 & 91.23 \\
\bottomrule
\end{tabular}
}
\label{tab:sft_instruction_comparison}
\end{table*}

At the same time, prompt quality continues to matter after supervised adaptation. Comparing prompt-engineered SFT with baseline-instruction SFT, the improvement is scale-dependent and follows a monotonically decreasing pattern. For \texttt{Llama-3.2-1B-Instruct}, prompt-engineered SFT improves Code F1 from 66.54 to 77.86 ($+$11.32) and Sub-code F1 from 53.13 to 63.21 ($+$10.08). For \texttt{Llama-3.2-3B-Instruct}, the differences narrow to $+$3.77 Code F1 and $+$4.40 Sub-code F1. At the 8B scale, the gaps contract further to $+$2.84 Code F1 and $+$0.92 Sub-code F1, and at 70B, they compress to $+$0.26 Code F1 and $+$0.40 Sub-code F1. This monotonic trend is itself informative: it suggests that larger models can partially compensate for weaker task specification through their greater parametric capacity, whereas smaller models remain more dependent on the quality of the instruction template that structures the training signal.

Importantly, the Span metric shows a qualitatively different pattern. Span F1 differences between prompt-engineered and baseline SFT are small and non-monotonic across scales (ranging from $+$0.41 to $+$2.02). This indicates that SFT is highly effective at teaching evidence extraction behavior regardless of prompt quality, and that the persistent prompt-quality effects are concentrated on \emph{semantic} decision-making (Code and Sub-code assignment) rather than on extraction behavior.

The harder part of the task is therefore not structural formatting but \emph{semantic attachment}: deciding which Code and Sub-code should be assigned to each extracted Span. SFT solves much of the formatting problem, but the quality of the instruction still affects how well the model internalizes the task's semantic factorization. This effect is strongest when model capacity is limited, suggesting that prompt engineering and SFT are complementary rather than competing interventions.

\subsection{STaR-DRO Improves Semantic Decisions Beyond SFT}
\label{sec:results-stardro}

Table~\ref{tab:standarddro_stardro} presents the main comparison among SFT, Standard DRO, and STaR-DRO across the Llama family. A clear pattern emerges: STaR-DRO primarily improves the semantically difficult dimensions of the task, namely Code and Sub-code prediction, while preserving already strong Span performance.

For \texttt{Llama-3.3-70B-Instruct}, STaR-DRO improves Code F1 from 79.24 to 81.47 and Sub-code F1 from 67.78 to 69.30, while maintaining Span F1 at 91.73. Notably, both Code precision and recall improve simultaneously, precision rises from 77.41 to 79.08 and recall from 81.16 to 84.00, indicating that the gain reflects better-calibrated semantic decisions rather than a simple precision--recall trade-off in which one axis improves at the expense of the other. The largest absolute gains appear at the largest scale, where the model likely has enough representational headroom to exploit the sharper group-aware training signal.

% ── Heatmap color palette ──────────────────────────────────────────────────────
% Code: light blue family (4 intensity levels)
\definecolor{codeL1}{RGB}{225, 240, 255}   % lightest  (least / no improvement)
\definecolor{codeL2}{RGB}{180, 215, 255}   % light
\definecolor{codeL3}{RGB}{130, 185, 250}   % medium
\definecolor{codeL4}{RGB}{80,  155, 240}   % darkest   (most improvement)

% Sub-code: light green family
\definecolor{subL1}{RGB}{220, 255, 215}    % lightest
\definecolor{subL2}{RGB}{170, 240, 165}    % light
\definecolor{subL3}{RGB}{120, 220, 115}    % medium
\definecolor{subL4}{RGB}{70,  200, 65}     % darkest

% Span: light peach / apricot family (all values below SFT → only 2 levels)
\definecolor{SpanL1}{RGB}{255, 232, 215}   % lightest  (more negative delta)
\definecolor{SpanL2}{RGB}{255, 208, 178}   % slightly richer (near-zero delta)
% ──────────────────────────────────────────────────────────────────────────────

% ── Heatmap color palette ──────────────────────────────────────────────────────
\definecolor{codeL1}{RGB}{225, 240, 255}
\definecolor{codeL2}{RGB}{180, 215, 255}
\definecolor{codeL3}{RGB}{130, 185, 250}
\definecolor{codeL4}{RGB}{80,  155, 240}

\definecolor{subL1}{RGB}{220, 255, 215}
\definecolor{subL2}{RGB}{170, 240, 165}
\definecolor{subL3}{RGB}{120, 220, 115}
\definecolor{subL4}{RGB}{70,  200, 65}

\definecolor{SpanL1}{RGB}{255, 232, 215}
\definecolor{SpanL2}{RGB}{255, 208, 178}
% ──────────────────────────────────────────────────────────────────────────────

\begin{table*}[!t]
\caption{Comparison of Standard DRO and STaR-DRO on the EPPC Miner benchmark.}
\centering
\resizebox{\textwidth}{!}{
\begin{tabular}{llccccccccc}
\toprule
\multirow{2}{*}{\textbf{Category}} & \multirow{2}{*}{\textbf{Model}} 
& \multicolumn{3}{c}{\textbf{Code}} 
& \multicolumn{3}{c}{\textbf{Sub-code}} 
& \multicolumn{3}{c}{\textbf{Span}} \\
\cmidrule(lr){3-5} \cmidrule(lr){6-8} \cmidrule(lr){9-11}
& & \textbf{Precision} & \textbf{Recall} & \textbf{F1}
& \textbf{Precision} & \textbf{Recall} & \textbf{F1}
& \textbf{Precision} & \textbf{Recall} & \textbf{F1} \\
\midrule

\multirow{4}{*}{Standard DRO}
 & Llama-3.3-70B-Instruct & 77.42 & 82.79 & 80.02 & 65.90 & 71.70 & 68.68 & 87.81 & 96.05 & 91.74 \\
 & Llama-3.1-8B-Instruct  & 74.96 & 79.03 & 76.94 & 61.94 & 66.87 & 64.31 & 87.93 & 95.87 & 91.73 \\
 & Llama-3.2-3B-Instruct  & 75.33 & 78.20 & 76.74 & 62.20 & 64.96 & 63.55 & 89.24 & 94.67 & 91.87 \\
 & Llama-3.2-1B-Instruct  & 74.12 & 77.78 & 75.91 & 59.82 & 64.06 & 61.86 & 87.71 & 95.94 & 91.64 \\
\midrule

\multirow{4}{*}{STaR-DRO}
 & Llama-3.3-70B-Instruct
   & \cellcolor{codeL4}79.08 & \cellcolor{codeL4}84.00 & \cellcolor{codeL4}81.47
   & \cellcolor{subL4}66.26  & \cellcolor{subL4}72.64  & \cellcolor{subL4}69.30
   & \cellcolor{SpanL1}87.57 & \cellcolor{SpanL1}96.30 & \cellcolor{SpanL1}91.73 \\
 & Llama-3.1-8B-Instruct
   & \cellcolor{codeL3}77.35 & \cellcolor{codeL3}82.33 & \cellcolor{codeL3}79.77
   & \cellcolor{subL1}63.83  & \cellcolor{subL1}70.05  & \cellcolor{subL1}66.80
   & \cellcolor{SpanL2}86.66 & \cellcolor{SpanL2}96.74 & \cellcolor{SpanL2}91.42 \\
 & Llama-3.2-3B-Instruct
   & \cellcolor{codeL2}78.13 & \cellcolor{codeL2}81.91 & \cellcolor{codeL2}79.98
   & \cellcolor{subL2}63.28  & \cellcolor{subL2}69.04  & \cellcolor{subL2}66.03
   & \cellcolor{SpanL1}86.46 & \cellcolor{SpanL1}96.05 & \cellcolor{SpanL1}91.00 \\
 & Llama-3.2-1B-Instruct
   & \cellcolor{codeL1}76.22 & \cellcolor{codeL1}79.41 & \cellcolor{codeL1}77.78
   & \cellcolor{subL3}62.26  & \cellcolor{subL3}66.79  & \cellcolor{subL3}64.45
   & \cellcolor{SpanL2}86.87 & \cellcolor{SpanL2}95.94 & \cellcolor{SpanL2}91.18 \\
\bottomrule
\end{tabular}
}
\label{tab:standarddro_stardro}
\end{table*}

%%% CHANGE 2: Cross-family STaR-DRO main results %%%
STaR-DRO's performance improvement extends beyond the Llama family. Figures~\ref{fig:qwen25_7b_eppc_results}, \ref{fig:qwen25_14b_eppc_results}, \ref{fig:gemma_2b_eppc_results}, and \ref{fig:gemma_9b_eppc_results} show that STaR-DRO improves Code and Sub-code F1 over SFT while preserving Span performance across both the Qwen and Gemma architectures. For \texttt{gemma-2-9b-it}, Code F1 improves from 0.7982 to 0.8138 ($+$1.56) and Sub-code F1 from 0.6730 to 0.6854 ($+$1.24), the largest non-Llama gains observed, while Span F1 remains at 0.9159. 

\subsubsection{Comparison with Standard DRO}
\label{sub:standard_dro_results}
The comparison with Standard DRO isolates the contribution of STaR-DRO's design choices. Standard DRO shows a sharply mixed pattern. At 70B, it improves Code F1 to 80.02 and Sub-code F1 to 68.68 over SFT, demonstrating that group-robust reweighting can provide a meaningful signal at sufficient model scale. However, for the three smaller models, Standard DRO degrades both Code and Sub-code performance relative to SFT, with average changes of $-1.91$ Code F1 and $-1.64$ Sub-code F1 across the smaller Llama models. These are not marginal regressions; they are consistent, multi-point losses across three model scales.

This contrast highlights two practical limitations of standard group DRO in structured LLM fine-tuning. First, the dense exponentiated-gradient update allocates adversarial mass across all groups at every step, diluting the training signal on the persistently hard groups that matter most. Second, direct minibatch-driven updates without smoothing are more sensitive to noise, and this instability is amplified at smaller model scales where the optimizer is more sensitive to gradient variance. STaR-DRO addresses both issues through stateful EMA-smoothed group statistics and Tsallis mirror updates with sparse back-projection, thereby concentrating emphasis on hard groups while preserving stability.

\begin{figure*}[!htbp]
    \centering
    \includegraphics[width=0.93\textwidth]{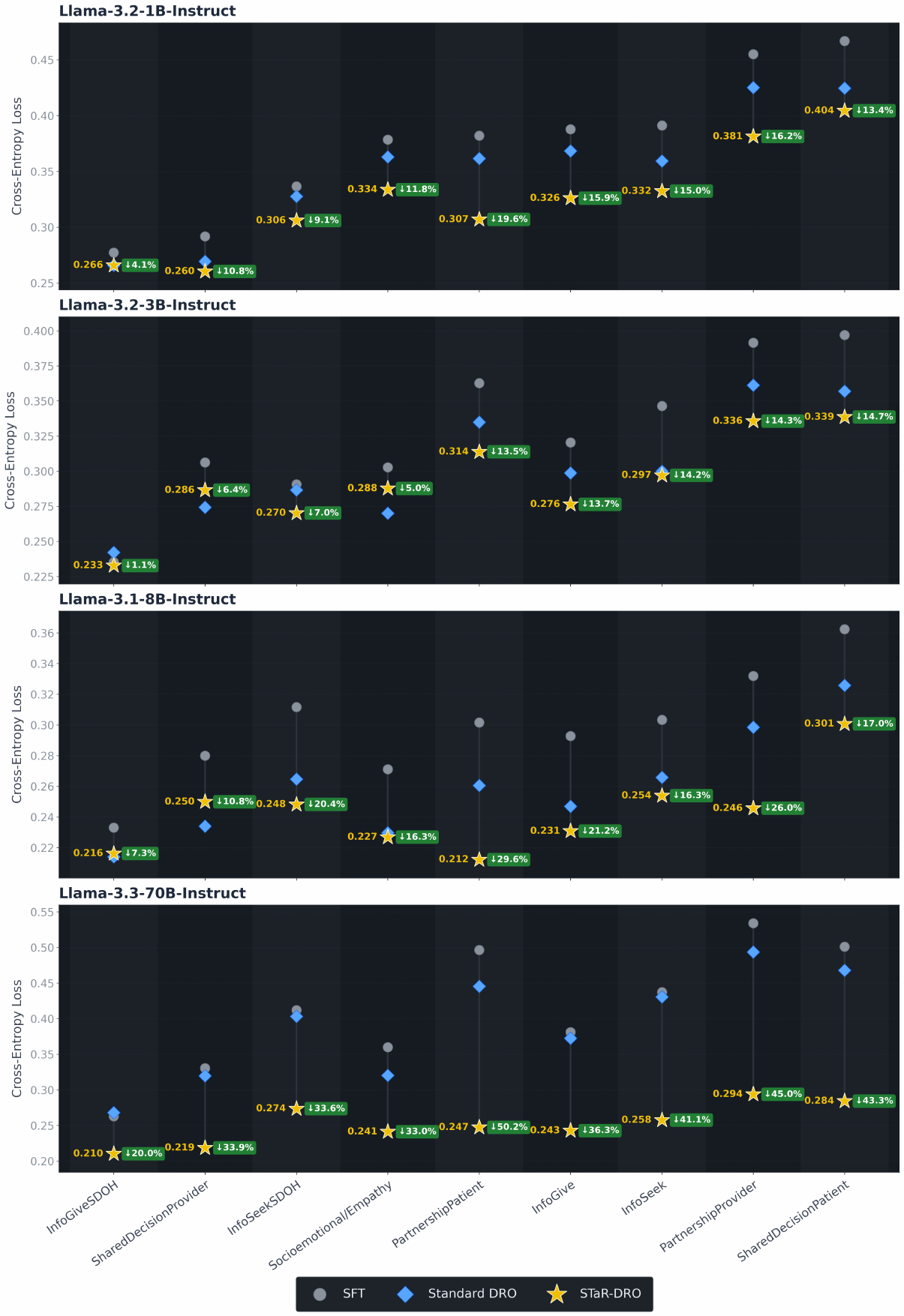}
    \caption{
STaR-DRO achieves near-uniform reductions in group-wise loss across Code-level groups and model scales, with the largest reductions emerging at the 70B scale and the strongest gains concentrated in clinically salient categories.
    }
    \label{fig:stardro_group_loss}
\end{figure*}

\subsubsection{Group-Wise Robustness Analysis}
\label{subsub:Group-Wise Robustness Analysis}

Figures~\ref{fig:stardro_group_loss}, \ref{fig:qwen25_7b_eppc_results}, \ref{fig:qwen25_14b_eppc_results}, \ref{fig:gemma_2b_eppc_results}, and \ref{fig:gemma_9b_eppc_results} provide group-wise validation loss comparisons across three model families. The dominant pattern is broad reduction rather than narrow worst-group rescue, and a consistent structure emerges in \emph{which} groups benefit most.

\paragraph{Representative results: Llama-3.3-70B-Instruct.}

The largest Llama model shows the strongest separation between STaR-DRO and the two baselines, suggesting that the group-aware robust signal becomes especially effective when the model has sufficient representational capacity to use it. On \texttt{Llama-3.3-70B-Instruct}, STaR-DRO reduces validation cross-entropy loss for all nine Code-level groups relative to SFT. The largest reduction occurs for PartnershipPatient, where loss decreases by 50.2\% from 0.496 to 0.247. The remaining groups also improve substantially. This pattern is important because the largest reductions are concentrated in the relational and shared-decision categories that are clinically meaningful but semantically difficult. Standard DRO also lowers loss relative to SFT for the 70B model, indicating that group-robust training can help once model capacity is high enough. However, STaR-DRO remains substantially below Standard DRO across nearly every group, with an average per-group loss reduction of 34.4\% relative to Standard DRO. This shows that the benefit is not merely due to applying group DRO, but to STaR-DRO's stateful, sparse, and bounded excess-only reweighting design. The 70B results therefore provide the clearest evidence that STaR-DRO improves group-wise robustness without simply shifting error from one group to another.

\paragraph{Cross-family consistency in group-level benefit.}
Extending the analysis to Qwen  and Gemma reveals a striking regularity: the same groups benefit most across all three model families. \textit{PartnershipPatient}, \textit{PartnershipProvider}, and \textit{SharedDecisionPatient} are among the top three beneficiaries regardless of architecture or model scale. These are precisely the groups that involve relational and shared-decision communication behaviors, clinically important but semantically ambiguous categories whose correct labeling depends on distinguishing speaker roles, navigating fine-grained ontological boundaries, and resolving direction-dependent semantics. The fact that STaR-DRO preferentially improves these groups across architecturally distinct model families suggests that the benefit is driven by the group structure of the task itself rather than by architecture-specific training dynamics.

Conversely, \textit{InfoGiveSDOH} consistently receives the smallest improvement across all seven models (ranging from $\downarrow$0.5\% for \texttt{gemma-2-9b-it} to $\downarrow$17.3\% for \texttt{Llama-3.1-8B-Instruct}). This group also exhibits the lowest baseline validation loss in every plot, suggesting that it is already well-learned under SFT and offers limited headroom for robust reweighting to exploit.

\paragraph{Exceptions and robustness pattern.}
The pattern is not perfectly uniform. For \texttt{gemma-2-9b-it}, two groups show modest loss \emph{increases}: \textit{SharedDecisionProvider} ($\uparrow$9.0\%) and \textit{Socioemotional/Empathy} ($\uparrow$3.3\%). However, the remaining seven groups all improve, and the aggregate Code and Sub-code F1 still rise (as reported in Section~\ref{sec:results-stardro}). For \texttt{gemma-2-2b-it}, all nine groups improve despite modest aggregate F1 movement, with reductions reaching $\downarrow$36.0\% on \textit{InfoSeekSDOH} and $\downarrow$31.7\% on \textit{PartnershipPatient}.

If STaR-DRO improved worst-group behavior only by sacrificing broader performance, one would expect a mixture of gains and regressions distributed unpredictably across groups. The observed pattern is fundamentally different: across 72 model--group combinations (seven models $\times$ nine groups), loss increases occur in only 2 cases, and both are modest. The broad, group-consistent improvement is a consequence of STaR-DRO's design choices.

\paragraph{Why validation loss rather than group-level classification accuracy.}
We analyze robustness primarily through group-level validation loss because the trained object in this paper is a structured generator, not a flat classifier. STaR-DRO reweights the completion distribution and therefore affects the entire generated object, including semantically decisive annotation tokens, evidence-span tokens, JSON keys, delimiters, and other structural tokens required for schema-valid output. Group-level cross-entropy is therefore the most faithful diagnostic of what the optimizer directly changes. By contrast, Code, Sub-code, and Span F1 are downstream projections of the generated completion onto discrete extracted decisions. They are less granular and only register improvements that change a final labeling or span-matching outcome. Consequently, group-level loss reductions and F1 gains should be expected to be directionally consistent but not proportional: validation loss reflects improved token prediction across the full structured completion, whereas F1 captures only the subset of those improvements that alter discrete extraction behavior.

\paragraph{Robustness to Grouping Strategy.}

Appendix~\ref{sec:appendix_stardro_grouping} examines STaR-DRO under five grouping strategies of increasing granularity in both annotation-level and sample-level attribution modes. The main takeaway is that Code-level annotation grouping is the strongest default setting, but STaR-DRO remains stable across nearby grouping granularities, suggesting that the method does not depend on a finely tuned \emph{a priori} group definition.

\subsection{Ablation Studies}

Table~\ref{tab:stardro_component_ablation} examines whether the gains from STaR-DRO arise from a single reweighting choice or from the interaction of its design components. The results support the latter interpretation. Full STaR-DRO is the only robust-training variant that simultaneously improves both semantic metrics over SFT, with gains of $+1.13$ Code F1 and $+1.13$ Sub-code F1, while also achieving the lowest macro group loss across all Code-level groups. Standard DRO reduces average group loss relative to SFT, but it degrades Code and Sub-code F1 by $-2.11$ and $-1.35$, respectively. This contrast shows that applying group-robust reweighting alone is not sufficient: the geometry and stability of the reweighting mechanism determine whether robustness gains translate into better structured predictions.

\begin{table*}[!t]
\caption{\textbf{Component Ablation of STaR-DRO on Llama-3.2-3B-Instruct.} 
Accuracy columns report absolute changes relative to SFT, 
$\Delta_{\mathrm{SFT}}F1_m = F1_m(\mathrm{Variant}) - F1_m(\mathrm{SFT})$, 
for $m\in\{\mathrm{Code},\mathrm{Sub\mbox{-}code},\mathrm{Span}\}$. 
$\bar{L}_{\mathcal{G}}$ denotes the macro-average cross-entropy over all Code-level groups, computed on the validation set. 
$\bar{L}_{\mathcal{G}_{\mathrm{rare}}}$ and 
$\bar{L}_{\mathcal{G}_{\mathrm{freq}}}$ denote macro-average cross-entropy over rare and frequent groups, respectively. 
Rare groups contain fewer than 50  annotations, whereas frequent groups contain at least 100 annotations.}
\centering
\resizebox{\textwidth}{!}{
\begin{tabular}{lcccccc}
\toprule
\multirow{2}{*}{\textbf{Variant}} 
& \multicolumn{3}{c}{\textbf{Accuracy Change Relative to SFT}} 
& \multicolumn{3}{c}{\textbf{Group Loss}} \\
\cmidrule(lr){2-4} \cmidrule(lr){5-7}
& $\boldsymbol{\Delta_{\mathrm{SFT}} F1_{\mathrm{Code}}}$ 
& $\boldsymbol{\Delta_{\mathrm{SFT}} F1_{\mathrm{Sub\mbox{-}code}}}$ 
& $\boldsymbol{\Delta_{\mathrm{SFT}} F1_{\mathrm{Span}}}$ 
& $\boldsymbol{\bar{L}_{\mathcal{G}}}$ 
& $\boldsymbol{\bar{L}_{\mathcal{G}_{\mathrm{rare}}}}$ 
& $\boldsymbol{\bar{L}_{\mathcal{G}_{\mathrm{freq}}}}$ \\
\midrule

SFT 
& $+0.00$ & $+0.00$ & $+0.00$ 
& 0.3281 & 0.3131 & 0.3582 \\

Standard DRO 
& $-2.11$ & $-1.35$ & $+0.41$ 
& 0.3014 & 0.2862 & 0.3317 \\

STaR-DRO 
& $\mathbf{+1.13}$ & $\mathbf{+1.13}$ & $-0.46$ 
& $\mathbf{0.2932}$ & $\mathbf{0.2855}$ & $\mathbf{0.3086}$ \\

STaR-DRO w/o bounded excess-only multiplier 
& $-0.55$ & $+0.14$ & $-0.10$ 
& 0.3096 & 0.2912 & 0.3463 \\

STaR-DRO w/o Stateful EMA 
& $+0.07$ & $+0.11$ & $-0.06$ 
& 0.3103 & 0.2978 & 0.3351 \\

STaR-DRO w/o scaled ascent signal 
& $+0.27$ & $+1.03$ & $-0.23$ 
& 0.3187 & 0.3077 & 0.3408 \\

\bottomrule
\end{tabular}
}
\label{tab:stardro_component_ablation}
\end{table*}

\paragraph{Component-level evidence for STaR-DRO's design principles.}
The component ablations in Table~\ref{tab:stardro_component_ablation} show that each part of STaR-DRO contributes to a distinct aspect of stable robust optimization:

\begin{itemize}
    \item \textbf{Bounded excess-only multipliers stabilize optimizer-facing reweighting.}
    Removing the bounded excess-only multiplier reverses the Code-level gain, changing Code F1 from $+1.13$ under full STaR-DRO to $-0.55$, and increases the macro group loss from $0.2932$ to $0.3096$. The degradation is especially large for frequent groups, where the loss rises from $0.3086$ to $0.3463$. This supports the role of the multiplier map as an optimizer-facing stabilizer: it converts adversarial mass into bounded upweighting of hard groups without translating simplex competition into excessive suppression or unstable reweighting of easier but still informative groups.

    \item \textbf{Stateful EMA preserves stable signal for frequently observed groups.}
    Removing the stateful EMA weakens both semantic accuracy and group robustness, increasing the macro group loss from $0.2932$ to $0.3103$. The frequent-group loss rises from $0.3086$ to $0.3351$, showing that EMA is not only useful for rare groups but also important for high-support groups that appear repeatedly during training. Because frequent groups are observed in many minibatches and are often already relatively easy under SFT, raw minibatch losses can cause unnecessary oscillations in the adversarial update. EMA smooths these fluctuations and helps distinguish persistent hardness from transient batch-level noise, allowing STaR-DRO to preserve the strong SFT behavior on frequent groups while still reallocating signal toward harder categories.

    \item \textbf{The scaled ascent signal is especially important for rare-group robustness.}
    Removing the scaled ascent signal produces the highest macro group loss among the STaR-DRO variants, increasing $\bar{L}_{\mathcal{G}}$ from $0.2932$ to $0.3187$. It also causes the rare-group loss to rise from $0.2855$ to $0.3077$, the largest rare-group degradation among the component ablations. This indicates that the scaled ascent signal is particularly important when groups are observed unevenly. Rare groups appear less often, so their EMA estimates can be stale, sparse, or influenced by earlier high-loss training phases. By normalizing group losses relative to the present batch's hardness scale, the scaled ascent signal allows rare groups to be compared by current relative difficulty rather than by raw absolute loss magnitude. This helps rare groups receive robust emphasis when they are genuinely hard, without letting outdated or poorly calibrated loss scales dominate the adversarial update.
\end{itemize}

Taken together, these ablations validate the design principles introduced in Section~\ref{sec:stardro}. The small changes in Span F1 should be interpreted in light of the task structure: Span performance is already high after SFT, whereas Code and Sub-code prediction remain the primary semantic bottlenecks. The main effect of STaR-DRO is therefore to improve the hardest remaining semantic decisions while preserving evidence extraction.

\subsection{Regularization interplay.}
Consistent with the analysis of ~\citep{sagawa2020distributionally}, 
the regularization study shows that STaR-DRO is most effective when robust 
reweighting is paired with sufficiently strong explicit regularization. 
Sagawa et al. show that, in overparameterized neural networks, reducing 
worst-group training loss is not sufficient for worst-group generalization: 
models can fit all training groups while still generalizing poorly on 
minority or difficult groups because of group-specific memorization. 
This observation is directly relevant to STaR-DRO. Although STaR-DRO improves where gradient signal is allocated by upweighting 
persistently hard groups, the amplified robust signal can still be spent 
memorizing narrow group-specific patterns unless the model is regularized 
strongly enough to learn group-generalizable rather than group-specific 
solutions. Thus, STaR-DRO and weight 
decay are complementary: STaR-DRO allocates additional gradient signal to hard 
groups, while regularization controls whether that signal produces transferable 
structure rather than memorized group-specific shortcuts.

\begin{figure*}[!t]
    \centering
    \includegraphics[width=0.93\textwidth]{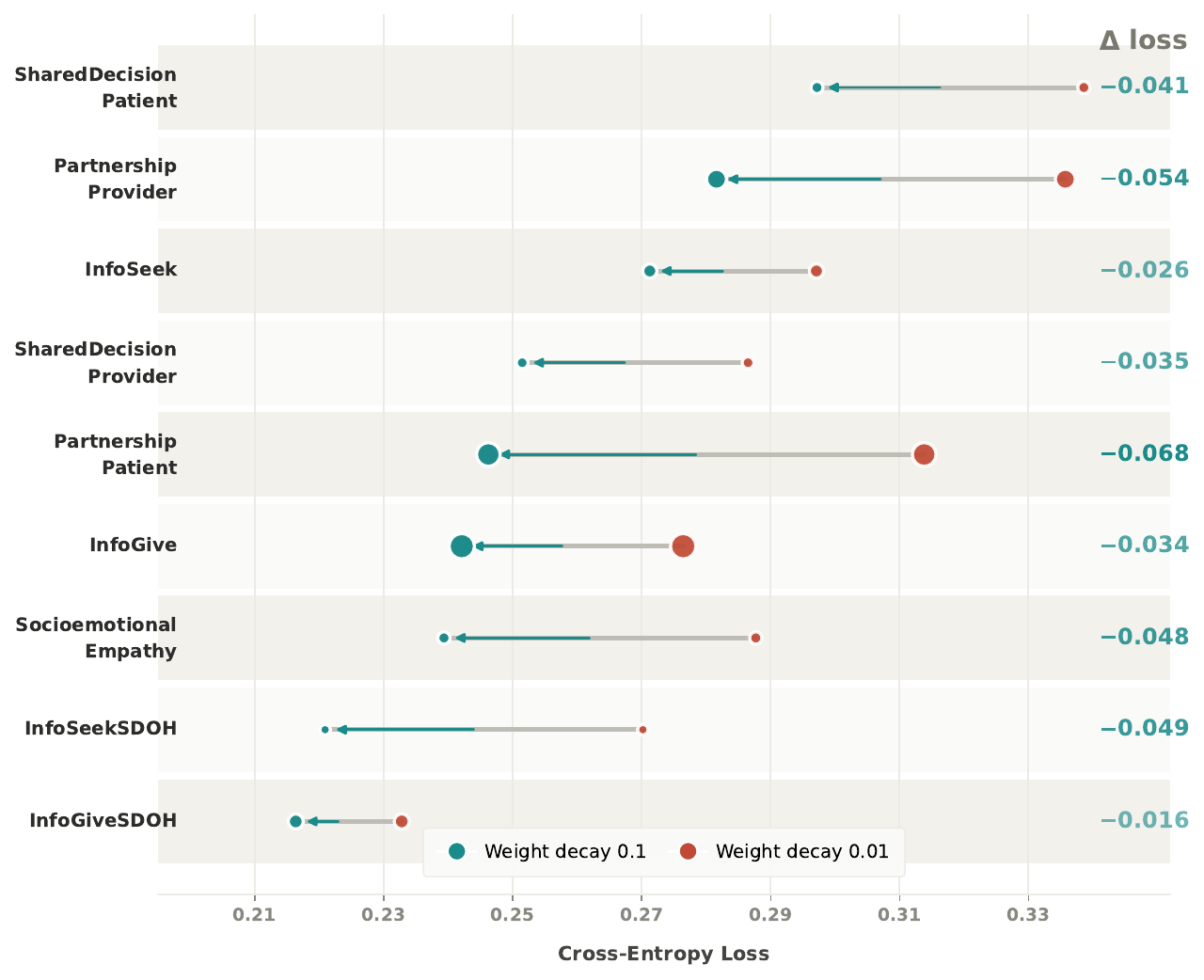}
    \caption{%
        Effect of $\ell_2$ regularization strength on per-group validation cross-entropy loss under STaR-DRO on Llama-3.2-3B-Instruct. Teal: $\lambda_{\mathrm{wd}} = 0.1$; red: $\lambda_{\mathrm{wd}} = 0.01$. Arrows indicate the direction and magnitude of loss reduction ($\Delta$ loss at right). Dot size reflects group sample count. Stronger weight decay uniformly reduces loss across all nine Code-level groups.
    }
    \label{fig:stardro_regularization}
\end{figure*}

Figure~\ref{fig:stardro_regularization} compares per-group validation cross-entropy under weight decay $\lambda_{\mathrm{wd}}=0.1$ versus $\lambda_{\mathrm{wd}}=0.01$, holding the remaining STaR-DRO settings fixed. The results show three clear patterns:

\begin{itemize}
    \item \textbf{Stronger regularization improves every Code-level group.}
    Increasing weight decay from $\lambda_{\mathrm{wd}}=0.01$ to $\lambda_{\mathrm{wd}}=0.1$ reduces validation loss for all nine Code-level groups. The improvements range from $\Delta=-0.016$ on InfoGiveSDOH to $\Delta=-0.068$ on PartnershipPatient, showing that the effect is not limited to a small subset of labels.

    \item \textbf{The largest gains occur on high-loss, semantically difficult groups.}
    The strongest reductions appear for groups such as \emph{PartnershipPatient}, \emph{PartnershipProvider}, and \emph{InfoSeekSDOH}. These are among the more semantically difficult groups, suggesting that stronger regularization helps STaR-DRO convert amplified hard-group signal into more generalizable decision boundaries rather than group-specific memorization.

    \item \textbf{Regularization compresses the group-loss spread.}
    Stronger regularization reduces the inter-group validation-loss range from approximately $[0.22,0.34]$ to $[0.21,0.29]$, roughly halving the spread across groups. This indicates that weight decay does not merely lower average loss; it also makes performance more uniform across heterogeneous Code-level groups.
\end{itemize}

\paragraph{Group-weight distribution regimes.}
Appendix~\ref{app:stardro_ablation} provides the full analysis of 
group-weight distribution regimes under different choices of $\alpha$, $\eta_t$, and 
$\rho_t$. In brief, overly aggressive settings of the effective dual step 
$\eta_{\mathrm{eff}}=(\alpha-1)\eta_t$ collapse the adversarial distribution toward 
a winner-take-all active set, whereas overly weak or delayed robust reweighting leaves 
the distribution nearly uniform and therefore ERM-like. The useful regime is selective 
but stable redistribution: enough concentration to identify persistently hard groups, 
but enough support to preserve signal from groups with moderate difficulty.

% \midrule
\section{Discussion}
\label{sec:discussion}
\paragraph{Extending STaR-DRO beyond SFT to preference optimization and LLM reasoning.}
STaR-DRO works beyond supervised fine-tuning. Although the present paper develops and evaluates the framework in a structured prediction setting under SFT, its core mechanism is not tied to that objective or to structured prediction alone. More broadly, STaR-DRO is applicable to any learning problem that exhibits meaningful group heterogeneity, that is, settings in which some groups are persistently harder, noisier, rarer, or less well served by average-case optimization. The central ingredients of STaR-DRO (stateful tracking of group difficulty, sparse Tsallis reweighting, and bounded group-aware multipliers)  operate on the presence of heterogeneous group difficulty rather than on any single training loss or task family. In our experiments here, the relevant groups are semantically meaningful hierarchical label categories together with grounded evidence extraction, and these same groups are likely to exhibit heterogeneous difficulty not only under SFT but also under preference optimization and reasoning \citep{fodeh2026tab,ouyang2022training,rafailov2024direct,huang2026groupdro_reasoning}. Preference learning signals may be systematically weaker or noisier for some semantic groups than for others, and reasoning chains may be systematically harder to learn for certain communication patterns than for others. In both cases, the underlying problem is the same one STaR-DRO already addresses: average-case optimization can obscure persistent failures on specific groups whose learning signal is sparse, noisy, or inherently more difficult.

This observation motivates a natural extension of the stateful Tsallis reweighting framework to preference optimization \citep{fodeh2026tab,ouyang2022training,rafailov2024direct} and to LLM reasoning over the same group structure \citep{huang2026groupdro_reasoning}. Integrating STaR-DRO into such objectives would allow robust reweighting to target groups where alignment or reasoning signals remain persistently weak, selectively allocating training compute without changing the group definitions themselves. The sparse Tsallis geometry is especially well suited to this regime because preference and reasoning datasets typically provide even fewer observations per group than SFT corpora, making dense reweighting across all groups more vulnerable to noise amplification, while the bounded excess-only multiplier design helps avoid unnecessarily suppressing groups that already receive adequate signal.

\paragraph{Granularity mismatch and the case for an auxiliary objective.}

The diagnostic distinction discussed in Section~\ref{subsub:Group-Wise Robustness Analysis}, under ``Why validation loss rather than group-level classification accuracy,'' also points to a limitation of the current objective. STaR-DRO currently applies group-robust reweighting to the structured completion loss, whereas downstream evaluation depends on discrete Code, Sub-code, and Span decisions extracted from that completion. This creates a granularity mismatch: improvements in token-level likelihood translate into F1 gains only when they affect decision-relevant parts of the generated object, such as label tokens or evidence-span boundaries. A natural future extension is therefore to augment the robust generative objective with an auxiliary classification-style objective directly over the task-defining semantic decisions. In this formulation, STaR-DRO's stateful multipliers could explicitly reweight the auxiliary Code/Sub-code classification loss, perfectly aligning the group-robust optimization pressure with the task definition. Persistently hard groups would receive concentrated adversarial signal not just on predicting the correct sequence of tokens, but on explicitly maximizing the margin between correct and incorrect semantic boundaries. Such an extension would preserve the flexibility of grounded structured generation while allowing the robust objective to act directly on the discrete classification decisions that drive worst-group performance.

\section{Conclusion}
\label{sec:conclusion}

This paper presents an end-to-end framework for robust structured prediction under group heterogeneity, combining sophisticated prompt engineering for structured prediction with STaR-DRO, a task-agnostic stateful sparse robust optimization method applicable to any supervised learning setting where predefined groups exhibit heterogeneous difficulty. Our results on the EPPC Miner benchmark reveal a clear division of labor across adaptation stages, each addressing a distinct failure mode. Prompt engineering substantially improves in-context reliability by making the task more explicit, verifiable, and direction-aware; on \texttt{Llama-3.1-8B-Instruct}, it recovers structured extraction from complete zero-shot failure, improving Code F1 from 0.00 to 52.84. Completion-only supervised fine-tuning then establishes strong schema-conformant generation, raising Code F1 above 77 across all model scales. STaR-DRO adds a distinct third benefit by improving the hardest remaining semantic decisions under group heterogeneity. Across four Llama models, STaR-DRO achieves an average Code F1 of 79.75 and Sub-code F1 of 66.65, outperforming Standard DRO by +2.35 and +2.05 F1 points, respectively, while maintaining strong Span extraction with an average Span F1 of 91.33. At the robustness level, STaR-DRO achieves the largest single-group reduction in validation cross-entropy, reaching 50.2\%. Together, these results show that STaR-DRO offers a stable path toward robust ontology-constrained generation, improving difficult semantic decisions and reducing group-wise loss disparities while preserving schema-valid, evidence-grounded outputs.

\clearpage

\bibliography{sn-bibliography}
\newpage

% Save the last figure/table numbers from the main article
\setcounter{mainfigurecounter}{\value{figure}}
\setcounter{maintablecounter}{\value{table}}

\begin{appendices}

% Keep appendix section numbering as A, B, C, etc.,
% but keep figures and tables numbered continuously as Arabic numerals.
\counterwithout{figure}{section}
\counterwithout{table}{section}
\renewcommand{\thefigure}{\arabic{figure}}
\renewcommand{\thetable}{\arabic{table}}

% Restore the main article counters so appendix figures/tables continue numbering
\setcounter{figure}{\value{mainfigurecounter}}
\setcounter{table}{\value{maintablecounter}}

\part*{Appendix}
\addcontentsline{toc}{part}{Appendix}

% \section*{Contents of the Appendix}
\localtableofcontents

\newpage

\section{Prompt Engineering Techniques}
\label{app:prompt-engineering}

This appendix provides design rationale, implementation details, and failure-mode analysis for the six prompt modules (M1--M6) introduced in Section~\ref{sec:prompt_engineering}. The main text describes \emph{what} each module does and \emph{what effect} it has on performance; this appendix documents \emph{why} each design choice was made and \emph{what specific failure modes} it addresses. The full EPPC Miner prompt instantiation is reproduced below, along with the baseline instruction.

% -------------------------
% Technique color palette
% -------------------------
\definecolor{techXML}{HTML}{1F77B4}   % XML Structuring
\definecolor{techCOT}{HTML}{FF7F0E}   % Chain-of-Thought
\definecolor{techVAL}{HTML}{D62728}   % Self-Validation
\definecolor{techLOG}{HTML}{17BECF}   % Decision Logic
\definecolor{techDIS}{HTML}{9467BD}   % Disambiguation Rules
\definecolor{techPERF}{HTML}{2CA02C}  % Performance Targets

% -------------------------
% Technique macros (monospace-safe)
% -------------------------
\newcommand{\XML}[1]{\textcolor{techXML}{#1}}
\newcommand{\COT}[1]{\textcolor{techCOT}{#1}}
\newcommand{\VAL}[1]{\textcolor{techVAL}{#1}}
\newcommand{\LOGIC}[1]{\textcolor{techLOG}{#1}}
\newcommand{\DIS}[1]{\textcolor{techDIS}{#1}}
\newcommand{\PERF}[1]{\textcolor{techPERF}{#1}}

\newcommand{\LegendItem}[2]{\textcolor{#1}{\rule{1.15ex}{1.15ex}}\hspace{0.5em}#2}
\newtcolorbox{legendbox}{
  enhanced,
  colback=white,
  colframe=black!35,
  boxrule=0.6pt,
  arc=1.2mm,
  left=5pt,right=5pt,top=4pt,bottom=4pt,
}

\newtcolorbox{promptbox}[1][]{
  enhanced,
  breakable,
  colbacktitle=purple!70!black,
  colframe=purple!60!black,
  colback=purple!4!white,
  coltitle=white,
  title={Prompt Engineered Instruction},
  boxrule=0.9pt,
  arc=2mm,
  left=6pt,right=6pt,top=6pt,bottom=6pt,
  #1
}

\newtcolorbox{promptbox_1}[1][]{
  enhanced,
  breakable,
  colbacktitle=purple!70!black,
  colframe=purple!60!black,
  colback=purple!4!white,
  coltitle=white,
  boxrule=0.9pt,
  arc=2mm,
  left=6pt,right=6pt,top=6pt,bottom=6pt,
  #1
}

% =========================
% In the paper body
% =========================

\begin{promptbox}[colbacktitle=purple!70!black,colframe=purple!60!black,colback=purple!4!white]
\ttfamily\footnotesize
% \vspace{0.35em}
\vspace{0.4em}
\begin{legendbox}
\LegendItem{techXML}{XML Structuring}\\
\LegendItem{techCOT}{Chain-of-Thought (4-step reasoning)}\\
\LegendItem{techVAL}{Self-Validation (quality gate)}\\
\LegendItem{techLOG}{Decision Logic (direction-aware)}\\
\LegendItem{techDIS}{Disambiguation Rules}\\
\LegendItem{techPERF}{Performance Targets}
\end{legendbox}

\begin{alltt}
\XML{<role>}
Expert patient-centered communication analyst with \PERF{>95\%} accuracy in medical message \\ multi-label classification.
\XML{</role>}

\XML{<performance_target>}
\PERF{CRITICAL REQUIREMENTS:}
\PERF{- Code Accuracy: >95\%}
\PERF{- Sub-code Accuracy: >95\%}
\PERF{- Span Accuracy: >98\% (character-perfect)}

\PERF{Every annotation must be defensible and verification-validated.}
\XML{</performance_target>}

\XML{<task>}
Extract Code, Sub-code, and Span triples from medical secure messages.

INPUT:
- Context (the message text)
- Message Direction (\LOGIC{TO\_PAT\_YN})

OUTPUT:
- JSON list of \{Code, Sub-code, Span\} objects

CONSTRAINT:
- MULTI-LABEL task (one message may contain multiple valid triples)
\XML{</task>}

\XML{<message_direction>}
\LOGIC{CRITICAL: Message direction determines Code selection.}

\LOGIC{- TO\_PAT\_YN = "Y": Provider speaking TO patient}
\LOGIC{- TO\_PAT\_YN = "N": Patient speaking TO provider}

\LOGIC{USE CASES:}
\LOGIC{- Provider to patient: Use PartnershipProvider, SharedDecisionProvider,}
\LOGIC{  CareCoordinationProvider when applicable}
\LOGIC{- Patient to provider: Use PartnershipPatient, SharedDecisionPatient,}
\LOGIC{  CareCoordinationPatient when applicable}
\LOGIC{- SDOH and SocioEmotionalBehaviour apply regardless of direction}
\XML{</message_direction>}
\end{alltt}
\end{promptbox}

\begin{promptbox_1}[colbacktitle=purple!70!black,colframe=purple!60!black,colback=purple!4!white]
\ttfamily\footnotesize
\begin{alltt}
\XML{<critical_rules>}
\VAL{RULE VIOLATIONS RESULT IN ANNOTATION FAILURE}

\VAL{1. Span Source:}
\VAL{   - Extract Spans ONLY from the provided message text}
\VAL{   - Context is for understanding only}
\VAL{   - Never invent, paraphrase, or infer Spans}

\VAL{2. Code and Sub-code Validity:}
\VAL{   - Every Sub-code MUST be valid for its Code}
\VAL{   - If a pairing is illogical or invalid, loop back and re-select}

\VAL{3. Span Exactness:}
\VAL{   - Copy EXACT text from the message}
\VAL{   - Preserve punctuation, capitalization, and spacing}
\VAL{   - No paraphrasing}

\VAL{4. Multi-label Requirement:}
\VAL{   - Identify ALL relevant Code and Sub-code pairs in the message}
\XML{</critical_rules>}

\XML{<reasoning_process>}
\COT{Follow this 4-step verification process:}

\COT{STEP 1: CONTEXT AND DIRECTION ANALYSIS}
\COT{- Read the full message carefully}
\COT{- Determine message direction using TO\_PAT\_YN}
\COT{- Understand speaker intent and conversational goal}

\COT{STEP 2: PHRASE DECOMPOSITION AND CODE MATCHING}
\COT{- Break the message into semantic units (phrases or clauses)}
\COT{- For each phrase, identify intent:}
\COT{  * SDOH}
\COT{  * PartnershipProvider or PartnershipPatient}
\COT{  * SharedDecisionProvider or SharedDecisionPatient}
\COT{  * CareCoordinationProvider or CareCoordinationPatient}
\COT{  * SocioEmotionalBehaviour}
\COT{- Use TO\_PAT\_YN to select Provider vs Patient variants}
\COT{- Match each phrase to the correct Code definition}
\COT{- Verify Code and Sub-code pairing is logical and valid}

\COT{STEP 3: Span EXTRACTION AND VERIFICATION}
\COT{- Extract the MINIMUM complete supporting phrase}
\COT{- Spans must come ONLY from the message text}
\COT{- Verify character-level exactness}
\COT{- If the Span does not exist exactly, reject the annotation}

\COT{STEP 4: CROSS-VALIDATION (MOST IMPORTANT)}
\COT{Verification priority:}
\COT{1. Best semantic match confirmed (if not, loop back to Step 2)}
\COT{2. Sub-code valid for Code}
\COT{3. Span is exact and present in message}
\COT{4. All relevant phrases analyzed}
\COT{5. Disambiguation rules applied correctly}
\COT{6. High-confidence annotation defensible to experts}
\XML{</reasoning_process>}

\XML{<codes_definitions>}
The following are authoritative ground-truth definitions. Names must match exactly.
(\textit{Full list omitted here for brevity}).

FORMAT (Code WITH Sub-codes):
CODE_NAME: <one-sentence operational definition>.
|- SUBCODE_1: <one-sentence operational definition>.
|- SUBCODE_2: <one-sentence operational definition>.
\end{alltt}
\end{promptbox_1}

\begin{promptbox_1}[colbacktitle=purple!70!black,colframe=purple!60!black,colback=purple!4!white]
\ttfamily\footnotesize
\begin{alltt}
|- SUBCODE_K: <one-sentence operational definition>.

FORMAT (Code WITHOUT Sub-codes):
CODE_NAME: <one-sentence operational definition>.
|- None: No Sub-codes are defined for this Code.
\XML{</codes_definitions>}

\XML{<disambiguation_rules>}
\DIS{Apply systematically to resolve ambiguity.}

\DIS{Salutation vs Signoff:}
\DIS{- Opening greetings indicate salutation}
\DIS{- Closing phrases indicate signoff}
\DIS{- Position determines classification}

\DIS{Appreciation/Gratitude vs Signoff:}
\DIS{- Simple closing thanks indicates signoff}
\DIS{- Specific appreciation indicates Appreciation/Gratitude}

\DIS{Provider vs Patient Codes:}
\DIS{- Use TO\_PAT\_YN strictly}
\DIS{- TO\_PAT\_YN = "Y" -> Provider codes}
\DIS{- TO\_PAT\_YN = "N" -> Patient codes}

\DIS{SharedDecision Codes:}
\DIS{- Use TO\_PAT\_YN to select Provider vs Patient variants}

\DIS{SDOH Sub-code Selection:}
\DIS{- EconomicStability: finances, income, food, housing}
\DIS{- EducationAccessAndQuality: education, literacy}
\DIS{- HealthCareAccessAndQuality: access to care, insurance, lifestyle}
\DIS{- NeighborhoodAndBuiltEnvironment: housing, transportation, environment}
\DIS{- SocialAndCommunityContext: social support, isolation, discrimination}

\DIS{CareCoordination vs maintainCommunication:}
\DIS{- maintainCommunication: future updates only}
\DIS{- CareCoordination: concrete coordination with other providers}

\DIS{requestsForOpinion vs inviteCollaboration:}
\DIS{- requestsForOpinion: asks patient views}
\DIS{- inviteCollaboration: invites joint participation}

\DIS{SocioEmotionalBehaviour:}
\DIS{- Emotional support, reassurance, empathy, politeness}
\DIS{- Only Sub-code "None" is valid}
\XML{</disambiguation_rules>}

\XML{<output_format>}
Return JSON with a "results" array:

\{
  "results": [
    \{
      "Code": "exact Code name",
      "Sub-code": "exact Sub-code name",
      "Span": "EXACT text from message"
    \}
  ]
\}

If no annotations apply:
\{"results": []\}
\XML{</output_format>}
\end{alltt}
\end{promptbox_1}

\begin{promptbox_1}[colbacktitle=purple!70!black,colframe=purple!60!black,colback=purple!4!white]
\ttfamily\footnotesize
\begin{alltt}
\XML{<quality_gate>}
\VAL{MANDATORY verification before submission:}

\VAL{1. JSON is parseable}
\VAL{2. All Sub-codes valid for Codes}
\VAL{3. All Spans are exact and present in message}
\VAL{4. Best semantic match verified}
\VAL{5. All disambiguation rules applied}
\VAL{6. High confidence suitable for expert review}

\VAL{Accuracy is paramount. Quality over speed.}
\XML{</quality_gate>}

INPUT:
\LOGIC{TO\_PAT\_YN: N (Patient speaking to provider)}

Context:
Dr. Person1 I need my prescription sent to the pharmacy for my flecainide acetate
100 mg tablets twice a day the pharmacist has try requesting it no success and I
don't have any pills. Person2
\end{alltt}
\end{promptbox_1}

% =========================
% PREAMBLE
% =========================

\newtcolorbox{promptboxone}[1][]{
  enhanced,
  breakable,
  colbacktitle=blue!70!black,
  colframe=blue!60!black,
  colback=blue!4!white,
  coltitle=white,
  title={Baseline Prompt},
  boxrule=0.9pt,
  arc=2mm,
  left=6pt,right=6pt,top=6pt,bottom=6pt,
  #1
}

\newtcolorbox{promptboxone_1}[1][]{
  enhanced,
  breakable,
  colframe=blue!60!black,
  colback=blue!4!white,
  boxrule=0.9pt,
  arc=2mm,
  left=6pt,right=6pt,top=6pt,bottom=6pt,
  #1
}

% =========================
% IN THE PAPER BODY
% =========================
\Needspace{12\baselineskip}
\begin{promptboxone}
\begin{Verbatim}[
  breaklines=true,
  breakanywhere=true,
  breaksymbolleft={},
  breaksymbolright={},
  xleftmargin=0pt,
  formatcom=\ttfamily\footnotesize\color{black},
  commandchars=\\\{\}
]
You are a patient-centered communication analyst tasked with identifying and classifying how patients and clinicians incorporate patient-centered communication (PCC) elements in secure messaging.

Your goal is to extract multiple Code-Sub-code pairs from the current sentence and identify specific Spans corresponding to each pair. This task requires careful, step-by-step reasoning to ensure accurate multi-label classification, with additional consideration of contextual information from surrounding sentences.

## Follow these steps systematically and step-by-step Instructions:
1. Understand the Input Sentence:
1.1 Analyze the message to establish the full context.
1.2 Note:
1.3 Carefully read and analyze every word in the message to determine its context and identify all relevant communication elements.
2. Identify Relevant Codes:
2.1 Match parts of the message to one or more Codes based on the intent and content described in the definitions below.
2.3 Acknowledge that a message may involve multiple Codes.
3. Determine Sub-codes for Each Code:
3.1 For each identified Code, assign the appropriate Sub-code(s) that further specify the meaning.
3.2 Use definitions of Sub-codes to ensure accuracy and consistency.
3.4 Important: Ensure that the Sub-code you select belongs to the Sub-code list under the identified Code. If it doesn't, reconsider whether the Code or Sub-code selection is correct.
4. Pair Codes with Sub-codes:
4.1 Form unique Code-Sub-code pairs for the message. These pairs should fully describe the meaning of the message.
4.2 If multiple Codes exist in the same message, their Sub-codes will differ.
5. Highlight Evidence for Each Pair:
5.1 Extract minimal, specific Spans of text from the message that support each identified Code-Sub-code pair.
5.2 Note: The extracted minimum Span should be a core phrase in the message instead of the entire sentence.

The following content provides definitions for Codes and Sub-Codes.

## Code and Definitions:

(\textit{Full list omitted here for brevity}).
\end{Verbatim}
\end{promptboxone}
\begin{promptboxone_1}
\begin{Verbatim}[
  breaklines=true,
  breakanywhere=true,
  breaksymbolleft={},
  breaksymbolright={},
  xleftmargin=0pt,
  formatcom=\ttfamily\footnotesize\color{black},
  commandchars=\\\{\}
]
FORMAT (Code WITH Sub-codes):
CODE_NAME: <one-sentence operational definition>.
|- SUBCODE_1: <one-sentence operational definition>.
|- SUBCODE_2: <one-sentence operational definition>.
|- SUBCODE_K: <one-sentence operational definition>.

FORMAT (Code WITHOUT Sub-codes):
CODE_NAME: <one-sentence operational definition>.
|- None: No Sub-codes are defined for this Code.

Ensure your reasoning is step-by-step to capture all relevant Code-Sub-code pairs and their corresponding Spans accurately. Remember, the Sub-code must belong to the list of Sub-codes under the identified Code.

IMPORTANT: Output your final result without any explanation and reasoning, you must output the JSON format like {"results": [{"Code": "<Identified Code>_1", "Sub-code": "<Identified Sub-code>_1", "Span": "<Extracted Span>_1"},...,{"Code": "<Identified Code>_n", "Sub-code": "<Identified Sub-code>_n", "Span": "<Extracted Span>_n"}]}
\end{Verbatim}
\end{promptboxone_1}

\subsection{M1: XML-Style Structural Segmentation}
\label{app:prompt-xml}

The prompt uses semantically tagged blocks (\texttt{<role>}, \texttt{<task>}, \texttt{<message\_direction>}, \texttt{<critical\_rules>}, \texttt{<reasoning\_process>}, \texttt{<disambiguation\_rules>}, \texttt{<output\_format>}, \texttt{<quality\_gate>}) as lightweight structural delimiters. The model does not parse these as formal XML; rather, the tags serve two functional roles that address specific failure modes observed during prompt development.

First, they impose \emph{scope boundaries} between modules within a long instruction set. In extended prompts containing role definitions, ontology inventories, disambiguation rules, reasoning steps, and output schemas, instruction-tuned models are susceptible to context rot: the progressive degradation of constraint adherence as the distance between a rule's declaration and its point of application increases. The tagged structure mitigates this by partitioning the instruction into discrete, self-contained blocks, each with a clear semantic scope. This reduces cross-module interference (e.g., disambiguation rules bleeding into output-format interpretation) and provides concrete referential anchors that other modules can point back to, when the quality gate (M6) directs the model to ``verify all disambiguation rules are applied,'' the existence of a discrete \texttt{<disambiguation\_rules>} block gives it a bounded scope to check against, rather than requiring recall of rules distributed across an unstructured paragraph.

Second, they support \emph{modular substitution}. Because each tagged block is self-contained, domain adaptation requires only replacing the content within the relevant tags, swapping label inventories, disambiguation criteria, metadata variables, and output schemas, while preserving the architectural relationships between blocks. This modularity is what makes the template applicable beyond EPPC Miner to other structured prediction tasks with hierarchical labeling and grounded evidence requirements.

\subsection{M2: Disambiguation Rule Selection and Coverage}
\label{app:prompt-disambiguation}

The disambiguation rules were not designed from first principles but were derived empirically from the annotation adjudication process. During calibration and consensus meetings, systematic disagreements between annotators were documented, and the resolution criteria established by senior investigators were translated into explicit decision rules within the prompt.

The current rule set covers five empirically frequent confusion pairs:

\begin{itemize}
    \item \textbf{salutation vs.\ signoff}: Distinguished by position (opening vs.\ closing) rather than lexical content. Both can involve names and pleasantries; the rule enforces positional classification.
    \item \textbf{Appreciation/Gratitude vs.\ signoff}: Simple closing thanks (e.g., ``Thank you'') is classified as signoff; specific appreciation referencing a concrete action or quality is classified as Appreciation/Gratitude.
    \item \textbf{Provider vs.\ Patient Code variants}: Resolved entirely by the \texttt{TO\_PAT\_YN} direction indicator (see M4 below). The rules are stated as hard conditionals rather than soft preferences.
    \item \textbf{carecoordination vs.\ maintainCommunication}: Distinguished by specificity: concrete coordination with other providers or services indicates CareCoordination, while statements about future updates or availability indicate maintainCommunication.
    \item \textbf{requestsForOpinion vs.\ inviteCollaboration}: Distinguished by the degree of participation solicited: asking for a patient's views is requestsForOpinion, whereas inviting active joint involvement is inviteCollaboration.
\end{itemize}

These five pairs were selected because they accounted for the majority of annotator disagreements during the adjudication phase. The rules are intentionally stated as crisp decision boundaries rather than probabilistic guidance, because the downstream evaluation is exact-match at the label level.

\subsection{M3: Reasoning Scaffold as Verification Routine}
\label{app:prompt-cot}

The four-step scaffold (context analysis $\to$ phrase decomposition $\to$ Span extraction $\to$ cross-validation) differs from standard chain-of-thought prompting \citep{wei2022chain} in two important respects.

First, it is \emph{task-specialized} rather than general-purpose. Each step targets a specific component of the structured output: Step~1 establishes the direction-dependent context that determines Code family eligibility; Step~2 performs the semantic decomposition that produces Code and Sub-code candidates; Step~3 grounds each candidate in verbatim evidence; and Step~4 performs ontology-consistency checks across the full annotation set. This specialization means the intermediate reasoning is not free-form explanation but a structured verification sequence aligned with the output contract.

Second, the scaffold includes an explicit \emph{loop-back condition}: if cross-validation in Step~4 reveals an inconsistency (e.g., a Sub-code that is invalid for its parent Code, or a Span that does not appear verbatim in the source text), the model is instructed to return to Step~2 and re-evaluate. This conditional retry within a single generation pass distinguishes the approach from both standard chain-of-thought (which is linear) and iterative self-refinement (which requires multiple inference calls).

\subsection{M4: Direction-Aware Control and the Label Search Space}
\label{app:prompt-direction}

The direction indicator \texttt{TO\_PAT\_YN} is exposed as an explicit control variable with hard decision rules rather than soft contextual guidance. This design choice addresses a specific structural property of the EPPC ontology: multiple Code pairs differ \emph{only} in speaker direction (\emph{PartnershipPatient} vs.\ \emph{PartnershipProvider}; \emph{SharedDecisionPatient} vs.\ \emph{SharedDecisionProvider}), while sharing identical Sub-code inventories and nearly identical communicative functions. Without explicit direction conditioning, the model must infer speaker role from surface cues in the message text, a task that is unreliable for messages with neutral or ambiguous phrasing.

By stating the mapping as a hard conditional (\texttt{TO\_PAT\_YN = Y} $\to$ Provider codes; \texttt{TO\_PAT\_YN = N} $\to$ Patient codes), the prompt effectively halves the eligible Code inventory for any given instance. This search-space reduction has a compounding benefit: fewer eligible Codes means fewer eligible Code--Sub-code pairs, which in turn reduces the probability of generating ontologically invalid combinations. The confusion matrix analysis in Section~\ref{sec:results-prompt} confirms that this module is the primary driver of reduced direction-conditioned misclassification.

\subsection{M5: Output Schema Contract and Structural Enforcement}
\label{app:prompt-schema}

The output schema is specified as a JSON contract with three hard constraints: (i) each entry must contain exactly the fields \texttt{Code}, \texttt{Sub-code}, and \texttt{Span}; (ii) Spans must be verbatim substrings of the source message, with no paraphrasing, insertion, or truncation; and (iii) the output must be a JSON array, enforcing the multi-label nature of the task (a single message may yield zero or more annotation triples).

These constraints serve a dual purpose. At inference time, they shift the generation task from open-ended text completion to constrained fill-in within a fixed structural template, which reduces format drift, a common failure mode in long structured outputs where the model gradually deviates from the declared schema over successive entries. At evaluation time, they enable deterministic parsing: the output can be programmatically extracted and evaluated without heuristic post-processing, which is necessary for reproducible metric computation. 

\subsection{M6: Single-Turn Quality Gate}
\label{app:prompt-selfval}

The quality gate is a six-point checklist appended after the output schema:

\begin{enumerate}
    \item JSON output is parseable under the declared schema.
    \item All Sub-codes are valid for their parent Codes.
    \item All Spans are exact substrings present in the source message.
    \item Best semantic match has been verified for each annotation.
    \item All disambiguation rules have been applied.
    \item Annotation is high-confidence and defensible to expert review.
\end{enumerate}

This checklist is designed as an \emph{intra-generation} audit: the model is expected to evaluate these conditions within the same inference pass that produces the output, without access to external feedback or a second generation call. This distinguishes M6 from iterative self-refinement methods such as Self-Refine \citep{madaan2023selfrefine}, which use multi-turn critique-and-revision loops. The single-turn constraint is motivated by two considerations. First, our evaluation protocol requires single-pass inference for efficiency and reproducibility. Second, recent evidence suggests that intrinsic self-correction, where the model revises its own output without external signal is often unreliable on reasoning-intensive tasks and can degrade performance relative to first-pass generation \citep{huang2024cannot}. The quality gate therefore does not ask the model to \emph{change} its output but to \emph{verify} it against concrete, checkable conditions before emission. In effect, it functions as a pre-submission audit rather than a post-hoc correction mechanism.

\subsection{Interactions Between Modules}
\label{app:prompt-interactions}

The six modules are designed to be individually interpretable but jointly effective. Several interaction patterns are worth noting.

M1 (XML structuring) supports M3 (reasoning scaffold) and M6 (quality gate) by providing referential anchors: the tagged blocks give the verification steps concrete scopes to check against. M2 (disambiguation rules) is most effective when combined with M3, because the reasoning scaffold's phrase-decomposition step is where ambiguous cases are first identified and where the disambiguation rules are applied. M4 (direction control) interacts multiplicatively with M5 (schema contract): direction conditioning halves the eligible Code inventory, and the schema contract then enforces that the Sub-codes selected are valid under the narrowed Code set. M6 (quality gate) functions as a final consistency check across all preceding modules. It verifies that the output satisfies the schema (M5), that disambiguation rules were applied (M2), and that Spans are grounded (M3 Step~3).

The compound effect of these interactions is visible in our experimental results. This pattern is consistent with the modules addressing qualitatively different failure modes: structural drift (M1, M5), semantic ambiguity (M2, M4), reasoning shortcuts (M3), and verification omission (M6).

\clearpage

\section{Cross-Family Results: Qwen and Gemma Models}

This section presents extended EPPC Miner results for the Qwen and Gemma model family, showing that the proposed prompting and STaR-DRO framework remains effective across Qwen and Gemma-scale variants.

\begin{figure*}[!htbp]
\centering
\includegraphics[width=0.9\textwidth]{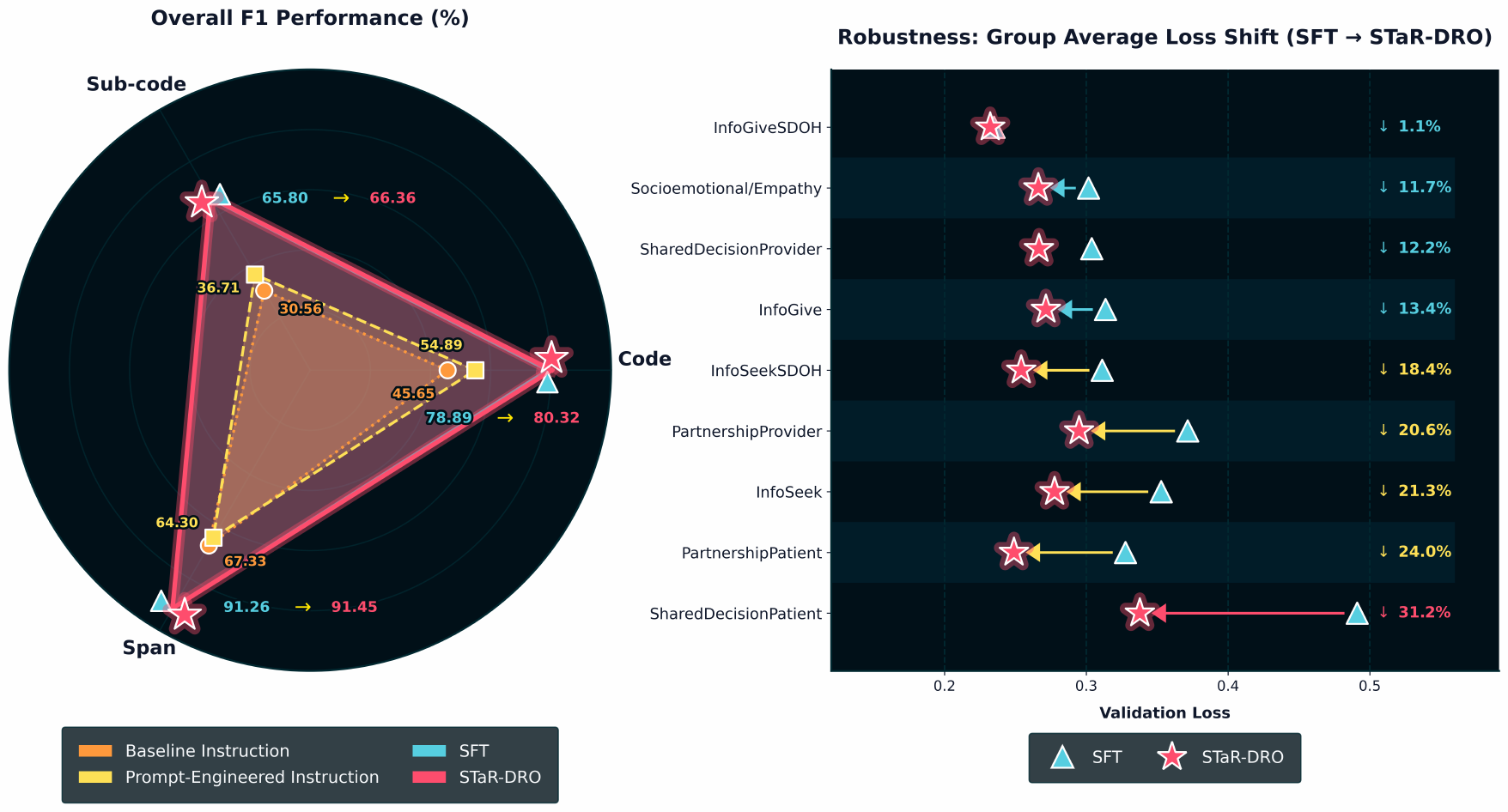}
\caption{F1 performance and group-wise robustness of Qwen2.5-7B-Instruct on the EPPC Miner benchmark.}
\label{fig:qwen25_7b_eppc_results}
\end{figure*}

\begin{figure*}[!htbp]
\centering
\includegraphics[width=0.9\textwidth]{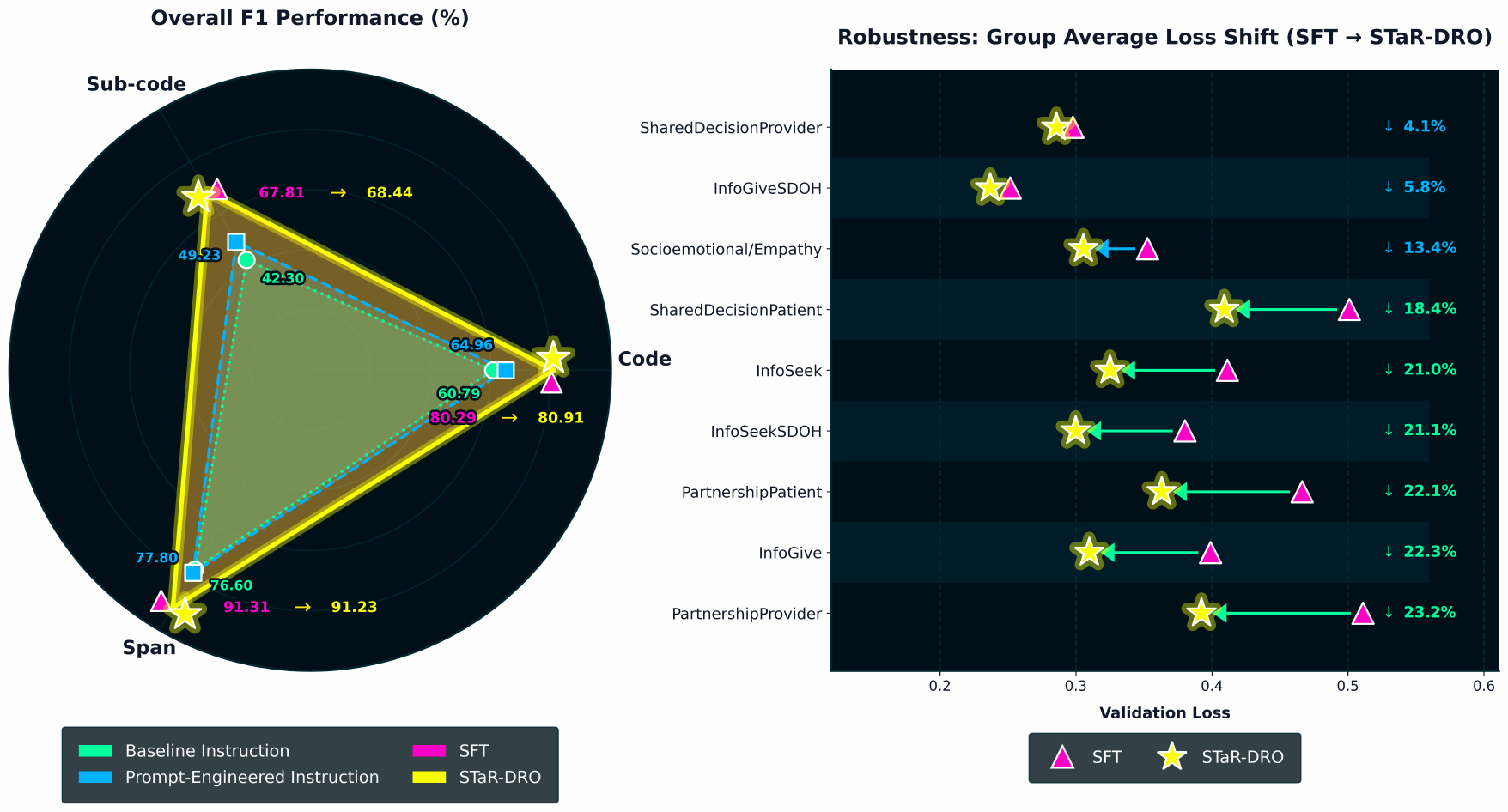}
\caption{F1 performance and group-wise robustness of Qwen2.5-14B-Instruct on the EPPC Miner benchmark.}
\label{fig:qwen25_14b_eppc_results}
\end{figure*}

\begin{figure*}[!htbp]
    \centering
    \includegraphics[width=0.9\textwidth]{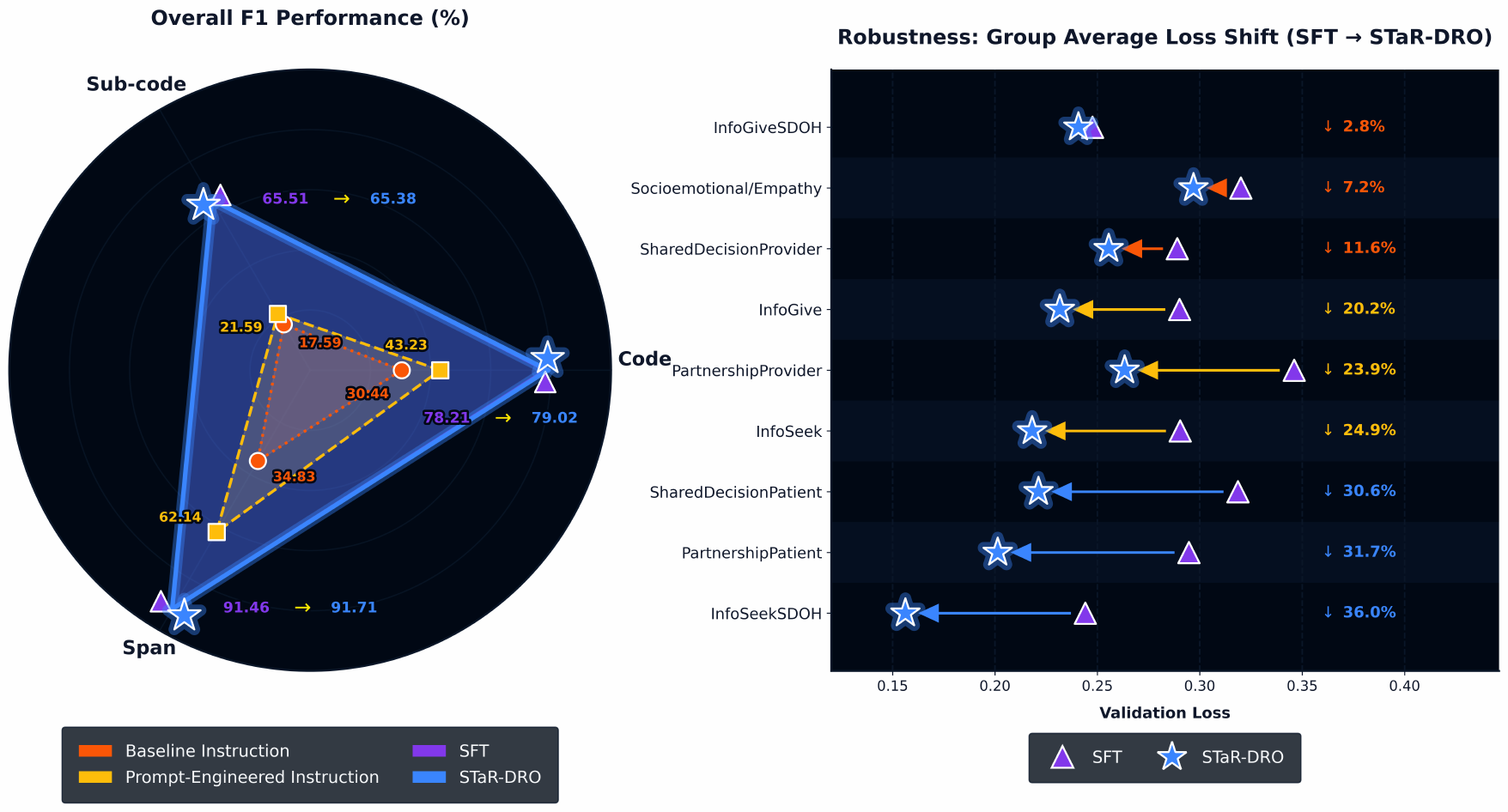}
    \caption{F1 performance and group-wise robustness of gemma-2-2b-it on the EPPC Miner benchmark.}
    \label{fig:gemma_2b_eppc_results}
\end{figure*}

\vspace{-2mm}

\begin{figure*}[!htbp]
    \centering
    \includegraphics[width=0.9\textwidth]{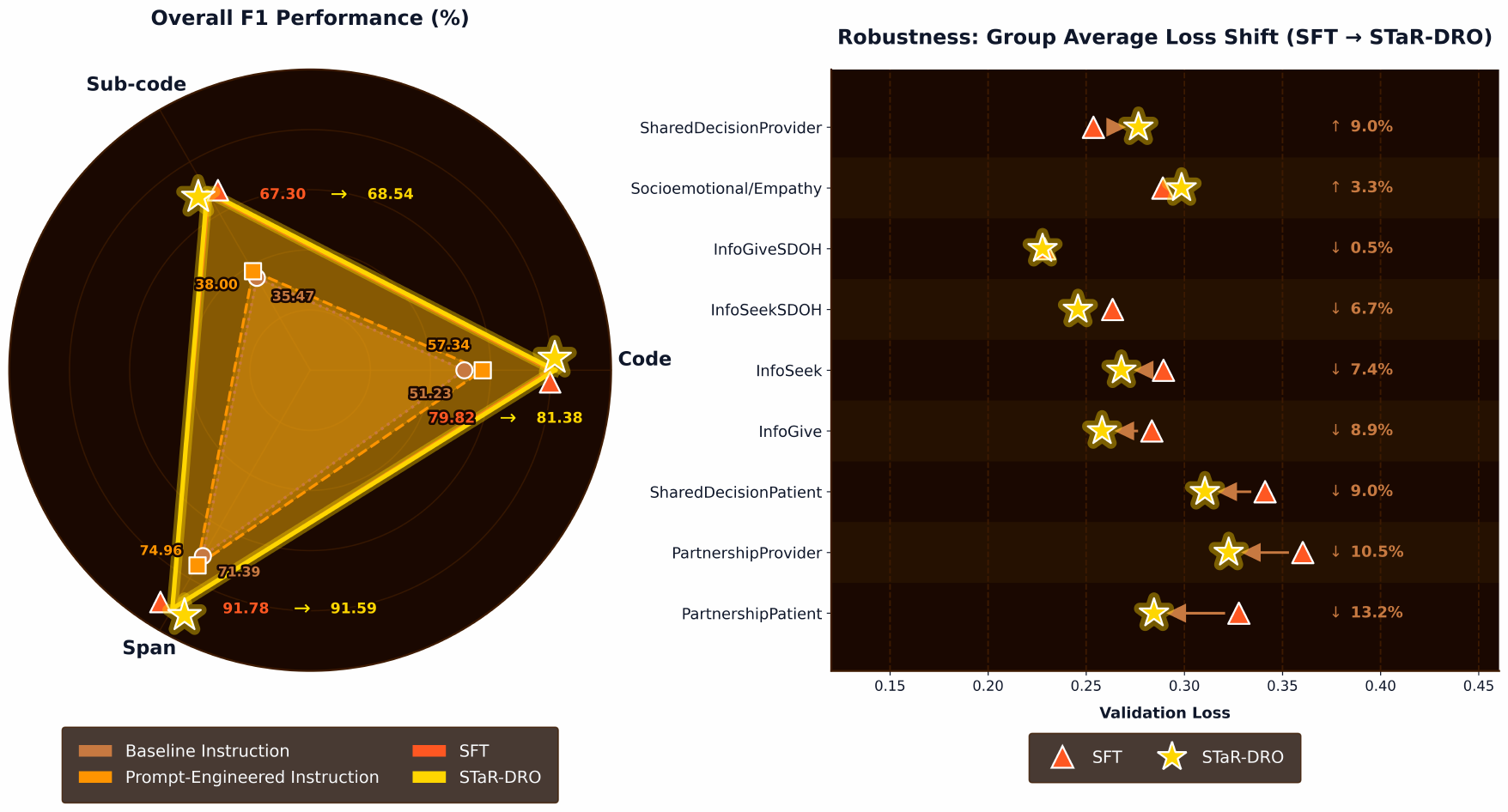}
    \caption{F1 performance and group-wise robustness of gemma-2-9b-it on the EPPC Miner benchmark.}
    \label{fig:gemma_9b_eppc_results}
\end{figure*}

\clearpage
\section{EPPC Miner Group Attribution Example}
\label{app:eppc_group_attribution}
% \begin{adjustbox}{scale=0.95,center}
% \begin{minipage}{\textwidth}

% \hlbox avoids conflict with soul package's \hl{text}
\newcommand{\hlbox}[2]{%
  \begingroup
  \setlength{\fboxsep}{1.5pt}%
  \colorbox{#1}{\strut\small #2}%
  \endgroup
}

% Define color scheme for groups
\definecolor{codeA}{RGB}{255,200,200}     % InfoGive - coral
\definecolor{codeB}{RGB}{200,220,255}     % InfoGiveSDOH - periwinkle
\definecolor{codeC}{RGB}{200,255,200}     % PartnershipProvider - mint

\definecolor{subA}{RGB}{255,220,180}      % Generalinformation - peach
\definecolor{subB}{RGB}{180,240,255}      % HealthCareAccessAndQuality - sky
\definecolor{subC}{RGB}{230,200,255}      % maintainCommunication - lavender

\definecolor{SpanAll}{RGB}{255,245,200}   % All Spans - light gold

% Section background colors
\definecolor{sect1bg}{RGB}{255,245,235}   % Warm peach
\definecolor{sect2bg}{RGB}{235,250,255}   % Cool cyan
\definecolor{sect3bg}{RGB}{245,240,255}   % Soft purple

% Grouping label colors (complementary pair — darkened for clarity)
\definecolor{multiGroupColor}{RGB}{0,80,80}      % Deep teal
\definecolor{singleGroupColor}{RGB}{120,10,40}    % Deep maroon

% ==================== FIGURE 1: Note + Sample Text + Grouped by Code ====================
\begin{figure*}[!htbp]
\centering
\begin{tcolorbox}[
  enhanced,
  width=\textwidth,
  colback=blue!8,
  colframe=blue!70!black,
  boxrule=1.2pt,
  arc=3mm,
  top=3pt, bottom=3pt, left=4pt, right=4pt,
  title={\normalsize\textbf{STaR-DRO: Group Loss Attribution (Part 1 of 2)}},
  fonttitle=\bfseries\normalsize,
  coltitle=white,
  colbacktitle=blue!75!black,
]

{\itshape\small
\textbf{Note:} Colored regions indicate annotation Spans contributing to each group's loss.
Color identity shows group membership. Each annotation contributes to exactly one group per grouping dimension.
}

\vspace{1mm}

% ==================== SAMPLE TEXT ====================
\begin{tcolorbox}[
  enhanced,
  colback=yellow!10,
  colframe=orange!60,
  boxrule=0.8pt,
  arc=2mm,
  top=2pt, bottom=2pt, left=4pt, right=4pt,
]
\noindent
\textbf{Sample Text:}\\[0.5mm]
{\small\ttfamily
Person1, I submitted application look for email from Org3 in spam mail as well- let us know if you do not receive anything by MM/DD/YYYY.
}

\vspace{1mm}
\noindent
\textbf{Annotations:}
\begin{itemize}[leftmargin=*, nosep, itemsep=0pt]
\item {\small\ttfamily \{"Code":"InfoGive", "Sub-code":"Generalinformation", "Span":"I submitted application"\}}
\item {\small\ttfamily \{"Code":"InfoGiveSDOH", "Sub-code":"HealthCareAccessAndQuality", "Span":"look for email from Org3 in spam mail as well"\}}
\item {\small\ttfamily \{"Code":"PartnershipProvider", "Sub-code":"maintainCommunication", "Span":"let us know if you do not receive anything by MM/DD/YYYY"\}}
\end{itemize}
\end{tcolorbox}

\vspace{-2mm}

% ==================== SECTION 1: Group by Code ====================
\begin{tcolorbox}[
  enhanced,
  colback=sect1bg,
  colframe=red!60,
  boxrule=1pt,
  arc=2mm,
  top=2pt, bottom=2pt, left=4pt, right=4pt,
  title={\textbf{Grouped by Code}},
  fonttitle=\bfseries,
  coltitle=red!30!black,
]

\noindent
{\small {\color{multiGroupColor}\textbf{Multi-Group Sample based Grouping:}} This sample belongs to \hlbox{codeA}{\textbf{Code = InfoGive}}, \hlbox{codeB}{\textbf{Code = InfoGiveSDOH}}, and \hlbox{codeC}{\textbf{Code = PartnershipProvider}}}

\vspace{2mm}

% Annotation 1
\noindent
{\footnotesize\ttfamily
\hlbox{codeA}{\{"Code":"InfoGive",}\\
\hlbox{codeA}{"Sub-code":"Generalinformation",}\\
\hlbox{codeA}{"Span":"I submitted application"\}}
}\\[1mm]
{\small {\color{singleGroupColor}\textbf{Single-Group Annotation based Grouping:}} \hlbox{codeA}{\textbf{InfoGive}}}

\vspace{2mm}

% Annotation 2
\noindent
{\footnotesize\ttfamily
\hlbox{codeB}{\{"Code":"InfoGiveSDOH",}\\
\hlbox{codeB}{"Sub-code":"HealthCareAccessAndQuality",}\\
\hlbox{codeB}{"Span":"look for email from Org3..."\}}
}\\[1mm]
{\small {\color{singleGroupColor}\textbf{Single-Group Annotation based Grouping:}} \hlbox{codeB}{\textbf{InfoGiveSDOH}}}

\vspace{2mm}

% Annotation 3
\noindent
{\footnotesize\ttfamily
\hlbox{codeC}{\{"Code":"PartnershipProvider",}\\
\hlbox{codeC}{"Sub-code":"maintainCommunication",}\\
\hlbox{codeC}{"Span":"let us know if..."\}}
}\\[1mm]
{\small {\color{singleGroupColor}\textbf{Single-Group Annotation based Grouping:}} \hlbox{codeC}{\textbf{PartnershipProvider}}}

\end{tcolorbox}

\end{tcolorbox}
\refstepcounter{figure}
\label{fig:star-dro-groups-part1}
\end{figure*}

% ==================== FIGURE 2: Grouped by Sub-code + Grouped by NSA ====================

\begin{figure*}[!htbp]
\centering
\begin{tcolorbox}[
  enhanced,
  width=\textwidth,
  colback=blue!8,
  colframe=blue!70!black,
  boxrule=1.2pt,
  arc=3mm,
  top=3pt, bottom=3pt, left=4pt, right=4pt,
  title={\normalsize\textbf{STaR-DRO: Group Loss Attribution (Part 2 of 2)}},
  fonttitle=\bfseries\normalsize,
  coltitle=white,
  colbacktitle=blue!75!black,
]

% ==================== SECTION 2: Group by Sub-code ====================
\begin{tcolorbox}[
  enhanced,
  colback=sect2bg,
  colframe=blue!60,
  boxrule=1pt,
  arc=2mm,
  top=2pt, bottom=2pt, left=4pt, right=4pt,
  title={\textbf{Grouped by Sub-code}},
  fonttitle=\bfseries,
  coltitle=blue!40!black,
]

\noindent
{\small {\color{multiGroupColor}\textbf{Multi-Group Sample based Grouping:}} This sample belongs to \hlbox{subA}{\textbf{Generalinformation}}, \hlbox{subB}{\textbf{HealthCareAccessAndQuality}}, and \hlbox{subC}{\textbf{maintainCommunication}}}

\vspace{2mm}

% Annotation 1
\noindent
{\footnotesize\ttfamily
\hlbox{subA}{\{"Code":"InfoGive",}\\
\hlbox{subA}{"Sub-code":"Generalinformation",}\\
\hlbox{subA}{"Span":"I submitted application"\}}
}\\[1mm]
{\small {\color{singleGroupColor}\textbf{Single-Group Annotation based Grouping:}} \hlbox{subA}{\textbf{Generalinformation}}}

\vspace{2mm}

% Annotation 2
\noindent
{\footnotesize\ttfamily
\hlbox{subB}{\{"Code":"InfoGiveSDOH",}\\
\hlbox{subB}{"Sub-code":"HealthCareAccessAndQuality",}\\
\hlbox{subB}{"Span":"look for email from Org3..."\}}
}\\[1mm]
{\small {\color{singleGroupColor}\textbf{Single-Group Annotation based Grouping:}} \hlbox{subB}{\textbf{HealthCareAccessAndQuality}}}

\vspace{2mm}

% Annotation 3
\noindent
{\footnotesize\ttfamily
\hlbox{subC}{\{"Code":"PartnershipProvider",}\\
\hlbox{subC}{"Sub-code":"maintainCommunication",}\\
\hlbox{subC}{"Span":"let us know if..."\}}
}\\[1mm]
{\small {\color{singleGroupColor}\textbf{Single-Group Annotation based Grouping:}} \hlbox{subC}{\textbf{maintainCommunication}}}

\end{tcolorbox}

\vspace{-2mm}

% ==================== SECTION 3: Group by Number of Annotations ====================
\begin{tcolorbox}[
  enhanced,
  colback=sect3bg,
  colframe=purple!60,
  boxrule=1pt,
  arc=2mm,
  top=2pt, bottom=2pt, left=4pt, right=4pt,
  title={\textbf{Grouped by Number of Annotations}},
  fonttitle=\bfseries,
  coltitle=purple!40!black,
]

\noindent
{\small {\color{multiGroupColor}\textbf{Multi-Group Sample based Grouping:}} This sample belongs to \hlbox{SpanAll}{\textbf{NA (Number of Annotations) = 3}} (single group)}

\vspace{2mm}

\noindent
{\footnotesize\ttfamily
\hlbox{SpanAll}{\{"Code":"InfoGive", "Sub-code":"Generalinformation",}\\
\hlbox{SpanAll}{"Span":"I submitted application"\}}
}\\[1mm]
{\small {\color{singleGroupColor}\textbf{Single-Group Annotation based Grouping:}} \hlbox{SpanAll}{\textbf{NA\_3}}}

\vspace{2mm}

\noindent
{\footnotesize\ttfamily
\hlbox{SpanAll}{\{"Code":"InfoGiveSDOH", "Sub-code":"HealthCareAccessAndQuality",}\\
\hlbox{SpanAll}{"Span":"look for email from Org3 in spam mail as well"\}}
}\\[1mm]
{\small {\color{singleGroupColor}\textbf{Single-Group Annotation based Grouping:}} \hlbox{SpanAll}{\textbf{NA\_3}}}

\vspace{2mm}

\noindent
{\footnotesize\ttfamily
\hlbox{SpanAll}{\{"Code":"PartnershipProvider", "Sub-code":"maintainCommunication",}\\
\hlbox{SpanAll}{"Span":"let us know if you do not receive anything by MM/DD/YYYY"\}}
}\\[1mm]
{\small {\color{singleGroupColor}\textbf{Single-Group Annotation based Grouping:}} \hlbox{SpanAll}{\textbf{NA\_3}}}

\vspace{1.5mm}
{\small\itshape
\textbf{Note:} Unlike Code/Sub-code grouping, annotation-count assigns the entire sample to one group based on total annotation count.
}
\end{tcolorbox}

\end{tcolorbox}
    \refstepcounter{figure}
\label{fig:star-dro-groups-part2}
\end{figure*}
\clearpage

\section{Impact of Grouping Criteria and Sample-Level Loss on STaR-DRO Performance}
\label{sec:appendix_stardro_grouping}

This appendix presents the results on sensitivity of STaR-DRO to the choice of grouping criterion, reporting results for both the annotation-level setting and the sample-level multi-grouping setting on EPPC Miner with \texttt{Llama-3.2-3B-Instruct}.

In annotation-level mode (Table~\ref{tab:stardro_grouping_criteria}), grouping by Code yields the strongest overall default configuration, achieving the highest Code F1 while remaining competitive on Sub-code and Span. Finer-grained schemes, including Sub-code and Code$\times$Sub-code groupings, remain tightly competitive. This narrow spread suggests that STaR-DRO is robust to the choice of grouping granularity, even though the exact grouping is not known \emph{a priori}.

In sample-level multi-group mode (Table~\ref{tab:stardro_grouping_criteria_sample_multi}), grouping by Number of Annotations gives the strongest Span result while maintaining balanced Code and Sub-code performance. Taken together, these results suggest a useful distinction. Annotation-level grouping produces clearer improvements on semantic disambiguation (Code and Sub-code) because the robust multiplier is routed to individual annotation tokens: a hard annotation within an example receives elevated weight while an easy annotation in the same example remains at the neutral multiplier. In sample-level multi-grouping, this granularity is lost: all annotations within a sample inherit the same upweighting regardless of their individual difficulty, which means that well-learned and poorly-learned groups within the same example receive identical emphasis. This dilutes the robust signal on the semantic distinctions that need it most.

\begin{table*}[htbp]
\caption{Performance of STaR-DRO under different grouping criteria on the EPPC Miner benchmark for Llama-3.2-3B-Instruct.}
\centering
\resizebox{\textwidth}{!}{
\begin{tabular}{lccccccccc}
\toprule
\multirow{2}{*}{\textbf{Grouping Criterion}} & \multicolumn{3}{c}{\textbf{Code}} & \multicolumn{3}{c}{\textbf{Sub-code}} & \multicolumn{3}{c}{\textbf{Span}} \\
\cmidrule(lr){2-4} \cmidrule(lr){5-7} \cmidrule(lr){8-10}
 & \textbf{Precision} & \textbf{Recall} & \textbf{F1} & \textbf{Precision} & \textbf{Recall} & \textbf{F1} & \textbf{Precision} & \textbf{Recall} & \textbf{F1} \\
\midrule

Code & 78.13 & 81.91 & 79.98 & 63.28 & 69.04 & 66.03 & 86.46 & 96.05 & 91.00 \\
Sub-code & 76.66 & 81.79 & 79.14 & 63.49 & 68.97 & 66.12 & 87.17 & 95.80 & 91.28 \\
Number of Annotations & 76.91 & 81.66 & 79.21 & 63.01 & 68.25 & 65.53 & 87.71 & 95.69 & 91.52 \\
Code $\times$ Sub-code & 76.81 & 81.75 & 79.20 & 62.45 & 68.70 & 65.43 & 86.51 & 96.45 & 91.21 \\
Code $\times$ Sub-code $\times$ Number of Annotations & 76.54 & 81.24 & 78.82 & 62.84 & 68.14 & 65.38 & 86.99 & 95.76 & 91.17 \\
\bottomrule

\end{tabular}
}
\label{tab:stardro_grouping_criteria}
\end{table*}

\begin{table*}[htbp]
\caption{Performance of STaR-DRO under different grouping criteria on the EPPC Miner benchmark for Llama-3.2-3B-Instruct. Results are reported for the sample-level multi-grouping setting.}
\centering
\resizebox{\textwidth}{!}{
\begin{tabular}{lccccccccc}
\toprule
\multirow{2}{*}{\textbf{Grouping Criterion}} & \multicolumn{3}{c}{\textbf{Code}} & \multicolumn{3}{c}{\textbf{Sub-code}} & \multicolumn{3}{c}{\textbf{Span}} \\
\cmidrule(lr){2-4} \cmidrule(lr){5-7} \cmidrule(lr){8-10}
 & \textbf{Precision} & \textbf{Recall} & \textbf{F1} & \textbf{Precision} & \textbf{Recall} & \textbf{F1} & \textbf{Precision} & \textbf{Recall} & \textbf{F1} \\
\midrule

Code & 76.40 & 81.66 & 78.94 & 62.46 & 68.67 & 65.42 & 86.50 & 96.34 & 91.15 \\
Sub-code & 76.48 & 81.37 & 78.85 & 62.22 & 68.14 & 65.04 & 87.06 & 96.09 & 91.35 \\
Number of Annotations & 77.27 & 81.79 & 79.46 & 62.94 & 68.18 & 65.46 & 87.80 & 96.23 & 91.82 \\
Code $\times$ Sub-code & 76.13 & 81.24 & 78.60 & 62.83 & 68.93 & 65.74 & 86.54 & 96.01 & 91.03 \\
Code $\times$ Sub-code $\times$ Number of Annotations & 76.29 & 82.12 & 79.10 & 62.13 & 68.63 & 65.22 & 86.54 & 96.45 & 91.22 \\

\bottomrule
\end{tabular}
}
\label{tab:stardro_grouping_criteria_sample_multi}
\end{table*}

% =============================================================================
% Appendix: Ablation Studies on STaR-DRO Hyperparameters
% =============================================================================
% Prerequisites: \usepackage{amsmath, amssymb, booktabs, graphicx, xcolor}
% Place figures in a figures/ directory in your Overleaf project.
% =============================================================================

\clearpage

\section{Additional STaR-DRO Hyperparameter and Stability Diagnostics}
\label{app:stardro_ablation}

This appendix complements the main-paper discussion by making the STaR-DRO
update dynamics explicit under different choices of the Tsallis order $\alpha$,
mirror-ascent step size $\eta_t$, EMA smoothing coefficient $\rho_t$,
weight-decay coefficient $\lambda_{\mathrm{wd}}$, and DRO activation epoch.
These choices control or modulate the transition between uniform, selective,
and over-concentrated group-weight regimes. Unless otherwise stated, we fix the
multiplier-shaping parameters at their default values $U=15$ and $\gamma=0.75$.

\subsection{Hyperparameter Roles in the STaR-DRO Update}
\label{app:star_dro_roles}

The STaR-DRO update is governed by the sequence
\[
L_{t,g}
\;\longrightarrow\;
s_t
\;\longrightarrow\;
a_{t,g}
\;\longrightarrow\;
u_{t+\frac12,g}
\;\longrightarrow\;
q_{t+1,g}
\;\longrightarrow\;
m_{t,g}.
\]

For present groups $g \in P_t$, the relative hardness signal is
\[
a_{t,g}=\frac{L_{t,g}}{s_t},
\qquad
s_t=\sum_{g\in P_t}\pi_t(g)L_{t,g},
\qquad
\pi_t(g)=\frac{n_{t,g}}{\sum_{h\in P_t}n_{t,h}},
\]
and we set $a_{t,g}=0$ for $g\notin P_t$.
Thus, the mirror-ascent step acts on \emph{relative} current difficulty rather than raw
loss magnitude. In the scaled dual coordinates,
\[
u_{t+\frac12,g}
=
q_{t,g}^{\alpha-1}
+
(\alpha-1)\eta_t a_{t,g},
\]
followed by the entmax-style thresholded projection back to the simplex,
\[
q_{t+1,g}
=
\bigl[u_{t+\frac12,g}-\lambda_t\bigr]_+^{1/(\alpha-1)},
\qquad
\sum_{g=1}^{G} q_{t+1,g}=1,
\]
where $\lambda_t$ is the unique projection threshold chosen by bisection. The resulting simplex weights are then converted into bounded multipliers by first
defining the excess-above-uniform term
\[
e_{t,g}
=
\left[
\frac{Gq_{t+1,g}-1}{G-1}
\right]_{[0,1]},
\]
and then setting
\[
m_{t,g}
=
1+(U-1)e_{t,g}^{\gamma}
\quad\text{for } g\in P_t,
\]
with $m_{t,g}=1$ otherwise. This final map is important: the adversarial simplex
weights are \emph{not} used directly as optimizer-facing weights. STaR-DRO instead
converts only the excess mass above the uniform baseline into additional emphasis,
thereby preserving neutral supervision for groups that are not currently harder than
uniform. The ablations are easiest to interpret in terms of four interacting control quantities.

\paragraph{Effective dual step.}
A useful diagnostic quantity is
\[
\eta_{\mathrm{eff}}=(\alpha-1)\eta_t.
\]
Although the full dynamics also depend on the projection, $\eta_{\mathrm{eff}}$ captures
the scale of the additive dual update. Large $\eta_{\mathrm{eff}}$ pushes the dual iterate
too aggressively toward a small active set; very small $\eta_{\mathrm{eff}}$ leaves
$q_t$ close to its initial uniform state within the available training budget.

\paragraph{EMA memory horizon.}
The coefficient $\rho_t$ determines how quickly $L_{t,g}$ tracks newly observed group
losses. Small $\rho_t$ yields smoother, longer-horizon estimates of group difficulty.
Large $\rho_t$ makes the update more reactive, but also more sensitive to minibatch
noise. In practice, $\rho_t$ should not be interpreted in isolation: aggressive
$\eta_t$ together with aggressive $\rho_t$ is precisely the combination most likely to
produce unstable weight swings.

\paragraph{Tsallis sparsity geometry.}
The Tsallis index $\alpha>1$ controls the geometry of the simplex update. As
$\alpha \to 1^+$, the update approaches the Shannon-entropy regime and becomes less
sparse. Increasing $\alpha$ weakens the barrier against boundary solutions and makes
it easier for the projection to threshold low-priority groups to zero. In other words,
$\alpha$ controls not only the effective step through $(\alpha-1)\eta_t$, but also the
degree of active-set sparsification induced by the projection.

\paragraph{Activation schedule.}
Delaying DRO activation until epoch~2 allows the initial SFT phase to build a more
stable representation before robust redistribution begins. This matters because the
relative hardness ordering of groups is less reliable early in training, when all losses
are still dominated by undertrained features.

For the distributional diagnostics used below, we define the Shannon entropy of the
adversarial group distribution as
\[
H(q_t)=-\sum_{g=1}^{G}q_{t,g}\log q_{t,g},
\]
whose maximum is $\log G$ under the uniform distribution. We also report the
entropy-effective active-set size
\[
N_{\mathrm{eff}}(q_t)=\exp(H(q_t)),
\]
which equals $G$ for a uniform distribution and decreases as the adversarial mass
concentrates on fewer groups. When exact sparsity is relevant, we additionally inspect
the thresholded support size
\[
A_{\epsilon}(q_t)=\left|\{g:q_{t,g}>\epsilon\}\right|,
\]
with a small numerical threshold $\epsilon$.

\setcounter{figure}{10}
\begin{figure*}[!htbp]
    \centering
    \includegraphics[width=\textwidth]{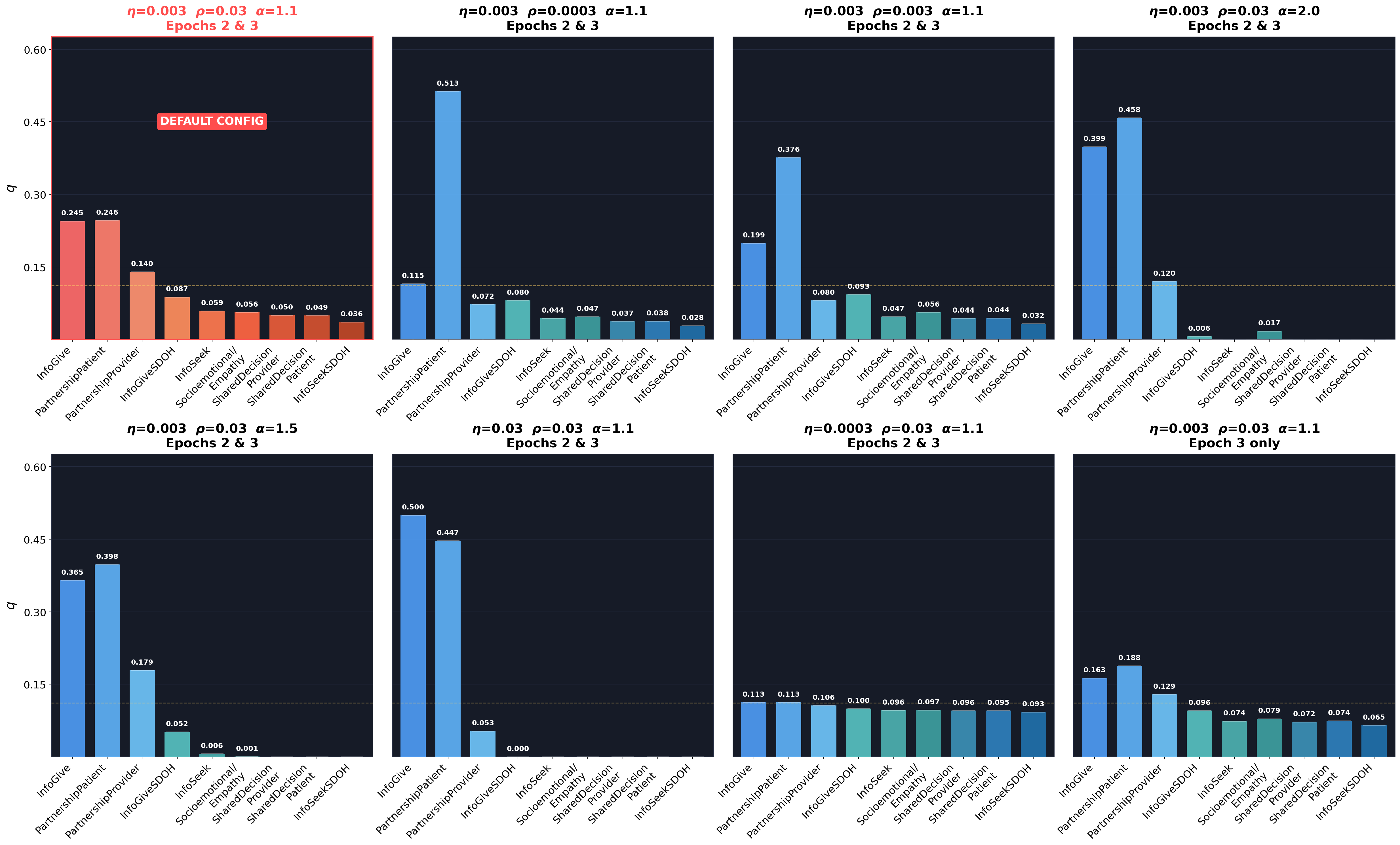}
    \caption{%
        Converged group-weight distributions $\mathbf{q}^{*}$ across eight STaR-DRO configurations on the nine Code-group EPPC Miner task with Llama-3.2-3B-Instruct. The \textbf{baseline} ($\eta_t{=}0.003$, $\rho_t{=}0.03$, $\alpha{=}1.1$, Epochs~2\,\&\,3; red bars) yields a balanced two-peak profile. Increasing $\eta_t$ or $\alpha$ drives concentration toward fewer groups; insufficiently strong settings produce near-uniform distributions. The dashed line denotes $1/G$.
    }
    \label{fig:stardro_q_distribution}
\end{figure*}

\subsection{Ablation on the Converged Group-Weight Distribution}
\label{app:star_dro_q_ablation}

Figure~\ref{fig:stardro_q_distribution} shows that the STaR-DRO hyperparameters produce three qualitatively
distinct regimes.

\paragraph{Balanced concentration.}
The default configuration produces a structured but not pathological distribution.
The converged weights form a two-peak profile, with
$q_{\mathrm{InfoGive}}^\star \approx 0.245$ and
$q_{\mathrm{PartnershipPatient}}^\star \approx 0.246$, while the remaining groups retain
non-trivial mass. The corresponding entropy
$H(q^\star)\approx 1.85$ nats remains well below the uniform maximum
$\log 9\approx 2.20$, indicating meaningful differentiation, but not active-set
collapse. This is the regime we seek: the adversary has clearly identified persistently
difficult groups, yet it still retains enough support to model intermediate group
difficulty rather than reducing the problem to a single worst-group trajectory.

\paragraph{Over-concentration.}
When $\eta_t$ is too large, when $\alpha$ is pushed too high, or when a reactive
$\rho_t$ is paired with an already aggressive dual step, the active set collapses.
Empirically, the extreme setting places more than $94\%$ of the adversarial mass on
two groups and thresholds one group to exactly zero. At that point, the update no
longer behaves like a balanced worst-group learner; it behaves like a near
winner-take-all procedure. This is undesirable for structured prediction because several
groups may have non-trivial generalization gaps even if they are not the single hardest
group at the current step. Over-concentration therefore throws away useful
group-difficulty signal.

\paragraph{Under-differentiation.}
At the other extreme, if $\eta_{\mathrm{eff}}$ is too small or if robust reweighting is
activated too late, the simplex iterate remains close to uniform. In the ablation, one
such configuration yields
$q_g \in [0.093, 0.113]$ for all nine groups, which is effectively indistinguishable
from ERM. Under-differentiation is therefore not a benign failure mode: it amounts to
paying the algorithmic complexity cost of DRO while recovering little or none of the
robustness benefit.

\paragraph{Interpretation.}
These three regimes clarify the practical role of $q_t$. The goal is not maximum
sparsity; nor is it to remain close to uniform. The goal is \emph{selective but stable}
redistribution: enough differentiation to identify persistently hard groups, but enough
spread to preserve signal from groups with moderate but still consequential
difficulty.

\paragraph{Scaling heuristic.}
The ablations support a simple rule of thumb:
\[
\eta_{\mathrm{eff}}=(\alpha-1)\eta_t \propto \frac{1}{G}.
\]
For the nine-group EPPC setting, the baseline
$\eta_{\mathrm{eff}}=0.1\times 0.003=3\times 10^{-4}$ lies in the balanced regime.
As the number of groups grows, both $\eta_t$ and $(\alpha-1)$ should be reduced so
that the dual step does not become too aggressive relative to the finer simplex
partition.

\subsection{Practical Recommendations}
\label{app:star_dro_recommendations}

The ablations suggest four practical recommendations.

\paragraph{Start conservative.}
It is easier to recover from mild under-differentiation by increasing $\eta_t$ or
$\alpha$ than it is to repair training after the active set has already collapsed.
Accordingly, practitioners should begin with $\alpha$ close to $1.1$ and a modest
mirror-ascent step. Equivalently, the effective dual step
$\eta_{\mathrm{eff}}=(\alpha-1)\eta_t$ should be initialized conservatively and reduced
as the number of groups increases.

\paragraph{Monitor distributional diagnostics, not just task loss.}
The entropy $H(q_t)$, the entropy-effective active-set size
$N_{\mathrm{eff}}(q_t)=\exp(H(q_t))$, and the thresholded support size
$A_{\epsilon}(q_t)$ provide immediate warning signals. Values of $H(q_t)$ near
$\log G$ or $N_{\mathrm{eff}}(q_t)$ near $G$ indicate that STaR-DRO is behaving too
much like ERM. Conversely, rapid collapse of $N_{\mathrm{eff}}(q_t)$ or
$A_{\epsilon}(q_t)$ indicates that the dual step is too aggressive.

\paragraph{Treat weight decay as a first-class robustness hyperparameter.}
For STaR-DRO, weight decay is not a generic optimization afterthought. It is part of
the robustness mechanism because it shapes how strongly the model can respond to
concentrated robust reweighting. In our setting, the useful regime lies roughly in the
range $\lambda_{\mathrm{wd}}\in[0.05,0.15]$, with stronger values often preferable as
group count and group sparsity increase.

\paragraph{Delay robust redistribution.}
Activating DRO only after an initial SFT phase yields more reliable estimates of
relative group hardness and avoids premature concentration on groups whose losses are
transiently large only because the representation is still immature. In practice, this
means enabling robust reweighting from epoch~2 or later, with later activation becoming
more useful as the group inventory becomes larger or sparser.

\clearpage

\section{Metric}\label{append:metric}

Each task adopts a different evaluation strategy tailored to its prediction format and semantic structure:

\paragraph{Code Classification}

The Code classification task is formulated as a multi-label classification problem over a predefined set of communicative Codes. Let $\widehat{\mathcal K}^{(i)}$ denote the predicted Code set and $\mathcal K^{(i)}$ the gold Code set for instance $i$. We compute precision recall, and F1-score as follows:

\begin{equation*}
    precision_{\text{Code}}
    =
    \frac{
    \sum_i \left|
    \widehat{\mathcal K}^{(i)}
    \cap
    \mathcal K^{(i)}
    \right|
    }{
    \sum_i \left|
    \widehat{\mathcal K}^{(i)}
    \right|
    }
\end{equation*}

\begin{equation*}
    recall_{\text{Code}}
    =
    \frac{
    \sum_i \left|
    \widehat{\mathcal K}^{(i)}
    \cap
    \mathcal K^{(i)}
    \right|
    }{
    \sum_i \left|
    \mathcal K^{(i)}
    \right|
    }
\end{equation*}

\begin{equation*}
    F1_{\text{Code}}
    =
    \frac{
    2 \times precision_{\text{Code}} \times recall_{\text{Code}}
    }{
    precision_{\text{Code}} + recall_{\text{Code}}
    }
\end{equation*}

\paragraph{Sub-code Classification}

Sub-code classification is also evaluated as a multi-label task, where each message may be annotated with one or more Sub-codes tied to a parent Code. Let $\widehat{\mathcal V}^{(i)}$ and $\mathcal V^{(i)}$ denote predicted and gold Sub-code sets, respectively. Metrics are calculated using:

\begin{equation*}
    precision_{\text{Sub-code}}
    =
    \frac{
    \sum_i \left|
    \widehat{\mathcal V}^{(i)}
    \cap
    \mathcal V^{(i)}
    \right|
    }{
    \sum_i \left|
    \widehat{\mathcal V}^{(i)}
    \right|
    }
\end{equation*}

\begin{equation*}
    recall_{\text{Sub-code}}
    =
    \frac{
    \sum_i \left|
    \widehat{\mathcal V}^{(i)}
    \cap
    \mathcal V^{(i)}
    \right|
    }{
    \sum_i \left|
    \mathcal V^{(i)}
    \right|
    }
\end{equation*}

\begin{equation*}
    F1_{\text{Sub-code}}
    =
    \frac{
    2 \times precision_{\text{Sub-code}} \times recall_{\text{Sub-code}}
    }{
    precision_{\text{Sub-code}} + recall_{\text{Sub-code}}
    }
\end{equation*}

\paragraph{Evidence Extraction}

For Span extraction, we evaluate each predicted evidence string against gold Spans using a \textbf{relaxed token-level matching strategy}, which combines:
\begin{itemize}
    \item Full containment (i.e., gold Span is fully included in predicted Span or vice versa).
    \item Jaccard similarity (between predicted and gold Spans), with a threshold of 0.6.
\end{itemize}

Let $\widehat{\Pi}^{(i)}$ and $\Pi^{(i)}$ denote the sets of predicted and gold evidence Spans (strings) for instance $i$.
A predicted Span $\hat{\pi}\in\widehat{\Pi}^{(i)}$ is counted as a \textbf{true positive (TP)} if it matches any gold Span $\pi\in\Pi^{(i)}$ such that
$\mathrm{Tok}(\pi)\subseteq \mathrm{Tok}(\hat{\pi})$ or
$\mathrm{Tok}(\hat{\pi})\subseteq \mathrm{Tok}(\pi)$ or
$\mathrm{Jaccard}(\hat{\pi},\pi)\ge 0.6$.
Spans in $\widehat{\Pi}^{(i)}$ that fail to match any gold Span are counted as \textbf{false positives (FP)},
and Spans in $\Pi^{(i)}$ not matched by any prediction are counted as \textbf{false negatives (FN)}.
Precision, recall, and F1-score are then computed as:

\begin{equation*}
    precision_{\text{Span}} = \frac{|\text{TP}|}{|\text{TP} + \text{FP}|}
\end{equation*}

\begin{equation*}
    recall_{\text{Span}} = \frac{|\text{TP}|}{|\text{TP} + \text{FN}|}
\end{equation*}

\begin{equation*}
    F1_{\text{Span}} = \frac{2 \times precision_{\text{Span}} \times recall_{\text{Span}}}{precision_{\text{Span}} + recall_{\text{Span}}}
\end{equation*}

\clearpage

\section{EPPC Miner Ontology and Label Distribution}

\begin{figure}[htbp]
\centering
\includegraphics[width=0.99\linewidth]{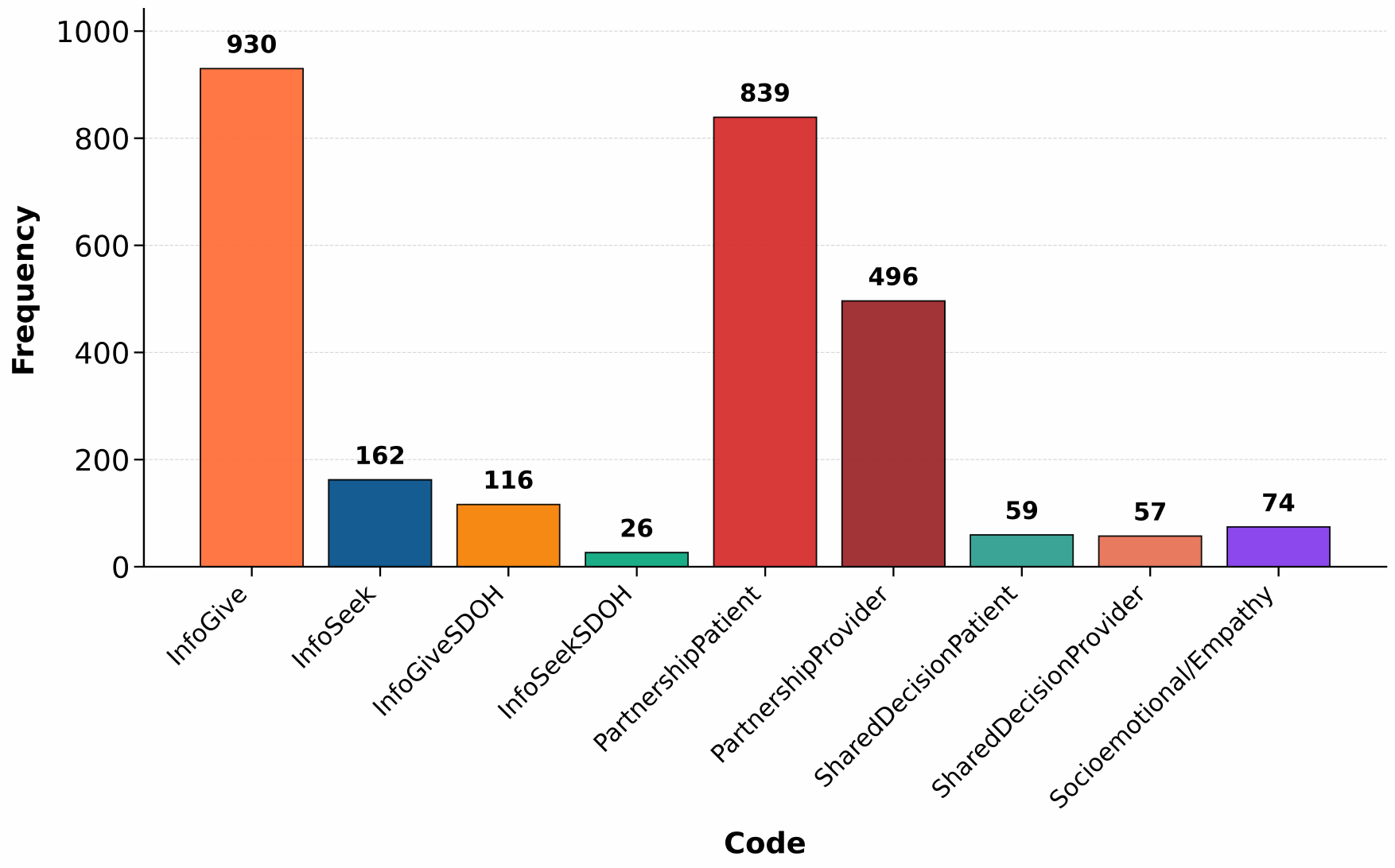}
\caption{The distribution of EPPC codes based on the annotation.}
\label{fig:eppc_codes}
\end{figure}

\begin{figure}[htbp]
\centering
\includegraphics[width=0.99\linewidth]{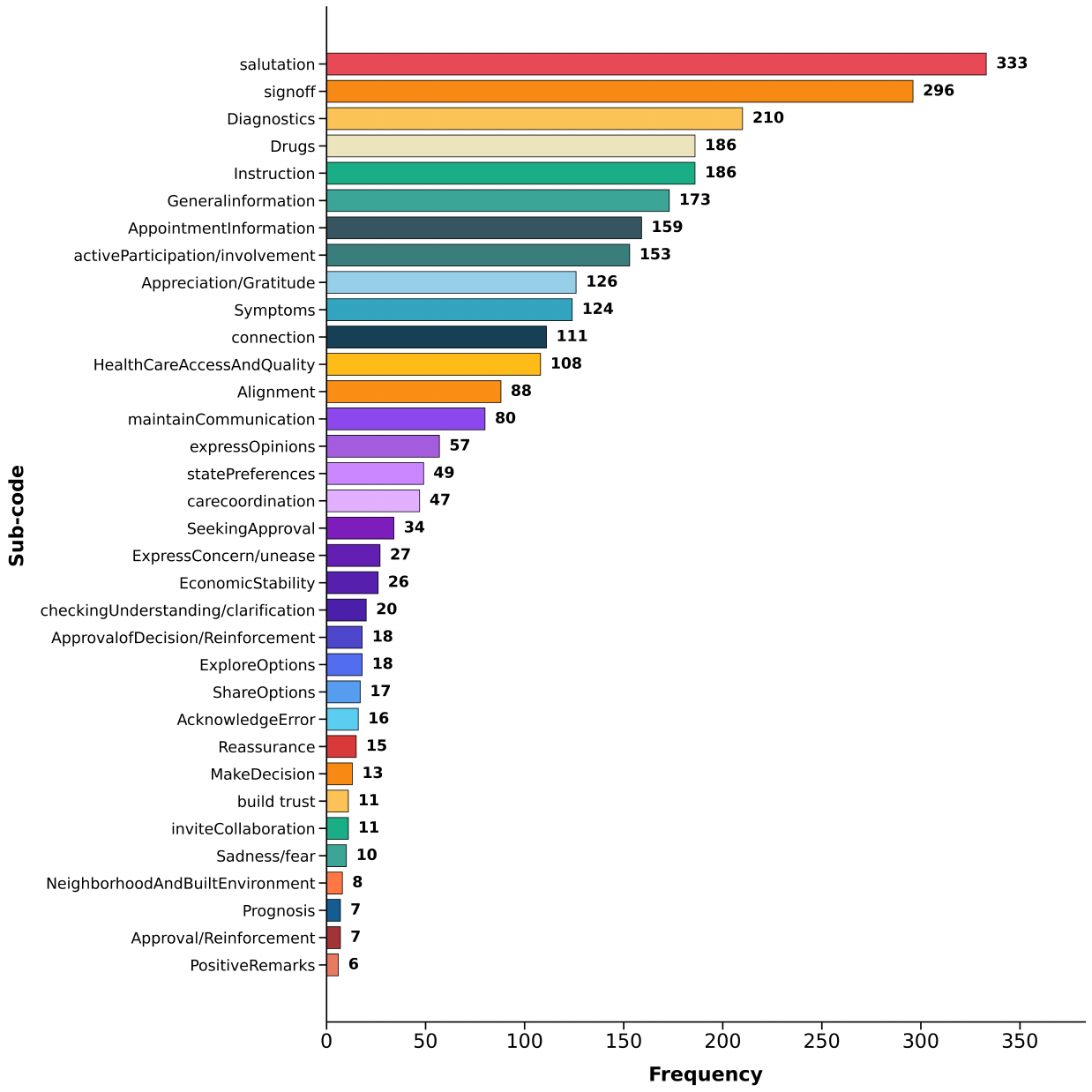}
\caption{The distribution of EPPC subcodes based on the annotation.}
\label{fig:eppc_subcodes}
\end{figure}

Figures~\ref{fig:eppc_codes} and~\ref{fig:eppc_subcodes} show the label frequency
distributions at the Code and Sub-code levels, respectively. The pronounced skew (the
two most frequent codes account for over 65\% of all Code-level labels, while the
frequency ratio between the most and least common Sub-code exceeds 55:1) is one
driver of group-difficulty heterogeneity, though not the only one; semantic ambiguity,
direction-dependent labeling, and ontological boundary proximity also contribute
independently of frequency. Figure~\ref{fig:eppc_four_categories} provides an overview of the EPPC Miner ontology by organizing the major Code categories and their associated Sub-code labels. Figure~\ref{fig:code_and_subcode} visualizes the
hierarchical validity mapping
$\mathcal{H}: L_{\text{Code}} \to 2^{L_{\text{Sub-code}}}$ that defines the
permissible Code--Sub-code pairings.

\clearpage

\begin{figure}[!htbp]
  \centering
  \includegraphics[width=\linewidth]{EPPCMiner_New.pdf}
  \caption{EPPC Miner communication codes and their associated Sub-codes.}
  \label{fig:eppc_four_categories}
\end{figure}

\clearpage

\begin{figure}[t]
  \centering
  \resizebox{0.8\linewidth}{!}{%
  \begin{tikzpicture}[
    code/.style={rectangle, rounded corners=8pt, draw=teal!80!black, fill=teal!22, thick, minimum width=2.05cm, minimum height=0.78cm},
    set/.style={rectangle, rounded corners=8pt, draw=orange!75!red, fill=orange!75!red!18, thick, minimum width=3.45cm, minimum height=1.05cm},
    node distance=1.85cm
  ]
    % ===== Group A (Codes 1–2) =====
    \node[code] (c1) {Code 1};
    \node[code, right=2.2cm of c1] (c2) {Code 2};
    \coordinate (midA) at ($(c1)!0.5!(c2)$);
    \node[set, below=1.4cm of midA] (setA) {Sub-code Set A};
    
    % ===== Group B (Codes 3–4) - ALIGNED WITH 1–2 =====
    \node[code, below=2.85cm of c1] (c3) {Code 3};
    \node[code, below=2.85cm of c2] (c4) {Code 4};
    \coordinate (midB) at ($(c3)!0.5!(c4)$);
    \node[set, below=1.4cm of midB] (setB) {Sub-code Set B};
    
    % ===== Group C (Codes 5–6) - ALIGNED WITH 1–2 =====
    \node[code, below=2.85cm of c3] (c5) {Code 5};
    \node[code, below=2.85cm of c4] (c6) {Code 6};
    \coordinate (midC) at ($(c5)!0.5!(c6)$);
    \node[set, below=1.4cm of midC] (setC) {Sub-code Set C};
    
    % ===== Disjoint group (Codes 7–9) - ADJUSTED SPACING =====
    \node[code, below=1cm of setC] (c8) {Code 8};
    \node[code, left=2.2cm of c8] (c7) {Code 7};
    \node[code, right=2.2cm of c8] (c9) {Code 9};
    \node[set, below=1.4cm of c7] (s7) {Sub-code Set D};
    \node[set, below=1.4cm of c8] (s8) {Sub-code Set E};
    \node[set, below=1.4cm of c9] (s9) {Sub-code Set F};
    
    % ===== Arrows =====
    \draw[->, thick] (c1) -- (setA);
    \draw[->, thick] (c2) -- (setA);
    \draw[->, thick] (c3) -- (setB);
    \draw[->, thick] (c4) -- (setB);
    \draw[->, thick] (c5) -- (setC);
    \draw[->, thick] (c6) -- (setC);
    \draw[->, thick] (c7) -- (s7);
    \draw[->, thick] (c8) -- (s8);
    \draw[->, thick] (c9) -- (s9);
    
    % ===== Striking curved background box =====
    \begin{scope}[on background layer]
      \node[
        rectangle,
        rounded corners=30pt,
        draw=violet!70!blue,
        fill=violet!70!blue!12,
        thick,
        inner sep=24pt,
        fit=(c1)(c2)(setA)(c3)(c4)(setB)(c5)(c6)(setC)(c7)(c8)(c9)(s7)(s8)(s9)
      ] {};
    \end{scope}
  \end{tikzpicture}
  }% end resizebox
  \caption{Illustration of the code–subcode relationship.
Codes~1--2 share identical subcode set~A (SDOH pair),
Codes~3--4 largely share set~B (InfoGive/InfoSeek pair, with one unique subcode in Code~3),
Codes~5--6 largely share set~C (Partnership pair, with several unique subcodes each),
while Codes~7--9 each correspond to fully disjoint subcode sets (D--F).
Each subcode set (A--F) is a subset of the subcode inventory $L_{sub}$,
i.e., $\text{Set} \subseteq 2^{L_{\text{Sub-code}}}$, consistent with the hierarchical
mapping $\mathcal{H}: L_{\text{Code}} \to 2^{L_{\text{Sub-code}}}$.}\label{fig:code_and_subcode}
\end{figure}

\clearpage
\section{Label Encoding Sensitivity Analysis}
\label{sec:label_encoding}

\begin{figure}[htbp]
    \centering
    \includegraphics[width=1\linewidth]{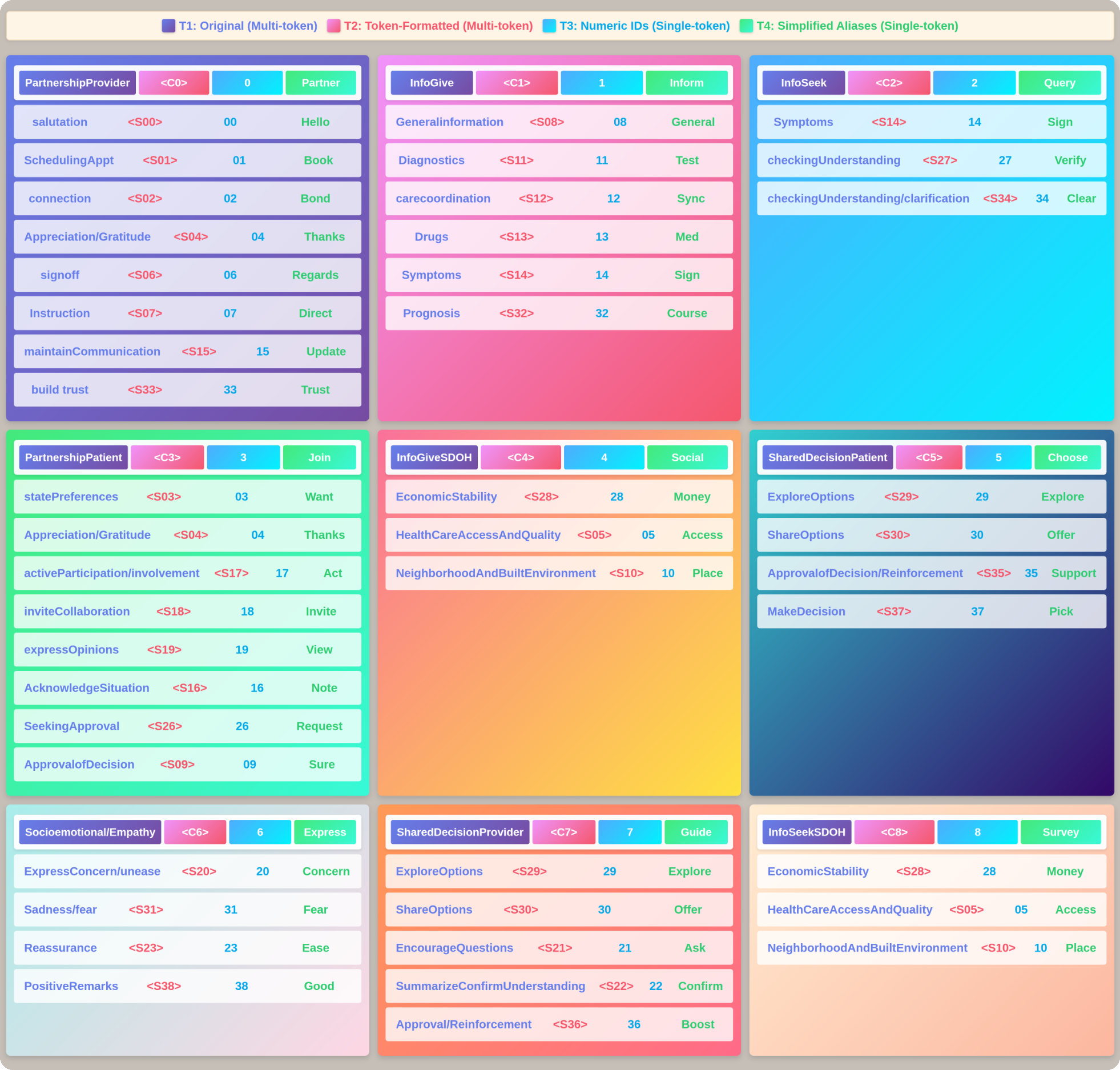}
    \caption{Label encoding schemes used in the sensitivity analysis, showing four alternative representations of EPPC Miner codes and Sub-codes: original multi-token labels, token-formatted labels, numeric IDs, and simplified aliases.}
    \label{fig:code_subcode}
\end{figure}

A defining characteristic of structured annotation tasks such as EPPC is that the target output comprises a \emph{closed set} of hierarchical labels, codes and Sub-codes drawn from a fixed codebook, rather than free-form natural language.
Because autoregressive LLMs generate outputs as sequences of tokens, the surface-form representation of these labels directly conditions the model's generation probability.
We therefore investigate a question that has received limited attention in the clinical NLP literature:
\textit{How sensitive is LLM-based structured annotation to the lexical and tokenization properties of the label encoding scheme, and does this sensitivity persist after supervised fine-tuning?}

We define four encoding schemes that vary along two orthogonal design axes: \emph{token cardinality} (whether a label decomposes into a single token or multiple tokens under the model's BPE vocabulary) and \emph{semantic transparency} (whether the surface form carries task-relevant meaning).
Figure~\ref{fig:code_subcode} illustrates the full codebook under each encoding.

\begin{itemize}
    \item \textbf{T1: Original (Multi-token).} The canonical EPPC labels as defined in the annotation codebook (e.g., \texttt{PartnershipProvider}, \texttt{salutation}, \texttt{Diagnostics}). These are descriptive, semantically transparent, and typically decompose into 2--5 BPE tokens.
    
    \item \textbf{T2: Token-Formatted (Multi-token).} XML-style identifiers using angle-bracket delimiters (e.g., \texttt{<C0>}, \texttt{<S00>}). These are semantically opaque and tokenize into four-token structured identifiers due to the bracket characters, but impose a rigid structural template.
    
    \item \textbf{T3: Numeric IDs (Single-token).} Plain integer identifiers (e.g., \texttt{0}, \texttt{00}, \texttt{08}). These are semantically opaque yet reliably single-token across all major BPE vocabularies, minimizing generation complexity at the expense of interpretability.
    
    \item \textbf{T4: Simplified Aliases (Single-token).} Short, mnemonic English words chosen to be single-token under the Llama tokenizer while retaining partial semantic relevance (e.g., \texttt{Partner}, \texttt{Hello}, \texttt{Test}, \texttt{Sync}). These represent an intentional compromise between tokenization efficiency and semantic grounding.
\end{itemize}

% ---------------------------------------------------------------------------

\begin{figure*}[htbp]
    \centering
    \begin{subfigure}{\textwidth}
        \centering
        \includegraphics[width=\textwidth]{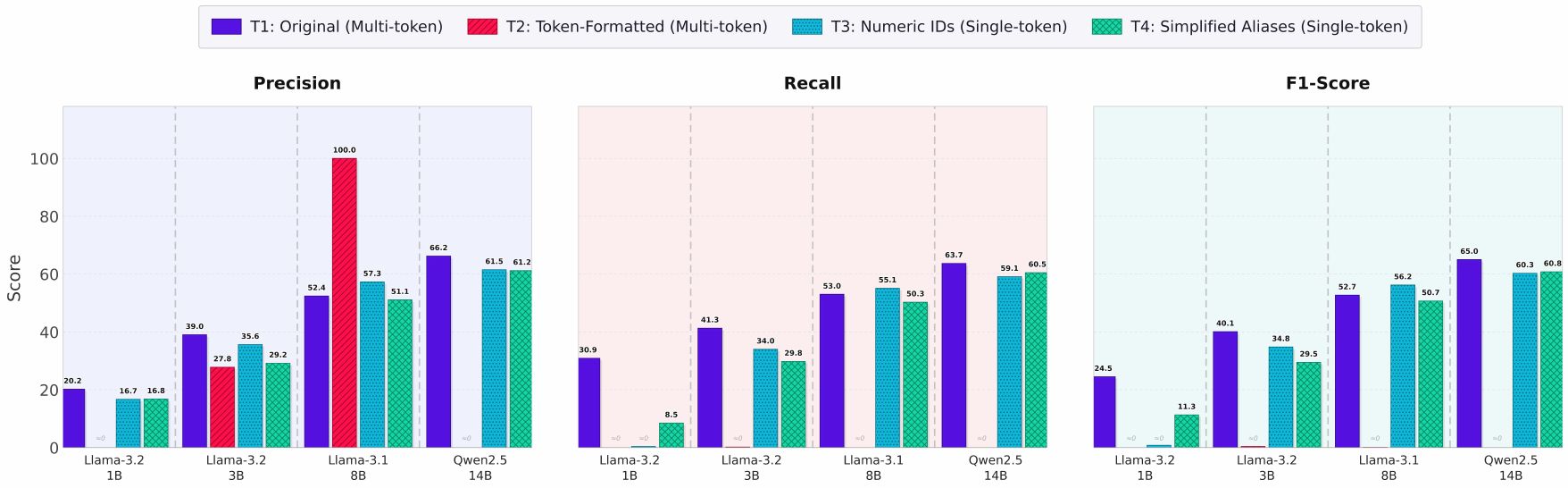}
        \label{fig:zeroshot_code}
    \end{subfigure}
    
    \vspace{0.25em}
    
    \begin{subfigure}{\textwidth}
        \centering
        \includegraphics[width=\textwidth]{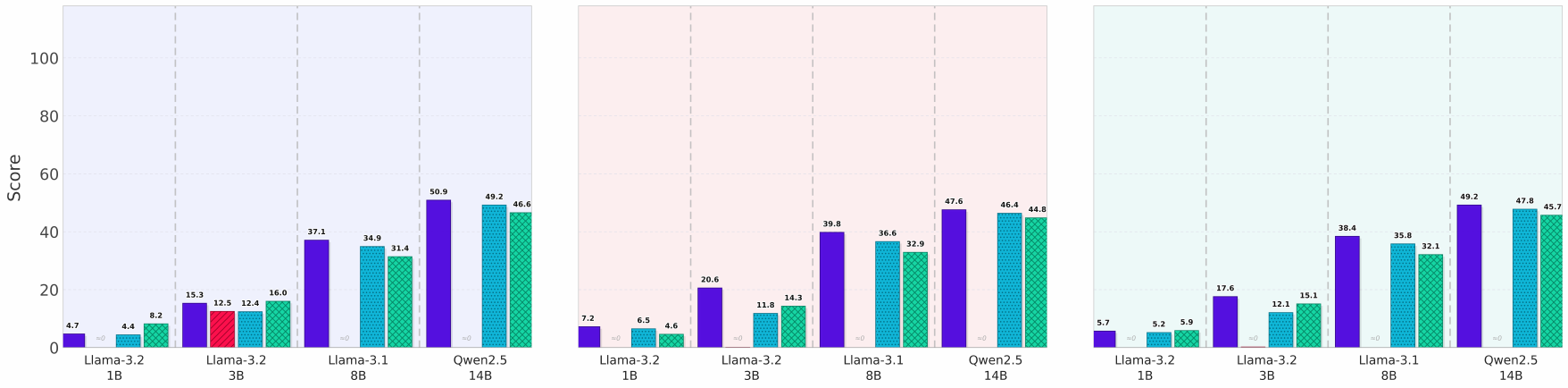}
        \label{fig:zeroshot_subcode}
    \end{subfigure}
    
    \vspace{0.25em}
    
    \begin{subfigure}{\textwidth}
        \centering
        \includegraphics[width=\textwidth]{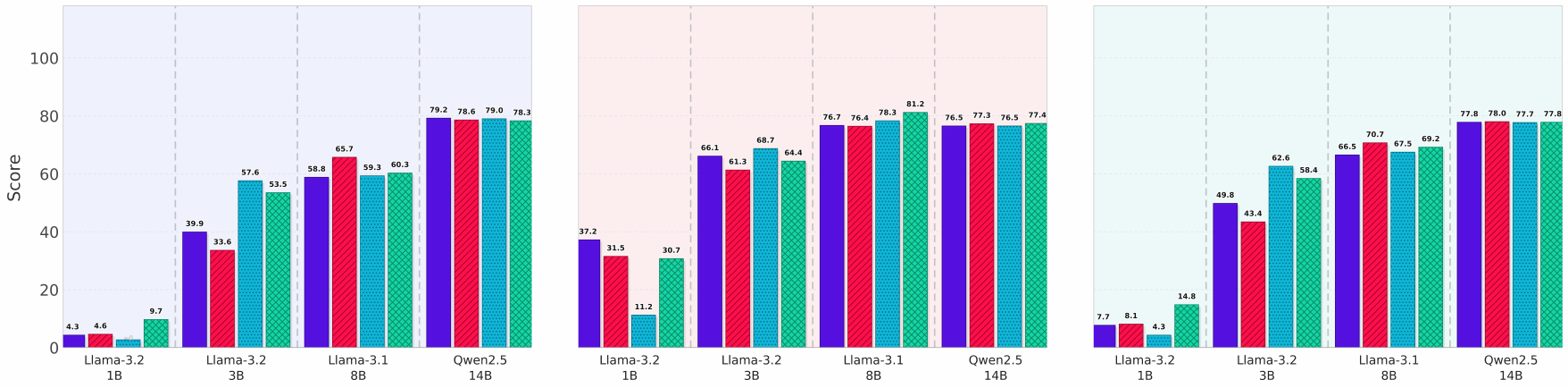}
        \label{fig:zeroshot_Span}
    \end{subfigure}
    
    \caption{Zero-shot code-level performance across label encoding strategies (T1--T4) and model scales. Precision, Recall, and F1-Score are reported for (a) code-level, (b) Sub-code-level, and (c) Span-level annotations.}
    \label{fig:zeroshot_all}
\end{figure*}

\subsubsection{Zero-Shot Performance}
\label{sec:zeroshot_encoding}

Figure~\ref{fig:zeroshot_all} report zero-shot precision, recall, and F1 across all five model scales (Llama-3.2~1B, Llama-3.2~3B, Llama-3.1~8B, Qwen2.5~14B) under each encoding scheme at the code, Sub-code, and Span levels, respectively.

\paragraph{Intuitive observations.}

\begin{enumerate}[label=(\roman*)]

\item \textbf{Semantic transparency dominates zero-shot performance.}
Across all model scales and evaluation granularities, the original multi-token encoding (T1) consistently achieves the highest F1 scores. At the Sub-ode level, T1 yields an F1 of 38.4 for Llama-3.1~8B, compared to 0, 35.8, and 32.1 for T2, T3, and T4, respectively.
The pattern is consistent with the hypothesis that semantically transparent labels activate relevant pre-trained representations, enabling the model to leverage its world knowledge about clinical communication concepts (e.g., ``Diagnostics,'' ``Reassurance,'' ``Prognosis'') even without task-specific training.

\item \textbf{Performance scales monotonically with model capacity.}
All encoding schemes exhibit monotonically increasing F1 with model size, consistent with established scaling laws for in-context learning~\citep{kaplan2020scaling, wei2022emergent}.
However, the \emph{rate} of improvement differs across encodings: T1 exhibits the steepest scaling trajectory, suggesting that larger models are disproportionately better at leveraging semantic label content for structured generation.
However, the \emph{rate} of improvement differs across encodings for fine-grained
labels. T1 shows the strongest absolute scaling at the Sub-code level, suggesting
that larger models are better able to leverage semantic label content for nuanced
structured-generation decisions.For instance, at the Sub-code level, T1 improves from 5.7 F1 on the 1B model
to 49.2 F1 on the Qwen2.5~14B model, an 8.63$\times$ increase and a +43.5 F1
gain. T3 improves from 5.2 to 47.8 F1 over the same model scale, a
9.19$\times$ increase and a +42.6 F1 gain. Thus, while both encodings benefit
substantially from scale, the semantically transparent T1 encoding reaches the
stronger endpoint and yields the larger absolute improvement.

\item \textbf{Sub-code granularity amplifies encoding sensitivity.}
The performance gap between T1 and alternative encodings widens as annotation granularity increases from code-level (9 classes) to Sub-code-level (38 classes).
This is expected: finer-grained label sets increase the discriminative burden on the model, making the semantic cues embedded in T1 labels more valuable for disambiguation among closely related Sub-codes (e.g., distinguishing \texttt{ExploreOptions} under \texttt{SharedDecisionProvider} from \texttt{ExploreOptions} under \texttt{SharedDecisionPatient}). This effect is amplified at the Sub-code level because, once the correct parent Code
is identified, the model must choose among a restricted but semantically fine-grained
Sub-code space. Consequently, for role-sensitive parent Codes such as
\texttt{SharedDecisionProvider}, \texttt{SharedDecisionPatient},
\texttt{PartnershipProvider}, and \texttt{PartnershipPatient}, lexical cues such as
\texttt{Provider} and \texttt{Patient} reinforce the metadata-conditioned logic used to
distinguish message direction and speaker role. These cues are absent or obscured in other encodings, making fine-grained disambiguation more difficult.

\end{enumerate}

\paragraph{Non-intuitive observations.}

\begin{enumerate}[label=(\roman*)]

\item \textbf{Angle-bracket-delimited labels (T2) are effectively unreachable regardless of model scale.}
Across model scales, T2 labels (e.g., \texttt{<C0>}, \texttt{<S00>}) yield near-zero Code F1, with the model almost entirely failing to produce them. The nominal precision of 1.00 observed at the 8B scales is an artifact of one or two isolated correct predictions rather than a meaningful capability. The most plausible explanation is that these token forms resemble reserved special tokens such as \texttt{<eos>}, \texttt{<bos>}, and \texttt{<pad>}: during pre-training, the model learns that angle-bracket-delimited sequences are control tokens that should not be emitted as part of natural generation. This learned suppression is sufficiently deep that even a 70B-parameter model, which otherwise demonstrates strong zero-shot structured generation with semantically transparent labels, cannot override it through in-context instruction alone. The observation delineates a hard boundary on zero-shot generalization: autoregressive LLMs generalize readily to novel \emph{compositions} of familiar tokens, but not to the spontaneous production of token forms that overlap with suppressed control-token patterns from pre-training. For structured annotation in zero-shot deployment, label design must therefore respect the model's pre-trained generation constraints as a non-negotiable requirement.

\item \textbf{Single-token encodings do not uniformly outperform multi-token encodings.}
A reasonable prior expectation is that single-token labels (T3, T4) should outperform multi-token labels (T1, T2) due to reduced autoregressive generation complexity: emitting a single token requires only one correct decoding step, whereas multi-token labels demand a sequence of correct predictions.
Our results contradict this hypothesis. T1 (multi-token, semantically transparent) consistently outperforms both T3 (single-token, semantically opaque) and T4 (single-token, partially semantic).
This suggests that in zero-shot structured generation, \emph{semantic alignment with pre-trained representations dominates tokenization efficiency} as a performance driver.
The model's ability to ground label semantics in its parametric knowledge outweighs the mechanical advantage of fewer decoding steps.

\item \textbf{Simplified aliases (T4) do not recover T1 performance despite partial semantic grounding.}
Although T4 labels were designed to preserve semantic relevance while maintaining single-token efficiency (e.g., \texttt{Trust} for \texttt{build trust}, \texttt{Concern} for \texttt{ExpressConcern/unease}), they consistently underperform T1.
This indicates that the \emph{specificity} and \emph{compositional structure} of the original labels which often encode hierarchical relationships (e.g., \texttt{PartnershipProvider} implying both the communication function and the speaker role) contribute meaningful signal that single-word aliases did not capture.
The performance gap between T1 and T4 thus reveals the importance of \emph{compositional semantics} in label design for autoregressive models, even at the cost of increased token count.

\item \textbf{Span-level metrics exhibit reduced encoding sensitivity.}
At the Span extraction level, performance differences across encodings narrow considerably relative to code- and Sub-code-level metrics.
For instance, at the 14B scale, Span F1 ranges from 77.7 to 78.0 across all four encodings, compared to Sub-code level F1 ranging from 45.7 to 49.2.
This attenuation reflects the fact that Span identification relies primarily on the model's ability to locate relevant text segments, a capability grounded in general language understanding rather than label-specific generation, and is therefore less sensitive to the surface form of the labels themselves.

\end{enumerate}

% ---------------------------------------------------------------------------

\begin{table*}[htbp]
\caption{\textbf{Label Encoding Sensitivity under SFT.} Results are reported as Precision, Recall, and F1. T1 = Original (Multi-token), T2 = Token-Formatted (Multi-token), T3 = Numeric IDs (Single-token), and T4 = Simplified Aliases (Single-token).}
\centering
\resizebox{\textwidth}{!}{
\begin{tabular}{llccccccccc}
\toprule
\multirow{2}{*}{\textbf{Category}} & \multirow{2}{*}{\textbf{Model}} & \multicolumn{3}{c}{\textbf{Code}} & \multicolumn{3}{c}{\textbf{Sub-code}} & \multicolumn{3}{c}{\textbf{Span}} \\
\cmidrule(lr){3-5} \cmidrule(lr){6-8} \cmidrule(lr){9-11}
 &  & \textbf{Precision} & \textbf{Recall} & \textbf{F1} & \textbf{Precision} & \textbf{Recall} & \textbf{F1} & \textbf{Precision} & \textbf{Recall} & \textbf{F1} \\
\midrule

\multirow{4}{*}{T1: Original (Multi-token)}
 & Qwen2.5-14B-Instruct & 77.86 & 83.12 & 80.40 & 64.84 & 71.33 & 67.93 & 86.75 & 96.34 & 91.29 \\
 & Llama-3.1-8B-Instruct & 75.96 & 81.45 & 78.61 & 63.58 & 69.75 & 66.52 & 86.86 & 96.52 & 91.43 \\
 & Llama-3.2-3B-Instruct & 76.29 & 81.58 & 78.85 & 62.05 & 68.03 & 64.90 & 86.84 & 96.59 & 91.46 \\
 & Llama-3.2-1B-Instruct & 75.65 & 80.20 & 77.86 & 61.41 & 65.97 & 63.21 & 86.91 & 96.01 & 91.23 \\
\midrule

\multirow{4}{*}{T2: Token-Formatted (Multi-token)}
 & Qwen2.5-14B-Instruct & 77.38 & 82.75 & 79.98 & 64.08 & 70.54 & 67.15 & 87.11 & 96.01 & 91.34 \\
 & Llama-3.1-8B-Instruct & 77.04 & 81.70 & 79.30 & 64.38 & 69.57 & 66.87 & 87.38 & 96.12 & 91.54 \\
 & Llama-3.2-3B-Instruct & 76.92 & 81.29 & 79.04 & 63.05 & 67.54 & 65.22 & 87.46 & 96.30 & 91.67 \\
 & Llama-3.2-1B-Instruct & 76.17 & 79.32 & 77.72 & 60.03 & 63.94 & 61.92 & 86.98 & 96.34 & 91.42 \\
\midrule

\multirow{4}{*}{T3: Numeric IDs (Single-token)}
 & Qwen2.5-14B-Instruct & 76.76 & 82.62 & 79.58 & 64.51 & 70.73 & 67.48 & 86.74 & 96.01 & 91.14 \\
 & Llama-3.1-8B-Instruct & 76.58 & 82.25 & 79.32 & 64.02 & 70.24 & 66.99 & 86.28 & 96.41 & 91.06 \\
 & Llama-3.2-3B-Instruct & 76.48 & 81.24 & 78.79 & 62.40 & 68.18 & 65.16 & 86.81 & 96.16 & 91.25 \\
 & Llama-3.2-1B-Instruct & 76.28 & 79.66 & 77.93 & 60.58 & 65.03 & 62.73 & 86.38 & 96.34 & 91.09 \\
\midrule

\multirow{4}{*}{T4: Simplified Aliases (Single-token)}
 & Qwen2.5-14B-Instruct & 77.55 & 82.96 & 80.16 & 64.99 & 71.33 & 68.01 & 86.77 & 96.01 & 91.16 \\
 & Llama-3.1-8B-Instruct & 76.53 & 81.70 & 79.03 & 63.67 & 69.23 & 66.33 & 87.04 & 95.94 & 91.28 \\
 & Llama-3.2-3B-Instruct & 75.95 & 82.04 & 78.88 & 62.47 & 68.37 & 65.28 & 87.05 & 96.70 & 91.62 \\
 & Llama-3.2-1B-Instruct & 75.17 & 78.40 & 76.75 & 59.99 & 64.28 & 62.06 & 87.00 & 95.83 & 91.20 \\
\bottomrule
\end{tabular}
}
\label{tab:label_encoding_sensitivity_sft}
\end{table*}

\subsubsection{Post-SFT Performance}
\label{sec:sft_encoding}

Tables~\ref{tab:label_encoding_sensitivity_sft} and~\ref{tab:label_encoding_sensitivity_tw} report the performance of all four encoding schemes after supervised fine-tuning (SFT) and SFT with token-level loss weighting (SFT+TW), respectively.

\paragraph{Intuitive observations.}

\begin{enumerate}[label=(\roman*)]

\item \textbf{SFT substantially improves performance across all encodings.}
Consistent with established findings on instruction tuning and task-specific adaptation~\citep{wei2022finetuned, hu2022lora}, supervised fine-tuning yields large absolute gains over zero-shot performance for every encoding scheme.

\item \textbf{Larger models retain their advantage post-SFT.}
The rank ordering of model scales is preserved after fine-tuning, with larger models achieving higher absolute performance.
This is consistent with the observation that SFT adapts rather than replaces pre-trained capabilities: larger models bring richer representational capacity to the fine-tuning process, yielding higher ceilings for task-specific performance.

\end{enumerate}

\paragraph{Non-intuitive observations.}

\begin{enumerate}[label=(\roman*)]

\item \textbf{SFT normalizes encoding-induced performance variance.}
Supervised fine-tuning dramatically compresses
the performance gap between encoding schemes. This \emph{encoding normalization
effect} indicates that, given sufficient task-specific supervision, the cross-entropy
training objective enables the model to learn reliable input--output mappings even
when the label surface forms differ in semantic transparency or tokenization structure.
In other words, SFT directly optimizes the conditional generation probability
$p_\theta(y\mid x)$ for the task-specific output strings, reducing the model's reliance
on pre-trained lexical associations or tokenization advantages present in the
zero-shot setting.

\item \textbf{Semantically opaque encodings close the gap most rapidly.}
T2 and T3, which exhibit the largest zero-shot deficits relative to T1, show the greatest relative improvement after SFT.
This asymmetric improvement pattern suggests that fine-tuning does not merely scale all encodings proportionally but preferentially benefits encodings that are poorly served by pre-trained representations.
The model effectively \emph{learns to generate} the arbitrary token sequences required by T2 and T3, establishing novel token--concept associations through gradient-based optimization.
This finding resonates with recent work on the capacity of parameter-efficient fine-tuning methods to rapidly adapt LLM generation distributions to out-of-distribution output formats~\citep{hu2022lora}.

\item \textbf{Token-level loss weighting does not preferentially benefit any encoding.}
The application of token-level loss weighting (SFT+TW), which upweights the loss contribution of label tokens relative to template tokens, yields modest and uniform improvements across all four encodings.
This uniformity indicates that the weighting mechanism addresses a \emph{task-level} challenge (focusing optimization on discriminative label tokens rather than predictable template syntax) rather than an \emph{encoding-level} one.
The finding further supports the conclusion that encoding differences are primarily a zero-shot phenomenon, effectively resolved by any form of supervised adaptation.

\end{enumerate}

\begin{table*}[htbp]
\caption{\textbf{Label Encoding Sensitivity under SFT + Token Weighting.} Results are reported as Precision, Recall, and F1 in percentage. Panel A uses $w_{\text{Code}}=3$, $w_{\text{Sub-code}}=3$, and $w_{\text{Span}}=3$. Panel B uses $w_{\text{Code}}=2$, $w_{\text{Sub-code}}=1.5$, and $w_{\text{Span}}=1$.}
\centering
\resizebox{\textwidth}{!}{
\begin{tabular}{lllccccccccc}
\toprule
\multirow{2}{*}{\textbf{Setting}} &
\multirow{2}{*}{\textbf{Category}} & \multirow{2}{*}{\textbf{Model}} &
\multicolumn{3}{c}{\textbf{Code}} & \multicolumn{3}{c}{\textbf{Sub-code}} & \multicolumn{3}{c}{\textbf{Span}} \\
\cmidrule(lr){4-6} \cmidrule(lr){7-9} \cmidrule(lr){10-12}
&  &  & \textbf{Precision} & \textbf{Recall} & \textbf{F1} & \textbf{Precision} & \textbf{Recall} & \textbf{F1} & \textbf{Precision} & \textbf{Recall} & \textbf{F1} \\
\midrule

\multirow{16}{*}{$w_{\text{Code}}=3,\; w_{\text{Sub-code}}=3,\; w_{\text{Span}}=3$}
& \multirow{4}{*}{T1: Original (Multi-token)}
 & Qwen2.5-14B-Instruct & 77.25 & 82.41 & 79.75 & 64.62 & 70.24 & 67.31 & 87.70 & 95.65 & 91.50 \\
&  & Llama-3.1-8B-Instruct & 76.31 & 81.83 & 78.98 & 63.56 & 69.57 & 66.43 & 87.36 & 96.23 & 91.58 \\
&  & Llama-3.2-3B-Instruct & 75.92 & 80.62 & 78.20 & 61.65 & 67.13 & 64.27 & 87.68 & 95.94 & 91.62 \\
&  & Llama-3.2-1B-Instruct & 75.56 & 78.78 & 77.14 & 61.56 & 64.66 & 63.07 & 89.00 & 95.32 & 92.05 \\
\cmidrule(lr){2-12}

& \multirow{4}{*}{T2: Token-Formatted (Multi-token)}
 & Qwen2.5-14B-Instruct & 77.83 & 83.29 & 80.47 & 65.48 & 71.51 & 68.36 & 87.94 & 96.16 & 91.86 \\
&  & Llama-3.1-8B-Instruct & 76.85 & 81.95 & 79.32 & 63.78 & 69.83 & 66.67 & 87.62 & 96.16 & 91.69 \\
&  & Llama-3.2-3B-Instruct & 76.89 & 80.62 & 78.71 & 63.78 & 68.25 & 65.94 & 88.45 & 96.01 & 92.08 \\
&  & Llama-3.2-1B-Instruct & 75.88 & 79.49 & 77.64 & 60.28 & 63.94 & 62.06 & 88.31 & 95.54 & 91.78 \\
\cmidrule(lr){2-12}

& \multirow{4}{*}{T3: Numeric IDs (Single-token)}
 & Qwen2.5-14B-Instruct & 76.43 & 82.37 & 79.29 & 63.90 & 70.39 & 66.99 & 87.25 & 96.01 & 91.42 \\
&  & Llama-3.1-8B-Instruct & 76.95 & 81.58 & 79.20 & 64.25 & 69.45 & 66.75 & 88.52 & 95.58 & 91.91 \\
&  & Llama-3.2-3B-Instruct & 76.66 & 80.66 & 78.61 & 63.17 & 67.62 & 65.31 & 87.68 & 95.43 & 91.39 \\
&  & Llama-3.2-1B-Instruct & 77.39 & 77.36 & 77.38 & 62.06 & 62.22 & 62.14 & 90.96 & 93.73 & 92.32 \\
\cmidrule(lr){2-12}

& \multirow{4}{*}{T4: Simplified Aliases (Single-token)}
 & Qwen2.5-14B-Instruct & 77.34 & 83.25 & 80.19 & 64.15 & 70.84 & 67.33 & 87.04 & 96.38 & 91.47 \\
&  & Llama-3.1-8B-Instruct & 76.33 & 82.04 & 79.08 & 63.76 & 69.38 & 66.45 & 88.05 & 95.90 & 91.81 \\
&  & Llama-3.2-3B-Instruct & 77.22 & 81.70 & 79.40 & 62.54 & 67.77 & 65.05 & 87.77 & 95.69 & 91.56 \\
&  & Llama-3.2-1B-Instruct & 77.00 & 77.07 & 77.04 & 62.86 & 62.11 & 62.48 & 91.84 & 91.74 & 91.79 \\
\midrule

\multirow{16}{*}{$w_{\text{Code}}=2,\; w_{\text{Sub-code}}=1.5,\; w_{\text{Span}}=1$}
& \multirow{4}{*}{T1: Original (Multi-token)}
 & Qwen2.5-14B-Instruct & 77.72 & 83.04 & 80.29 & 64.34 & 70.80 & 67.42 & 86.85 & 96.23 & 91.30 \\
&  & Llama-3.1-8B-Instruct & 76.68 & 82.41 & 79.44 & 63.38 & 69.60 & 66.35 & 87.20 & 96.34 & 91.54 \\
&  & Llama-3.2-3B-Instruct & 76.24 & 81.62 & 78.84 & 63.01 & 69.42 & 66.06 & 86.52 & 96.30 & 91.15 \\
&  & Llama-3.2-1B-Instruct & 74.72 & 79.87 & 77.21 & 60.99 & 65.82 & 63.31 & 87.00 & 96.09 & 91.32 \\
\cmidrule(lr){2-12}

& \multirow{4}{*}{T2: Token-Formatted (Multi-token)}
 & Qwen2.5-14B-Instruct & 77.03 & 82.50 & 79.67 & 64.51 & 71.33 & 67.75 & 86.82 & 96.48 & 91.40 \\
&  & Llama-3.1-8B-Instruct & 76.33 & 81.75 & 78.94 & 63.66 & 69.68 & 66.54 & 87.14 & 96.52 & 91.59 \\
&  & Llama-3.2-3B-Instruct & 77.43 & 81.41 & 79.37 & 63.63 & 69.30 & 66.34 & 86.66 & 96.27 & 91.21 \\
&  & Llama-3.2-1B-Instruct & 75.86 & 79.95 & 77.85 & 60.10 & 64.54 & 62.24 & 86.85 & 96.23 & 91.30 \\
\cmidrule(lr){2-12}

& \multirow{4}{*}{T3: Numeric IDs (Single-token)}
 & Qwen2.5-14B-Instruct & 77.58 & 83.25 & 80.31 & 63.71 & 69.94 & 66.68 & 86.79 & 96.45 & 91.36 \\
&  & Llama-3.1-8B-Instruct & 75.93 & 81.83 & 78.77 & 63.23 & 69.04 & 66.01 & 86.99 & 96.23 & 91.38 \\
&  & Llama-3.2-3B-Instruct & 76.46 & 81.66 & 78.97 & 62.05 & 68.33 & 65.04 & 86.60 & 96.70 & 91.37 \\
&  & Llama-3.2-1B-Instruct & 76.66 & 79.45 & 78.03 & 60.39 & 64.81 & 62.52 & 86.90 & 96.41 & 91.41 \\
\cmidrule(lr){2-12}

& \multirow{4}{*}{T4: Simplified Aliases (Single-token)}
 & Qwen2.5-14B-Instruct & 77.57 & 83.21 & 80.29 & 63.41 & 69.64 & 66.38 & 86.88 & 96.52 & 91.45 \\
&  & Llama-3.1-8B-Instruct & 76.43 & 81.83 & 79.04 & 62.83 & 68.67 & 65.62 & 87.30 & 96.41 & 91.63 \\
&  & Llama-3.2-3B-Instruct & 76.69 & 82.58 & 79.53 & 62.87 & 68.55 & 65.59 & 86.98 & 96.12 & 91.32 \\
&  & Llama-3.2-1B-Instruct & 76.28 & 79.24 & 77.73 & 61.06 & 65.07 & 63.00 & 87.66 & 95.51 & 91.41 \\
\bottomrule
\end{tabular}
}
\label{tab:label_encoding_sensitivity_tw}
\end{table*}

% ---------------------------------------------------------------------------
\subsubsection{Synthesis and Recommendations}
\label{sec:encoding_recommendations}

The above findings yield several practical recommendations for structured annotation in clinical and other specialized domains:

\begin{enumerate}
    \item \textbf{In zero-shot or few-shot deployment, label design is a first-order decision.} Semantically transparent, descriptive labels that align with the model's pre-trained token distribution should be preferred. Angle-bracket-delimited tokens that resemble control sequences are effectively unreachable, and arbitrary numeric identifiers sacrifice the semantic prior that drives zero-shot performance. When fine-tuning is not feasible, label verbalization should be treated as a tunable hyperparameter subject to systematic optimization, analogous to prompt engineering itself.

    \item \textbf{After supervised fine-tuning, encoding choice becomes secondary.} SFT normalizes encoding-induced performance differences to within 1--2 F1 points, so practitioners may select encodings based on engineering considerations: parsing reliability, integration with downstream pipelines, or output token efficiency, rather than expected task-level performance impact.

\item \textbf{For hierarchical label sets, compositional structure matters more than token economy.}
Multi-token labels that preserve role-sensitive and ontology-relevant structure
(e.g., the \texttt{Provider}/\texttt{Patient} distinctions in
\texttt{PartnershipProvider}, \texttt{PartnershipPatient},
\texttt{SharedDecisionPatient}, and \texttt{SharedDecisionProvider}) often outperform
single-token simplifications in zero-shot settings, even when the single-token alias
retains partial semantic relevance. This advantage suggests that the model can exploit
compositional semantics in the label surface form, especially when those components
align with metadata-conditioned decision logic and hierarchical Code--Sub-code
constraints. In contrast, compressed single-token encodings reduce decoding length but
can discard lexical cues that help resolve role-sensitive and fine-grained label
distinctions.
\end{enumerate}

\end{appendices}

\end{document}